\RequirePackage{fix-cm}
\documentclass[twocolumn]{svjour3}          % twocolumn
\smartqed  % flush right qed marks, e.g. at end of proof
\usepackage{graphicx}
\usepackage{url}
\usepackage{array}
\usepackage{amsmath}
\usepackage{float}
\usepackage{color}

\usepackage{booktabs}
\usepackage{multirow} 
\usepackage{ulem}
\usepackage{soul}
\usepackage{algorithm}
\usepackage{algorithmicx}  
\usepackage{algpseudocode}
\usepackage{amssymb}
\usepackage{booktabs}
\usepackage{bbding}
\usepackage{overpic}
\usepackage{subfigure}
\usepackage{makecell}
\usepackage{array}
\usepackage{colortbl,xcolor}
\usepackage{amssymb}
\usepackage{subfigure}
\usepackage{colortbl,xcolor}
\usepackage{pifont}
\usepackage[breaklinks=true,colorlinks,citecolor=blue,urlcolor=blue,linkcolor=blue,bookmarks=false,pagebackref=true]{hyperref}

\definecolor{Gray}{gray}{0.8}
\newcommand{\cmark}{\ding{51}}
\newcommand{\xmark}{\ding{55}}

\newcolumntype{M}[1]{>{\centering\arraybackslash}m{#1}}

\definecolor{bblue}{rgb}{0,150,230}
\definecolor{mygray}{gray}{.9}
\definecolor{lightgray}{gray}{.96}
\definecolor{myy}{RGB}{126,95,0}
\definecolor{ggray}{RGB}{127,127,127}
\definecolor{mygreen}{RGB}{93,173,85}
\definecolor{myred}{RGB}{240,16,89}
\definecolor{myblue}{RGB}{0,114,188}
\definecolor{darkgreen}{rgb}{0.0, 0.5, 0.0}
\definecolor{demphcolor}{RGB}{100,100,100}
\newcolumntype{C}[1]{>{\centering\let\newline\\\arraybackslash\hspace{0pt}}m{#1}}

\def\swten{0.11\linewidth}

\DeclareRobustCommand\onedot{\futurelet\@let@token\@onedot}
\newcommand{\etal}{\textit{et al}. }
\newcommand{\ie}{\textit{i.e., }}
\newcommand{\eg}{\textit{e.g., }}

\makeatother

\newcolumntype{d}[1]{>{\raggedright\arraybackslash}p{#1pt}}
\newcolumntype{e}[1]{>{\raggedleft\arraybackslash}p{#1pt}}

\newcommand{\firstone}[1]{\colorbox{red!15}{#1}}
\newcommand{\secondone}[1]{\colorbox{blue!15}{#1}}

\mathchardef\mhyphen="2D
\begin{document}

\title{LLDiffusion: Learning Degradation Representations in Diffusion Models for Low-Light Image Enhancement
}

\author{Tao Wang         \and
        Kaihao Zhang        \and
        Ziqian Shao        \and
        Wenhan Luo  \and \\
        Bjorn Stenger  \and  
        Tong Lu   \and
        Tae-Kyun Kim  \and
        Wei Liu \and
        Hongdong Li 
}

\institute{
         Tao Wang, Ziqian Shao and Tong Lu  \at
         Nanjing University, Nanjing, 210023, China \\
         \email{taowangzj@gmail.com, ziqian.shao@outlook.com, lutong@nju.edu.cn}        \\
         \and
        Kaihao Zhang and Hongdong Li \at
        Australian National University, Australia\\
        \email{\{super.khzhang, hongdong.li\}@gmail.com}        \\
        \and
        Wenhan Luo \at
        Shenzhen Campus of Sun Yat-sen University, Shenzhen, 518107, China\\
        \email{whluo.china@gmail.com} \\
        \and
         Bjorn Stenger \at Rakuten Institute of Technology, Japan \\
           \email{bjorn@cantab.net} \\
        \and
        Tae-Kyun Kim \at Imperial College London, London, UK \& KAIST, Daejeon, South Korea\\
           \email{tk.kim@imperial.ac.uk} \\  \and
        Wei Liu \at Tencent, Shenzhen, 518107, China \\
           \email{wl2223@columbia.edu} \\
}

\date{Received: date / Accepted: date}
% The correct dates will be entered by the editor

\maketitle

\begin{abstract}
Current deep learning methods for low-light image enhancement (LLIE) typically rely on pixel-wise mapping learned from paired data. However, these methods often overlook the importance of considering degradation representations, which can lead to sub-optimal outcomes. In this paper, we address this limitation by proposing a degradation-aware learning scheme for LLIE using diffusion models, which effectively integrates degradation and image priors into the diffusion process, resulting in improved image enhancement. Our proposed degradation-aware learning scheme is based on the understanding that degradation representations play a crucial role in accurately modeling and capturing the specific degradation patterns present in low-light images. To this end, First, a joint learning framework for both image generation and image enhancement is presented to learn the degradation representations. Second, to leverage the learned degradation representations, we develop a Low-Light Diffusion model (LLDiffusion) with a well-designed dynamic diffusion module. This module takes into account both the color map and the latent degradation representations to guide the diffusion process. By incorporating these conditioning factors, the proposed LLDiffusion can effectively enhance low-light images, considering both the inherent degradation patterns and the desired color fidelity. Finally, we evaluate our proposed method on several well-known benchmark datasets, including synthetic and real-world unpaired datasets. Extensive experiments on public benchmarks demonstrate that our LLDiffusion outperforms state-of-the-art LLIE methods both quantitatively and qualitatively. 
The source code and pre-trained models are available at https://github.com/TaoWangzj/LLDiffusion.

\end{abstract}

\section{Introduction}\label{section:Introduction}

Images captured under low-light conditions often contain a variety of degradation factors, such as low contrast, low visibility, and high noise, which can negatively impact downstream vision tasks and can lead to failures in many computer vision systems. As a result, low-light image enhancement (LLIE) has become an important area of research in the generation of high-quality normal-exposure images from low-light images.

%%%%% use a paragraph to review the LLIE Methods 
During the early development of LLIE, methods focused on using image priors to improve visibility by adjusting the image contrast. 
Some of these methods include histogram equalization-based techniques~\cite{kim1997contrast,stark2000adaptive}, as well as retinex-based techniques~\cite{kimmel2003variational,wang2014variational}. However, approaches were limited in their applicability and could lead to undesirable outcomes, such as color distortion or over-enhancement. In recent years, deep learning techniques have emerged as a more successful approach to LLIE, with studies showing that they can achieve superior performance~\cite{wei2018deep,zamir2020learning,wang2022ultra}.

\begin{figure}[t]
	\begin{center}
		\begin{tabular}[t]{c} \hspace{-4mm} 
			\includegraphics[width=0.45\textwidth]{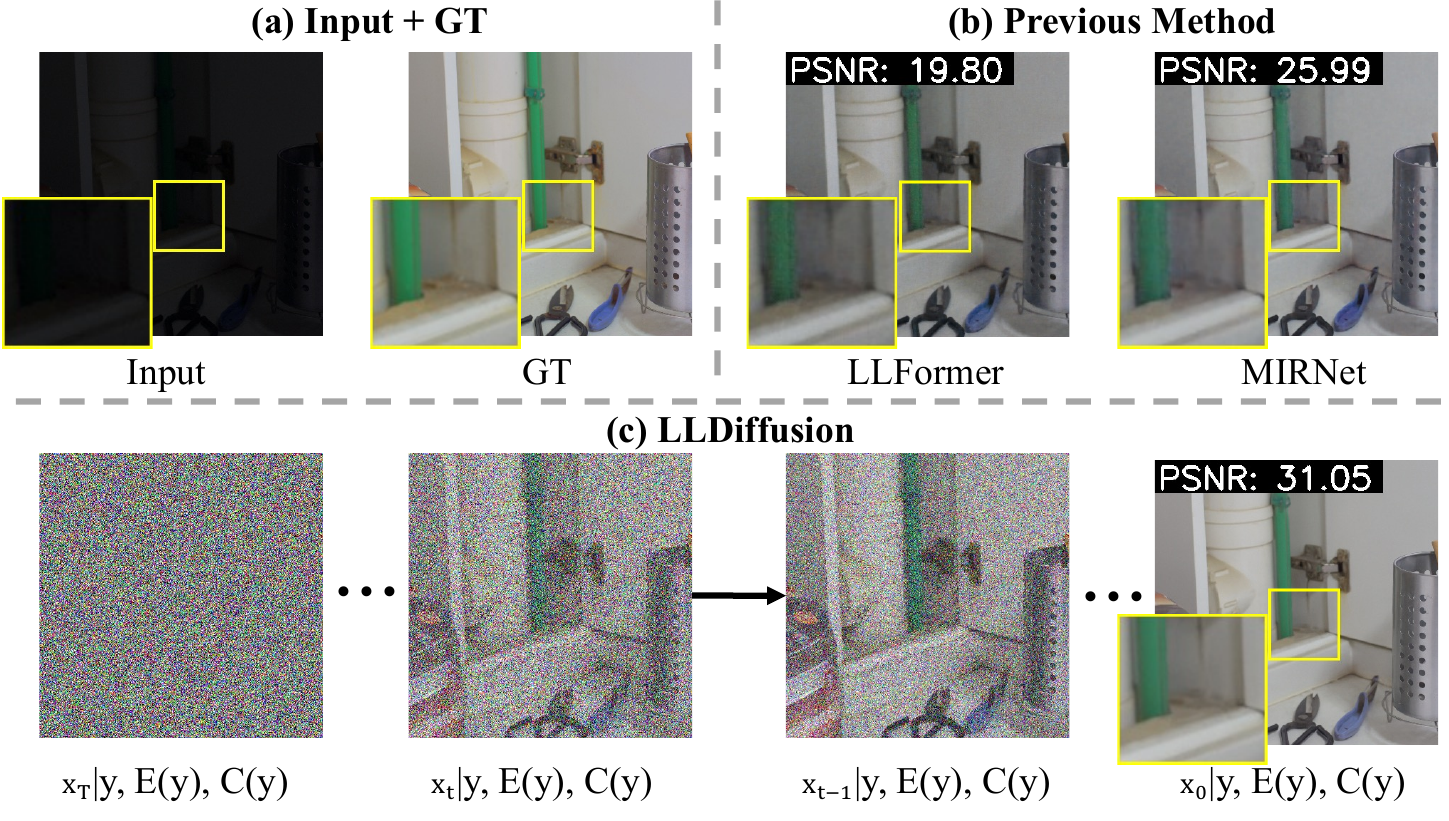}
		\end{tabular}
	\end{center}
	% \vspace{-4mm}
	\caption{(a) Input low-light image and ground truth, (b) results of two recent SOTA methods LLFormer~\cite{wang2022ultra} and MIRNet~\cite{zamir2020learning}, and (c) our LLDiffusion iteratively (T$\rightarrow$0) enhances the low-light image, where the $x_{0}$ is the final enhanced result.}
	\label{fig:results}
	% \vspace{-7mm}
\end{figure}
Learning-based LLIE methods typically consider the enhancement task as a pixel-wise mapping between low-light and normal-light images, and employ pixel-wise losses, such as $\mathbf{L}{1}$ and $\mathbf{L}{2}$ losses, during training. However, these methods often struggle to maintain visual fidelity, leading to unsatisfactory outcomes~\cite{bruna2016super,hao2022decoupled}. To address this issue, some LLIE methods use generative adversarial networks (GANs) and introduce an adversarial loss into the training process to improve visual fidelity~\cite{jiang2021enlightengan,liu2021pd}. However, GAN-based methods are susceptible to unstable training and mode collapse, which can produce artifacts in the enhanced images~\cite{mescheder2018training}. 

In more recent developments, diffusion models, including the diffusion denoising probability model ~\cite{ho2020denoising}, have shown impressive results in image restoration tasks. These models possess a superior capability to model distributions compared to other deep generative models like GANs, VAE, and Normalizing flow. However, to date, diffusion models have not been studied for low-light image enhancement, and current deep learning-based LLIE methods and diffusion models do not fully incorporate degradation representations in their network design.

%%%%%% use a paragraph to introduce the proposed LLDiffusion 
To address these issues, we propose a new degradation-aware learning scheme, which explicitly considers degradation representations from low-light images in the enhancement stage. Based on this scheme, we propose a low-light diffusion model called LLDiffusion for low-light image enhancement. 
To model degradation representations, we first propose estimating explicit low-light degradation representations through a joint learning method of low-light generation learning and image enhancement learning. Then, based on the pre-learned degradation, we propose a dynamic degradation-aware diffusion module (DDDM) that explicitly integrates degradation and image prior into the diffusion model for image enhancement. Specifically, LLDiffusion mainly contains a latent map encoder, a degradation generation network (DGNET), and a dynamic degradation-aware diffusion module (DDDM). DGNET is designed to learn the degradation process from normal-light images to low-light ones. The DDDM is proposed to achieve image enhancement. These three components are first trained together to guide the encoder to learn how to accurately encode degradation representations from the low-light input. After obtaining the pre-trained encoder, we freeze the encoder and retrain the DDDM to effectively perform image enhancement conditioned on the color map (denoting image prior), the degradation representations, and the input low-light image. Experimental results show that LLDiffusion significantly outperforms state-of-the-art LLIE methods (see Fig.~\ref{fig:results}).

%%%%%% summary of the contribution of this work
In summary, the contributions of this work are four-fold:
\begin{itemize}
\item To the best of our knowledge, we are the first to incorporate degradation representations into the diffusion model for addressing low-light image enhancement. This novel approach allows us to effectively capture and utilize degradation patterns in the enhancement process.

\item We propose a degradation-aware learning scheme specifically designed for low-light image enhancement. Building on this scheme, we develop a novel and unified diffusion model-based framework called LLDiffusion, which leverages the benefits of both low-light image generation and image enhancement tasks.

\item We propose a novel dynamic degradation-aware diffusion module in the diffusion network that explicitly integrates degradation and image priors to enhance image quality. Additionally, to assess the generation capability of our proposed model, we collect a new real-world testing dataset called the Real World Test (RWT) for evaluating real-world low-light image enhancement.

\item Our proposed method is evaluated on various benchmark datasets, including synthetic datasets such as LOL~\cite{wei2018deep}, LOL-v2~\cite{yang2021sparse}, and VE-LOL~\cite{liu2021benchmarking}, as well as real-world unpaired datasets like DICM~\cite{lee2012contrast}, MEF~\cite{ma2015perceptual}, NPE~\cite{wang2013naturalness}, and our RWT dataset. Extensive experiments on these benchmarks show that our method outperforms state-of-the-art LLIE methods both quantitatively and qualitatively.
\end{itemize}

The subsequent sections are structured as follows: Section~\ref{sec:related_work} provides an overview of related works pertaining to low-light image enhancement. 
In Section~\ref{sec:method}, we present the details of our novel approach, outlining the architecture, methodology, and key components employed to address the existing challenges in low-light image enhancement. Then, the experimental results are reported and analyzed in Section~\ref{sec:experiment}. Section~\ref{sec:limitations} discusses limitations and future work. Finally, Section~\ref{sec:conclusion} summarizes the conclusion of this paper.

%%%%%%% two days to write related work 
%------------------------------------------------------------------------
\section{Related Work}\label{sec:related_work}
% \subsection{Image Restoration in Adverse Weather Conditions}
Our proposed method is related to the LLIE task, diffusion models, and learning degradation representations in the model, which are reviewed in the following.
\subsection{Low-light Image Enhancement}
As an important research topic in recent decades, many LLIE approaches have been proposed to enhance low-light images. LLIE methods can be roughly categorized into two families, \ie non-learning-based methods and learning-based methods.

%%%%%% no-learning-based methods 
Non-learning-based methods primarily rely on histogram equalization, gamma correction, and retinex theory. Histogram-based and gamma correction-based methods~\cite{kim1997contrast,reza2004realization,huang2012efficient,lee2013contrast,srinivas2020low} expand the dynamic range of pixel intensities and increase image contrast. While these methods are simple and fast, they do not consider the spatial information of images, which can lead to undesirable artifacts like over- or under-enhancement in the restored image. Retinex-based methods~\cite{fu2016weighted,li2018structure,zhang2018high,hao2020low} are based on retinex theory~\cite{land1977retinex}, which proposes that an image can be decomposed into an illumination map and a reflectance map, where the reflectance map describes the inherent properties of the object that remain unchanged. Retinex-based methods aim to correct the illumination of the image. However, they often fail to correct color distortion and may even introduce additional color distortion in the recovered images, especially when dealing with complex degraded images.~\cite{wang2019underexposed,hao2022decoupled}.

%%%%%% learning-based methods
Recently, there have been several proposed learning-based methods for enhancing low-light images. LLNet~\cite{lore2017llnet} adaptively improves image contrast using a depth auto-encoder/decoder model. Retinex-Net~\cite{wei2018deep} uses two CNN-based networks for image decomposition and enhancement, based on the retinex model. Instead of directly learning an image-to-image mapping, Wang~\etal proposed a method to estimate the illumination map for enhancing under-exposed images~\cite{wang2019underexposed}. Yang~\etal designed a semi-supervised approach to learn linear-band feature representations from low-light images~\cite{yang2020fidelity}. Additionally, other methods have adopted unsupervised learning or zero-shot learning paradigms. For example, EnlightenGAN~\cite{jiang2021enlightengan} is an unsupervised generative adversarial model that aims to use global and local information extracted from unpaired data to enhance real-world images. Zero-DCE~\cite{guo2020zero} is a zero-shot learning-based method that treats enhancement as a task of image-specific curve estimation using a deep network. Recent LLIE methods have introduced new techniques such as normalizing flow or transformer networks into the model. For example, in~\cite{wang2022low}, a normalizing flow model is applied to deal with low-light image enhancement. 
Wang \etal \cite{wang2022ultra} proposed a transformer-based network called LLFormer to address the problem of ultra-high-definition low-light image enhancement and achieved state-of-the-art performance.

\subsection{Diffusion-based Model for Low-level Tasks}

In recent years, diffusion-based generative models~\cite{sohl2015deep} have achieved remarkable results with improved denoising diffusion probabilistic models~\cite{ho2020denoising,nichol2021improved}. These diffusion models can be divided into a fixed forward diffusion process and a learnable reverse diffusion process. The forward process gradually adds Gaussian noise to the image, whereas the reverse process uses a Markov chain structure to denoise and restore the clean image. Thanks to the fixed forward diffusion process and the learnable reverse diffusion process employed in diffusion models, these models have demonstrated state-of-the-art performance in various low-level tasks, including image super-resolution~\cite{ho2022cascaded,saharia2022image}, image-to-image translation~\cite{choi2021ilvr,lugmayr2022repaint}, and image deblurring~\cite{whang2022deblurring,ren2022image}. For example, \cite{saharia2022image} proposed using a conditional diffusion model for single image super-resolution, which yielded better performance compared to state-of-the-art GAN-based and normalizing flow-based models. Additionally, \cite{ozdenizci2022restoring} proposed a patch-based diffusion model for size-agnostic image restoration, which uses a guided denoising process across overlapping patches during inference.

To improve generalization ability on unseen domains, a multi-scale domain-generalizable guidance network~\cite{ren2022image} was embedded into the diffusion model for image deblurring. \cite{guo2022shadowdiffusion} proposed the ShadowDiffusion model for shadow removal, which is an unrolling diffusion framework that aims to explicitly integrate a degradation prior and a diffusive generative prior. Although these diffusion models outperform GAN-based methods, their performance is still limited due to ignoring explicit degradation modeling. In contrast, we propose a diffusion model that can adaptively learn degradation representations to help enhance images under low-light conditions. 

\subsection{Learning Degradation Representations in Low-level Vision}
In real-world scenarios, low-level vision methods are commonly employed to tackle a wide range of complex degradations present in images. Degradation representations have been widely acknowledged as a vital component in various low-level image restoration tasks, such as image super-resolution and image denoising. The degradation representations play a pivotal role in capturing and modeling the underlying degradation patterns, allowing for effective restoration and enhancement of images affected by diverse forms of degradation.  In the task of image super-resolution, several approaches~\cite{zhang2018learning,shocher2018zero,bell2019blind,zhang2020deep} employed degradation representations to address the challenge of image super-resolution in the real world. The fundamental concept behind these methods is to train the model using various types of degradations, including different combinations of Gaussian blur, motion blur, and noise. During inference, these methods assume that the degradations are known for the low-resolution test images. Notably, these methods exhibit improved results in super-resolving images when the degradation is known in advance.
In the task of image denoising, several approaches~\cite{mildenhall2018burst,guo2019toward,zhang2020deblurring} incorporated the noise variance as an input to the network, enabling adaptive handling of different noise strengths. Moreover, various practical denoising methods address the issue of noise variance by considering factors such as the characteristics of the Poisson-Gaussian distribution~\cite{li2022efficient} and different ISO settings~\cite{wang2020practical}. However, in the task of low-light image enhancement, the utilization of degradation representations in network design has been overlooked, limiting the performance of the network, particularly in real-world scenarios. To address this limitation, we propose a degradation-aware diffusion model that incorporates degradation representations into the enhancement process, aiming to improve the overall enhancement performance.

%------------------------------------------------------------------------
\section{Methodology}\label{sec:method}
In this section, we first highlight some limitations of the previous scheme for LLIE and introduce our proposed degradation-aware learning scheme. Then, a joint learning framework for both image enhancement and low-light generation to learn the degradation representations from low-light inputs is presented. Finally, a dynamic degradation-aware diffusion model of our proposed method is illustrated, which helps effectively integrate the latent degradation representations for LLIE.

\begin{figure*}[t]
\begin{center}
	\includegraphics[width=\textwidth]{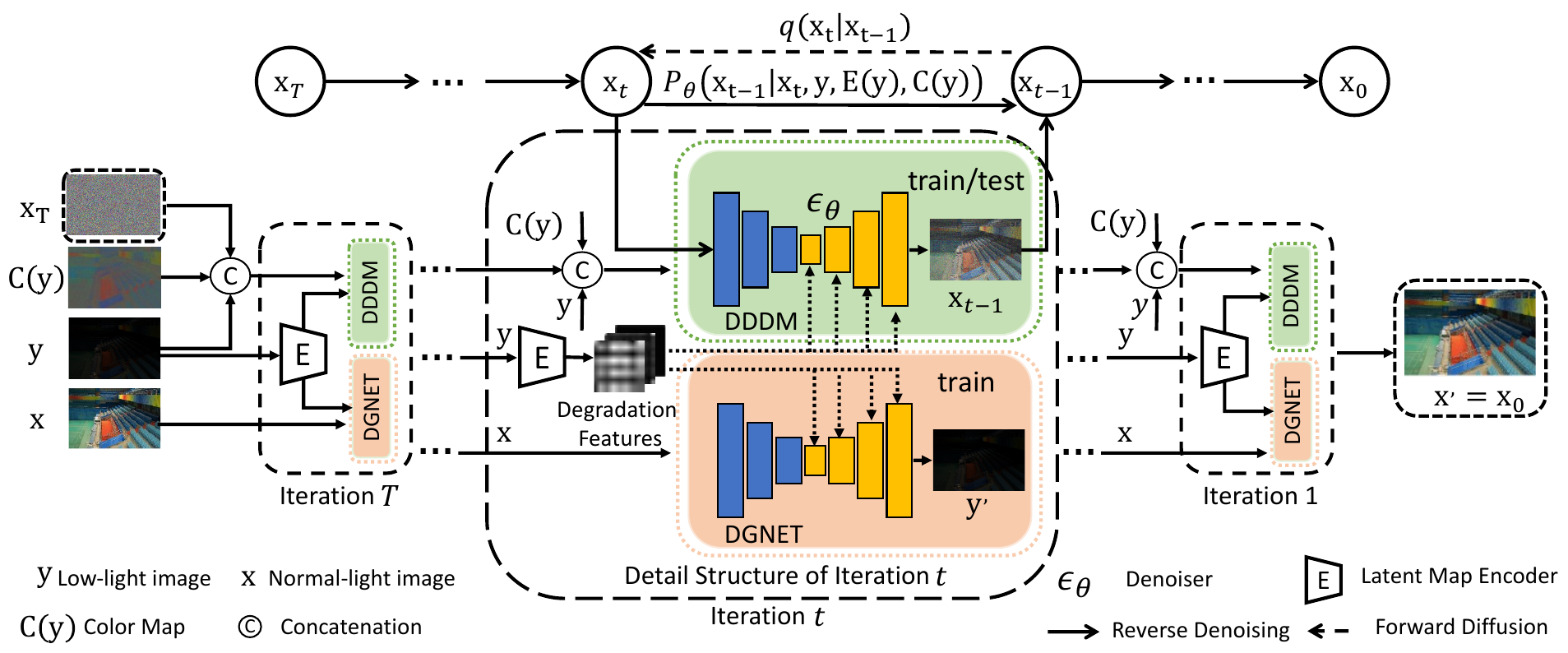}
	  % \vspace{-0.1in}
   % \vspace{-4mm}
	\caption{An overview of the forward diffusion (dashed line) and reverse denoising (solid line) processes for our LLDiffusion. Our LLDiffusion model consists of a degradation generation network (DGNET), a dynamic degradation-aware diffusion module (DDDM), and a latent map encoder (E). Our LLDiffusion model first learns the low-light degradation representations with the joint task of image enhancement using DDDM, and low-light generation using DGNET. DDDM learns to effectively perform image enhancement
conditioned on the color map, $\mathrm{c(y)}$, the degradation representations, and the input low-light image.}
	\label{fig:overall}
 \end{center}
   % \vspace{-5mm}
\end{figure*}

\subsection{Degradation-aware Learning Scheme} 
Recent low-light image enhancement methods can be divided into two families according to the imaging model under low-light conditions.  

\textbf{Retinex-based methods}: Inspired by retinex theory, a low-light image $\mathbf{y}$ can be decomposed into two components, the reflectance layer $\mathbf{R}$ and the illumination layer $\mathbf{I}$, which is formulated as:
\begin{equation}
\small
\mathbf{y} = \mathbf{R} \circ \mathbf{I},
\end{equation}
where $\circ$ refers to element-wise multiplication. The reflectance $\mathbf{R}$ describes a fixed object-inherent property. In this scheme, the enhancement process is transformed into estimating the illumination layer $\mathbf{I}$, or joint learning features from the reflectance and illumination to generate normal-light images~\cite{wei2018deep,wang2019underexposed,zhang2019kindling}.  

\textbf{End-to-end learning methods}: The low-light image $\mathbf{y}$ is the result of the normal-light image $\mathbf{x}$ being corrupted using a degradation function  $\theta$~\cite{jiang2022degrade}, which can include noise, exposure changes, or other artifacts:
\begin{equation}
\small
\mathbf{y}=\theta(\mathbf{x}).
\end{equation}
The goal of low-light image enhancement is to generate a normal exposure image $\mathbf{x}$ given a low-light image $\mathbf{y}$.
In this scheme, a large number of methods~\cite{jiang2021enlightengan,zamir2020learning,Zamir2022MIRNetv2,wang2022ultra} use deep networks to learn the latent inverse function $\theta^{-1}$ from training image pairs. 

Although these schemes are elegant, they have certain limitations. The retinex-based scheme focuses only on adjusting the illumination layer to enhance images, failing to generate high-fidelity and natural results. End-to-end learning methods are heavily based on the network architecture and training strategy. End-to-end models ignore the intrinsic degradation representations of the low-light image, which often results in artifacts or unnatural enhanced images. To this end, we propose a new scheme called \textbf{degradation-aware learning scheme} for low-light image enhancement, which considers the degradation between the normal-light image $\mathbf{x}$ and the low-light image $\mathbf{y}$. The low-light degradation model is formulated as:
\begin{equation}
\small
\mathbf{y}=\psi\left(\mathbf{x}, \mathbf{x}_{D}(\mathbf{x};\mathbf{y})\right),
\end{equation}
where $\mathbf{x}_{D}(\mathbf{x};\mathbf{y})$ denotes the pre-learned degradation from the normal-light image $\mathbf{x}$
to the low-light image $\mathbf{y}$, and $\psi$ refers to the low-light degradation model. Since there is no explicit analytic function to represent the degradation of low-light images, we formulate this complex mapping as a high-dimensional non-analytic transfer map. Thus, given the low-light image $\mathbf{y}$ and degradation representations $\mathbf{x}_{D}(\mathbf{x};\mathbf{y})$, the enhancement process is formulated as:
\begin{equation}
\small
\mathbf{x}=\psi^{-1}\left(\mathbf{y}, \mathbf{x}_{D}(\mathbf{x};\mathbf{y})\right),
\end{equation}
where $\psi^{-1}$ denotes the latent function of the model's enhancement process. 

\subsection{LLDiffusion}
Based on the proposed degradation-aware learning scheme, we propose a new model called LLDiffusion for low-light image enhancement, which is shown in Fig.~\ref{fig:overall}. LLDiffusion consists of two primary stages: a joint learning stage for both image enhancement and low-light generation to learn the degradation representations from low-light inputs and a dynamic degradation-aware diffusion module that integrates the pre-learned degradation representations for effectively low-light image enhancement. We introduce the details of the two stages in the following.

\textbf{Learning degradation representations through joint learning}. The degradation representation $\mathbf{x}_{D}$ is crucial for low-light image enhancement according to the degradation-aware learning scheme, and inaccurate degradation representations will directly affect the performance of the model~\cite{jiang2021enlightengan}. Thus, we treat the degradation estimation as an auxiliary task to help produce high-quality images with normal light. Inspired by previous work~\cite{zhang2020deblurring,li2022learning}, we jointly learn low-light image generation and image enhancement to encode the low-light degradation representations from low-light images. Specifically, as illustrated in Fig.~\ref{fig:overall}, we first design a latent map encoder $\mathbf{E}$ to transfer the low-light image $\mathbf{y}$ into the degradation representations $\mathbf{E(y)}$. A degradation generation network (DGNET) simulates the degradation process from $\mathbf{x}$ to $\mathbf{y'}$. A dynamic degradation-aware diffusion module (DDDM) generates the enhanced image $\mathbf{x'}$. The encoder $\mathbf{E}$ extracts multi-scale features with four convolution layers and uses upsampling layers to match the feature size of DGNET and DDDM's decoders. The degradation representations are integrated into each decoder layer of DGNET and DDDM via feature concatenation. DGNET and DDDM are jointly trained in a framework to help the encoder learn degradation representations that cover as many low-light degradation cases as possible.

\textbf{(1) Learning low-light image degradation}. Given the normal-light image $\mathbf{x}$ and the degradation representations $\mathbf{E(y)}$ from the latent map encoder $\mathbf{E}$, the proposed DGNET $\mathbf{D}_{g}$ learns how to synthesize the low-light image:
% \begin{small}
% \begin{align}
\begin{equation}
\small
\mathbf{y'}=\mathbf{D}_{g}(\mathbf{x}, \mathbf{E(y)}),
\end{equation}
% \end{align}
% \end{small}
where $\mathbf{y'}$ represents the synthesized low-light image given the normal light image $\mathbf{x}$ and degradation representations $\mathbf{E(y)}$. Thanks to the degradation representations, DGNET is aware of the complex degradation patterns in low-light conditions and effectively learns how to degrade images. We use the same U-Net network as in the DDDM diffusion module to build DGNET, which is trained using an $\mathcal{L}_{1}$ loss, denoted as $\mathcal{L}_{1}(y, y')$. 

\textbf{(2) Learning image enhancement}. A dynamic degradation-aware diffusion module (DDDM) is trained jointly with low-light image generation. Benefiting from the learned degradation representations, the DDDM function $\mathbf{D}_{\mathrm{diff}}$ can handle more complex degradation patterns. The diffusion process is formulated as:
% \begin{small}
% \begin{align}
\begin{equation}
\small
\mathbf{x'}=\mathbf{D}_{\mathrm{diff}}(\mathbf{y}, \mathbf{E(y)}),
% \end{align}
% \end{small}
\end{equation}
where $\mathbf{x'}$ refers to the enhanced image. $\mathbf{D}_{\mathrm{diff}}$ is trained by minimizing the denoising loss $\mathcal{L}_{\mathrm{diff}}$. During the joint training stage, the overall loss function is represented as:
% \begin{small}
% \begin{align}
\begin{equation}
\small
\mathcal{L}_{\mathrm{total}}=\mathcal{L}_{\mathrm{diff}} +\alpha \cdot \mathcal{L}_{1},
% \end{align}
% \end{small}
\end{equation}
where $\mathcal{L}_{1}$ is used in DGNET for low-light generation learning and $\mathcal{L}_{\mathrm{diff}}$ is used in DDDM for image enhancement learning. 
We use hyper-parameters $\alpha$ in the final loss $\mathcal{L}_{\mathrm{total}}$ for DGNET and DDDM.  

\begin{algorithm}[t]
\hspace*{\algorithmicindent}\noindent \textbf{Input:} low-light image $\mathbf{y}$, normal-light image $\mathbf{x}$, degradation $\mathbf{E(y)}$, and invariant color maps $\mathbf{C(y)}$. 
\caption{Dynamic degradation-aware diffusion training}\label{algo_train}
\begin{algorithmic}[1]
        \While{not converged}
        \State $t \sim \text{Uniform}\{1, \ldots, T\}$
        \State $\epsilon \sim \mathcal{N}(\mathbf{0},\mathbf{I})$
        \State $\mathbf{e}_t = \epsilon_\theta\left(\sqrt{\bar{\alpha}_t} \mathbf{x}_0+\sqrt{1-\bar{\alpha}_t} \epsilon,  \mathbf{y}, \mathbf{E(y)}, \mathbf{C(y)}, t\right)$
      \If{Join Learning} 
      \State Perform Gradient descent steps on $\nabla_\theta 
        \;\mathcal{L}_{\mathrm{total}}(\theta)$ 
     \Else
      \State Perform Gradient descent steps on $\nabla_\theta 
        \;\mathcal{L}_{\mathrm{diff}}(\theta)$ 
     \EndIf
        \EndWhile
      \\ $\textbf{return}\;\; \theta$ 
\end{algorithmic}
  % \vspace{-1mm}
\end{algorithm}

\textbf{Enhancement with dynamic degradation-aware diffusion}.
After obtaining the pre-trained encoder, we freeze the pre-trained encoder to explore how the diffusion model of embedded learning degradation representations can be used for the image enhancement task. The core idea of previous conditional diffusion-based methods for image restoration~\cite{saharia2022palette,saharia2022image,ozdenizci2022restoring} is learning a conditional reverse process $\mathbf{p}_\theta(\mathbf{x}_{0: T}|y)$ without modifying the diffusion process $\mathbf{q}(\mathbf{x}_{1: T}|x)$ for $\mathbf{x}$. However, these conditional diffusion-based methods are typically implemented via concatenation of $\mathbf{y}$ and $\mathbf{x}_{t}$ at the input level, which easily leads to poor performance in complex cases. To this end, we propose a dynamic degradation-aware diffusion module (DDDM) with a more efficient modulation mechanism that progressively enhances the image using degradation representations (from the encoder) and invariant color maps~\cite{wang2022low} as conditions. The invariant color maps, which are inspired by the Retinex theory, serve as an image prior to the low-light image enhancement task. It enriches the saturation and reduces color distortion, leading to improved image quality. We adopt the color map as a condition in our diffusion model to guide the low-light image restoration process and leverage its benefits in enhancing color and brightness consistency. In the following,
we present the training and sampling states of DDDM.

\begin{algorithm}[t!]
\hspace*{\algorithmicindent}\noindent \textbf{Input:} low-light image $\mathbf{y}$, degradation representations $\mathbf{E(y)}$, invariant color maps $\mathbf{C(y)}$, and number of implicit sampling steps $T$.
\caption{Degradation-based diffusive sampling}\label{algo}
\begin{algorithmic}[1]
      \State $\mathbf{x_{t}} \sim \mathcal{N}(\mathbf{0}, \mathbf{I})$
      \For{$t =T, \ldots, 1$}
        \State $\mathbf{e}_{t-1} = \epsilon_\theta\left(\mathbf{x}_{t}, \mathbf{y}, \mathbf{E(y)}, \mathbf{C(y)}\right)$
        \State $\mathbf{x}_{t-1} \!\!= \!\!\sqrt{\bar{\alpha}_{t-{1}}}\left(\frac{\mathbf{z}_{t}-\sqrt{1-\bar{\alpha}_t} \cdot \mathbf{e}_{t-1}}{\sqrt{\bar{\alpha}_t}}\right)
       \! +\!\sqrt{1-\bar{\alpha}_{t-{1}}} \cdot \mathbf{e}_{t-1}$
      \EndFor
      \State \Return $\mathbf{x}_t$
\end{algorithmic}
  % \vspace{-1mm}
\end{algorithm}

\quad \textbf{(1) Diffusion training}:
In the training stage, we sample $(\mathbf{x}_0, \mathbf{y}) \sim q(\mathbf{x}, \mathbf{y})$ from a paired data distribution (\eg a low-light image $\mathbf{y}$ and its corresponding normal-light image $\mathbf{x}$ ). The low-light image $\mathbf{y}$ is used to generate the degradation $\mathbf{E(y)}$ and the invariant color maps $\mathbf{C(y)}$. The training method is illustrated in Algorithm~\ref{algo_train}. It learns the dynamic degradation-aware reverse process:
% \begin{small}
% \begin{align}
\begin{equation}
\small
\setlength\abovedisplayskip{2pt}%
\setlength\belowdisplayskip{2pt}
p_\theta\left(\mathbf{x}_{0: T} \!\mid \!\mathbf{y}\right)=p\left(\mathbf{x}_T\right) \prod_{t=1}^T p_\theta\left(\mathbf{x}_{t-1} \!\mid\! \mathbf{x}_t, \mathbf{y}, \mathbf{E(y)},\mathbf{C(y)} \right).
% \end{align}
% \end{small}
\end{equation}

\begin{figure*}[t]
\begin{center}
	\includegraphics[width=\textwidth]{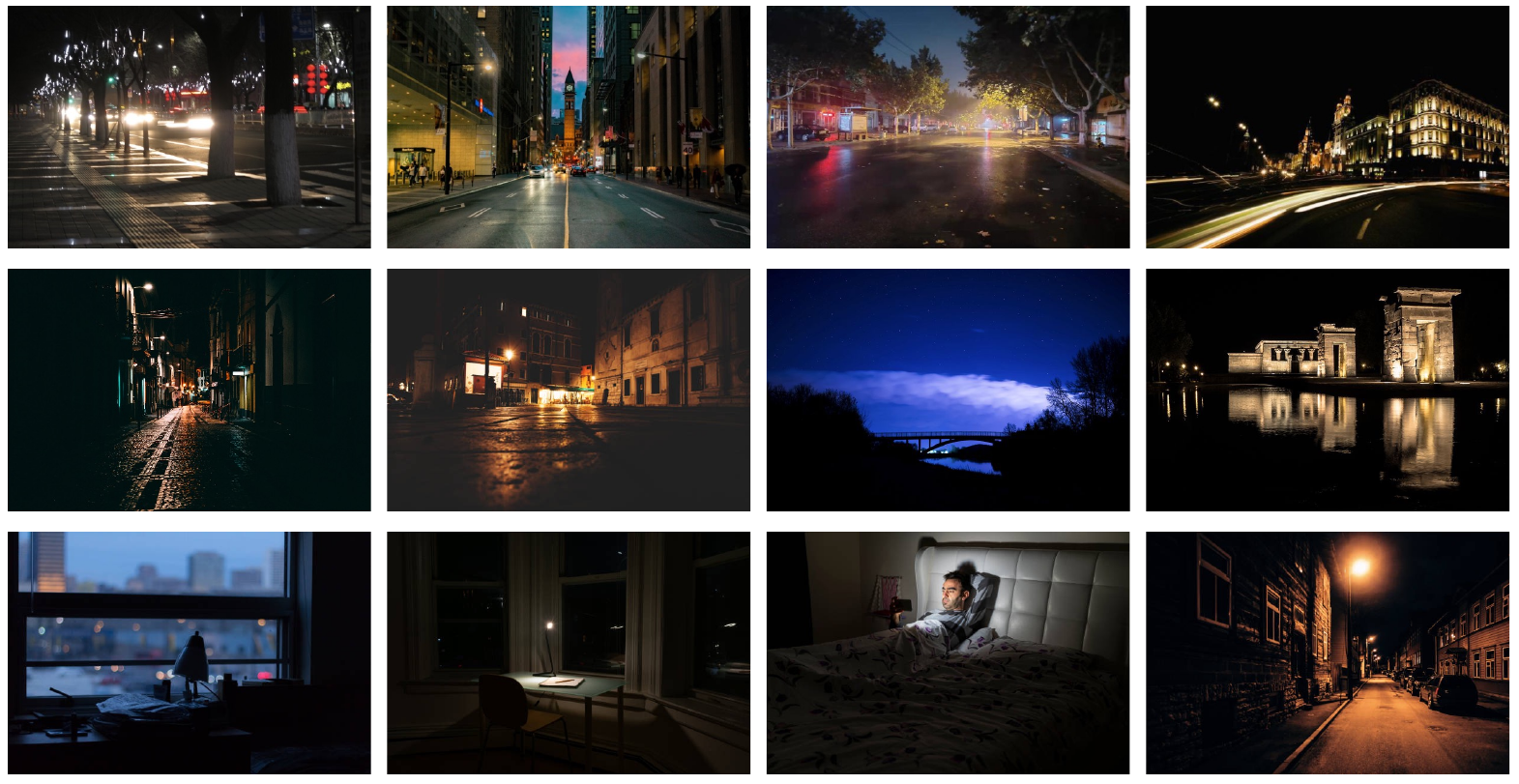}
	  % \vspace{-0.1in}
   % \vspace{-4mm}
	\caption{Some images sampled from our RWT dataset. The images in our RWT dataset are collected from a wide range of environments, including buildings, streets, people, and natural landscapes. The diversity of scenes within our dataset aims to encompass the broad spectrum of real-world scenarios encountered in low-light conditions, enabling comprehensive evaluations of low-light image enhancement methods.}
	\label{fig:rwt_dataset}
 \end{center}
   % \vspace{-5mm}
\end{figure*}

Specifically, a forward Gaussian diffusion process is to sample an intermediate noisy version $\mathbf{x}_t$ from the normal-light image $\mathbf{x}_{0}$ at a diffusion step $t$ via $\mathbf{x}_t = \sqrt{\bar{\alpha}_t} \mathbf{x}_0+\sqrt{1-\bar{\alpha}_t} \epsilon$, $\epsilon \sim \mathcal{N}(\mathbf{0},\mathbf{I})$, $\alpha_t = 1-\beta_t$, where $\beta_t$ refers to the noise schedule. A denoiser $\epsilon_\theta$ takes the low-light image $\mathbf{y}$, the intermediate noise map $\mathbf{x}_t$, the degradation representations $\mathbf{E(y)}$, the color maps $\mathbf{C(y)}$, and the time step $t$ as inputs to estimate the noise map $\mathbf{e}_t$, which is represented as:

\begin{equation}
\small
\setlength\abovedisplayskip{2pt}%
\setlength\belowdisplayskip{2pt}
    \mathbf{e}_t = \epsilon_\theta\left(\sqrt{\bar{\alpha}_t} \mathbf{x}_0+\sqrt{1-\bar{\alpha}_t} \epsilon,  \mathbf{y}, \mathbf{E(y)}, \mathbf{C(y)}, t\right).
% \end{align}
% \end{small}
\end{equation}

We adopt the denoising loss to optimize the proposed DDDM following the previous work~\cite{ho2020denoising}, which is presented as:
\begin{small}
\begin{align}
    \mathcal{L}_{\mathrm{diff}}=\mathbb{E}_{\mathbf{x}_0, t, \epsilon}\left\|\mathbf{e}_t - \epsilon\right\|^2_F.
\end{align}
\end{small}

\quad \textbf{(2) Diffusion Sampling for Enhancement}: DDDM generates the enhanced image $\mathbf{x}_{0}$ from random noise $\mathbf{x}_{T}$ by iteratively sampling $\mathbf{x}_{t-1}$ from $p_\theta(\mathbf{x}_{t-1}|\mathbf{x}_{t}, \mathbf{y},\mathbf{E(y)}, \mathbf{C(y)})$.  The sampling process is shown in Algorithm~\ref{algo}, in which we incorporate the degradation representations and the image prior to the sampling without additional cost. The proposed DDDM with pre-learned degradation and color maps learns to handle complicated degradation patterns and maintains the color and brightness consistency of the enhanced images.

\section{Experiments and Analysis}\label{sec:experiment} 
In this section, we introduce the experimental setup and evaluate the performance of the proposed LLDiffusion model on various benchmarks. Finally, we conduct comprehensive ablation studies to evaluate the contributions of the different model components.

\subsection{Datasets and Implementation Details}
% Datasets and Implementation Details

We evaluate our proposed model on popular benchmarks including LOL~\cite{wei2018deep}, LOL-V2~\cite{yang2021sparse}, VE-LOL~\cite{liu2021benchmarking}, and four real-world datasets for LLIE.

\textbf{LOL}. LOL dataset~\cite{wei2018deep} includes $485$ and $15$ paired images for training and testing, respectively. The image size of LOL images is $600 \times 400$, where each pair of images consists of a low-light image and its corresponding normal brightness ground truth.

\textbf{LOL-v2}. LOL-v2 dataset~\cite{yang2021sparse} consists of two different subsets, \ie LOL-v2-real and LOL-v2-synthetic. The LOL-v2-real subset is captured in real scenes by changing the ISO and exposure time and includes $689$ and $100$ paired images for training and testing. In the LOL-v2-synthetic subset, low-light images are synthesized from RAW images by analyzing the illumination distribution of low-light images. There are $1,000$ low-/normal image pairs, and $900$ pairs are chosen for training and $100$ pairs for testing. 

\textbf{VE-LOL}. VE-LOL~\cite{liu2021benchmarking} is a large-scale dataset for low-light image enhancement, which provides $2,500$ low-light/normal-light paired images with more diverse scenes. Unlike LOL, VE-LOL additionally considers noise modeling at the RAW image level, and the noise is captured by using four different cameras: Sony A7R, Olympus E-M10, Sony RX100 IV, and Huawei Nexus 6P.

\def \rootrealworld {image/LOL/111/}
\begin{figure*}[t]
    \centering
    \subfigure[Input]{
        \includegraphics[width=0.190\linewidth]{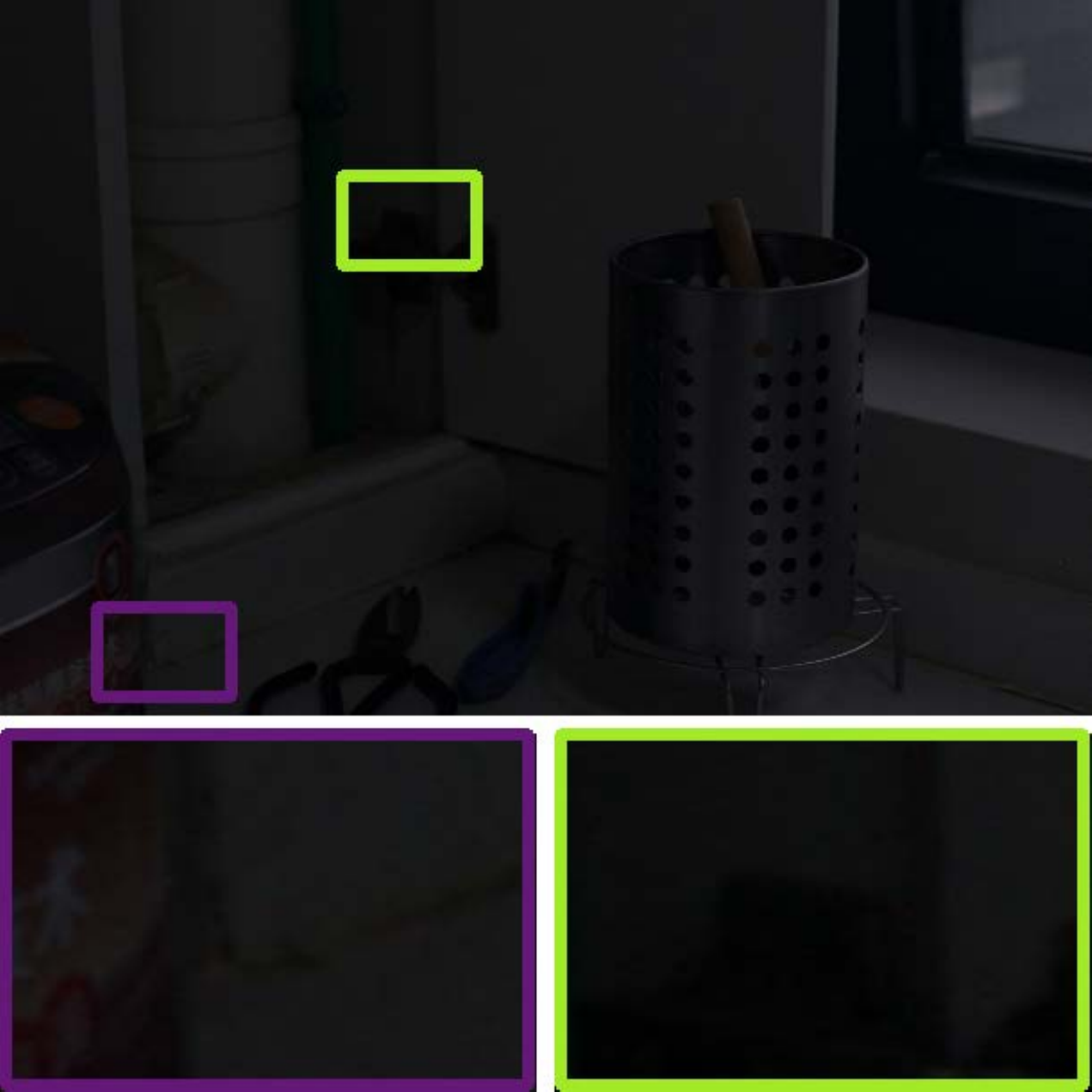}
    }\hspace{-5pt}
    \subfigure[RetinexNet]{
        \includegraphics[width=0.190\linewidth]{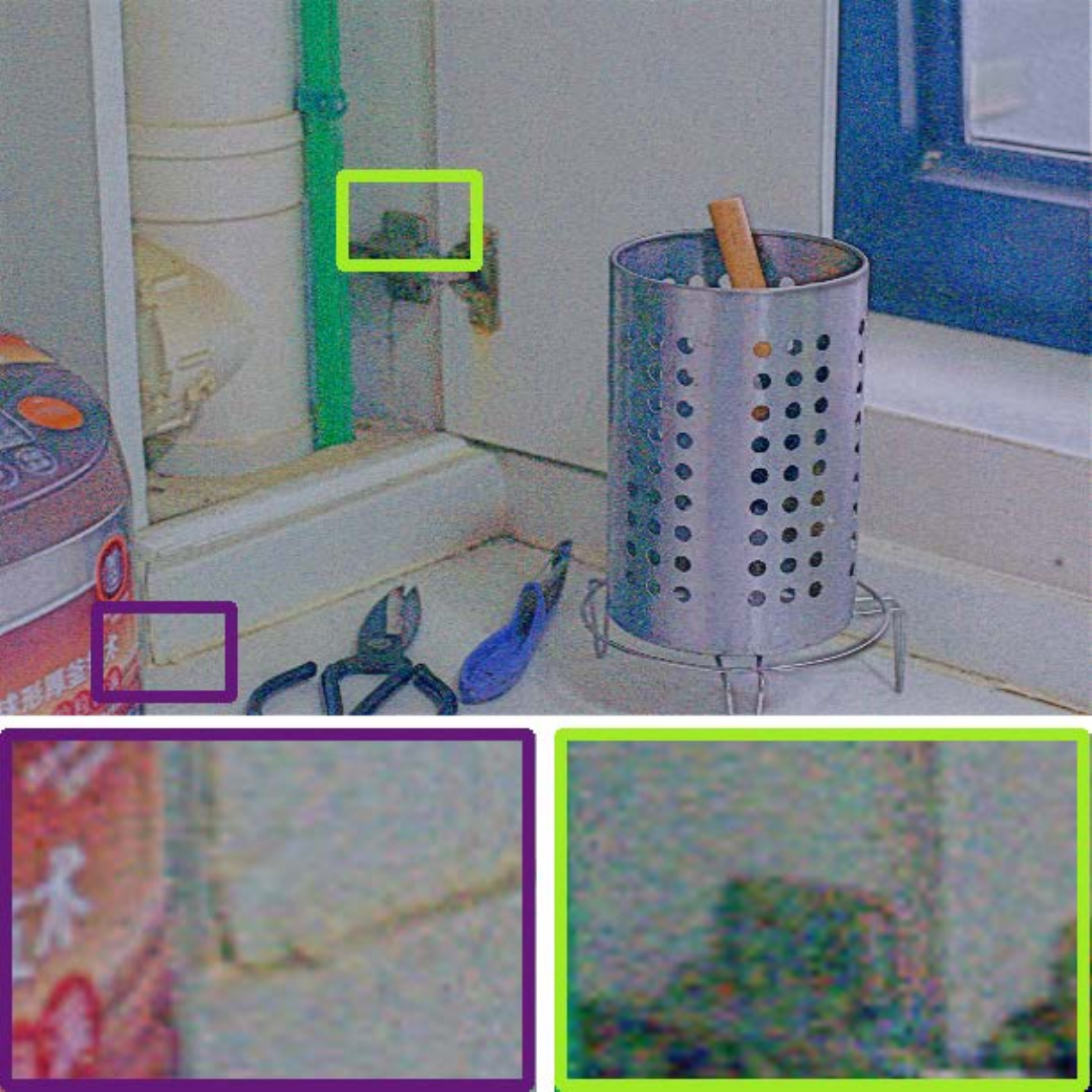}
    }\hspace{-5pt}
    \subfigure[KinD]{ 
        \includegraphics[width=0.190\linewidth]{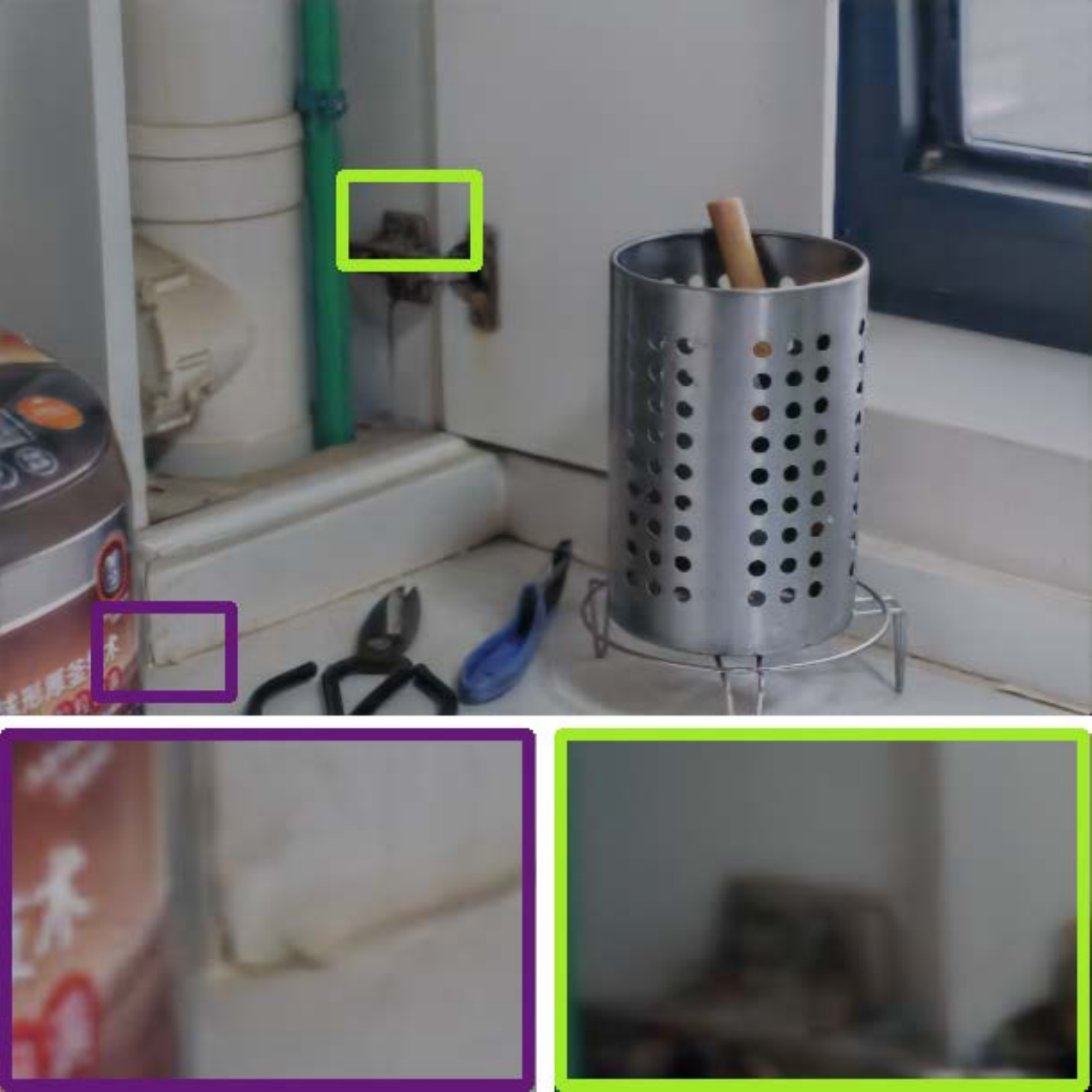}
    }\hspace{-5pt}
    \subfigure[KinD++]{
        \includegraphics[width=0.190\linewidth]{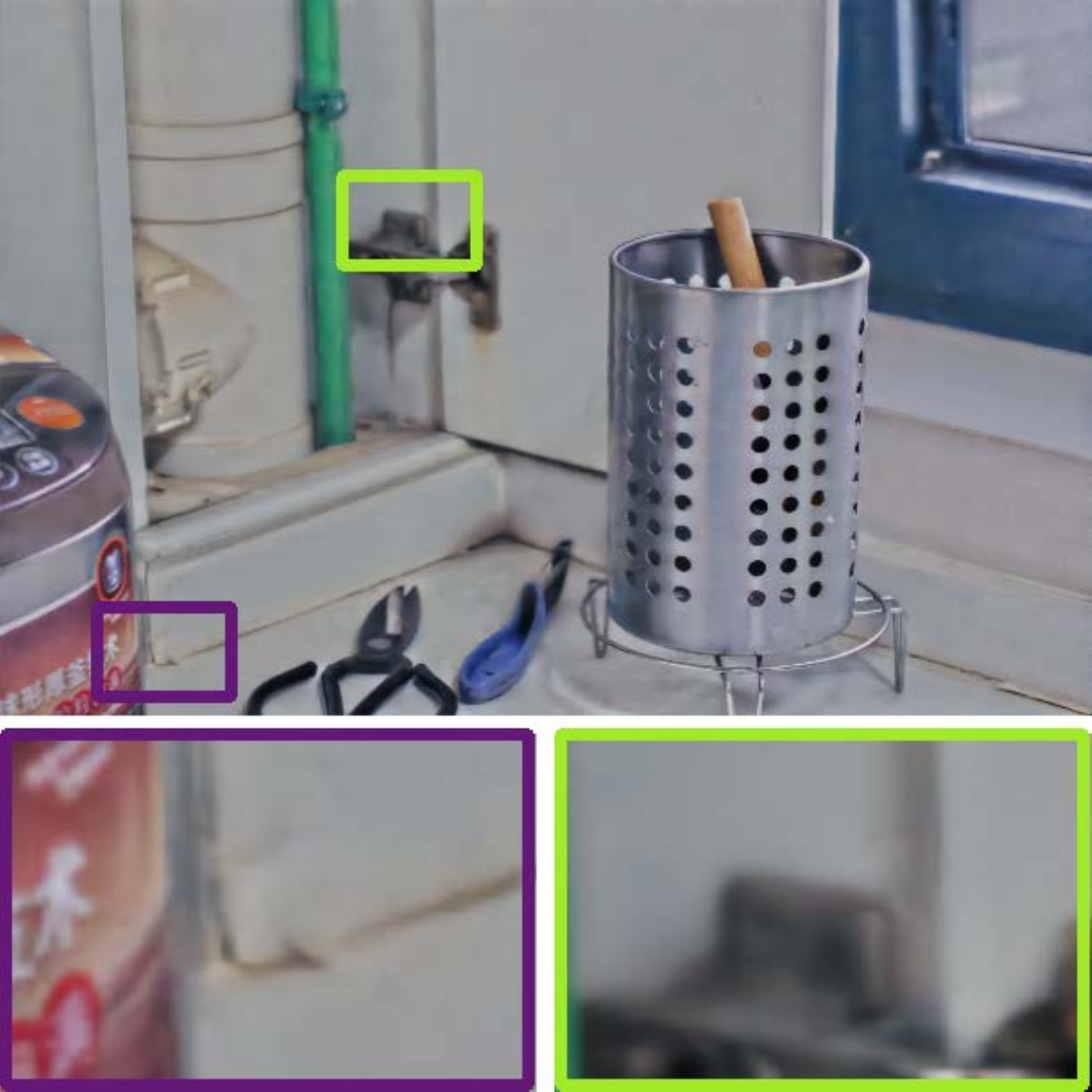}
    }\hspace{-5pt}
    \subfigure[UFormer]{
        \includegraphics[width=0.190\linewidth]{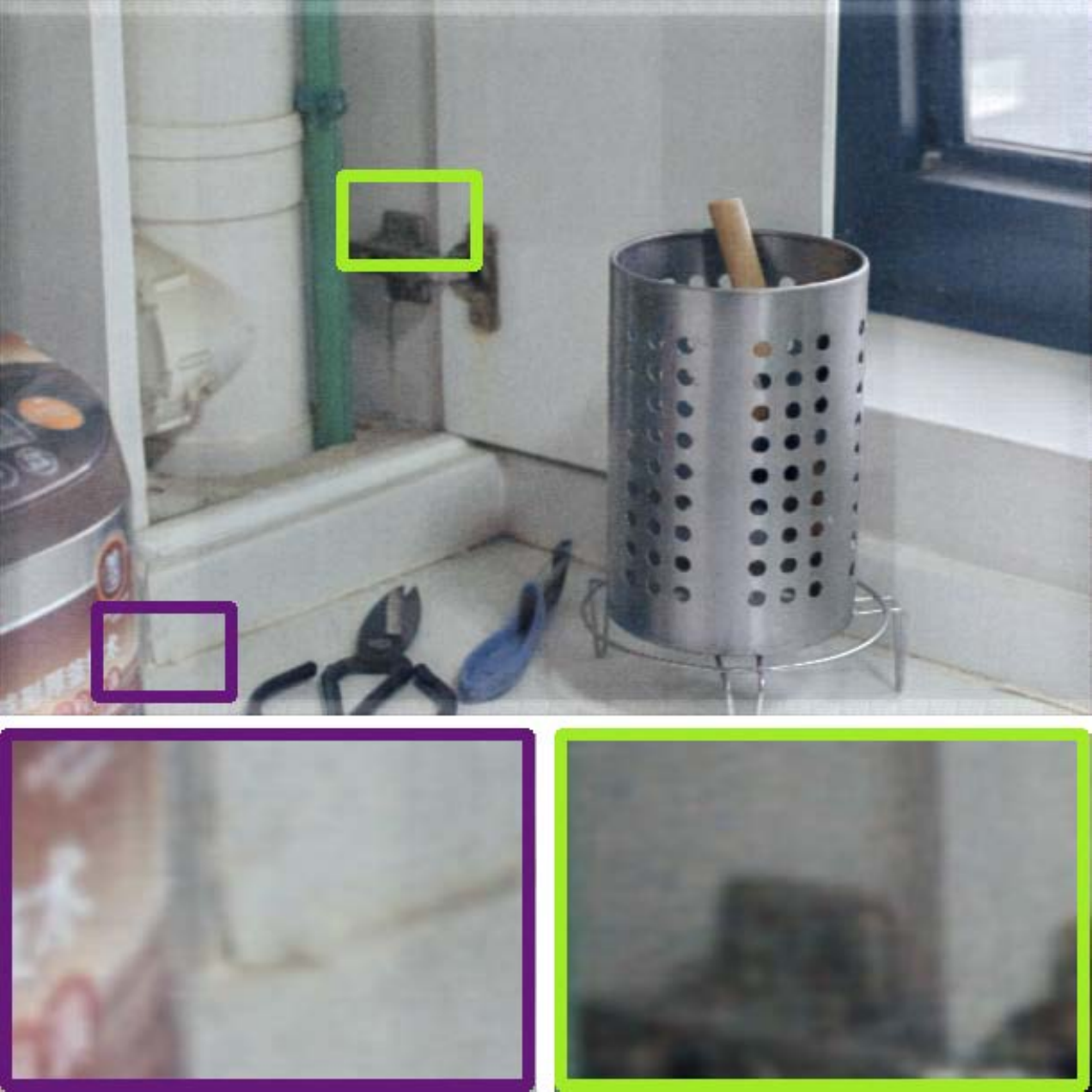}
    } \vspace{-1.0mm}
    \subfigure[Restormer]{
        \includegraphics[width=0.190\linewidth]{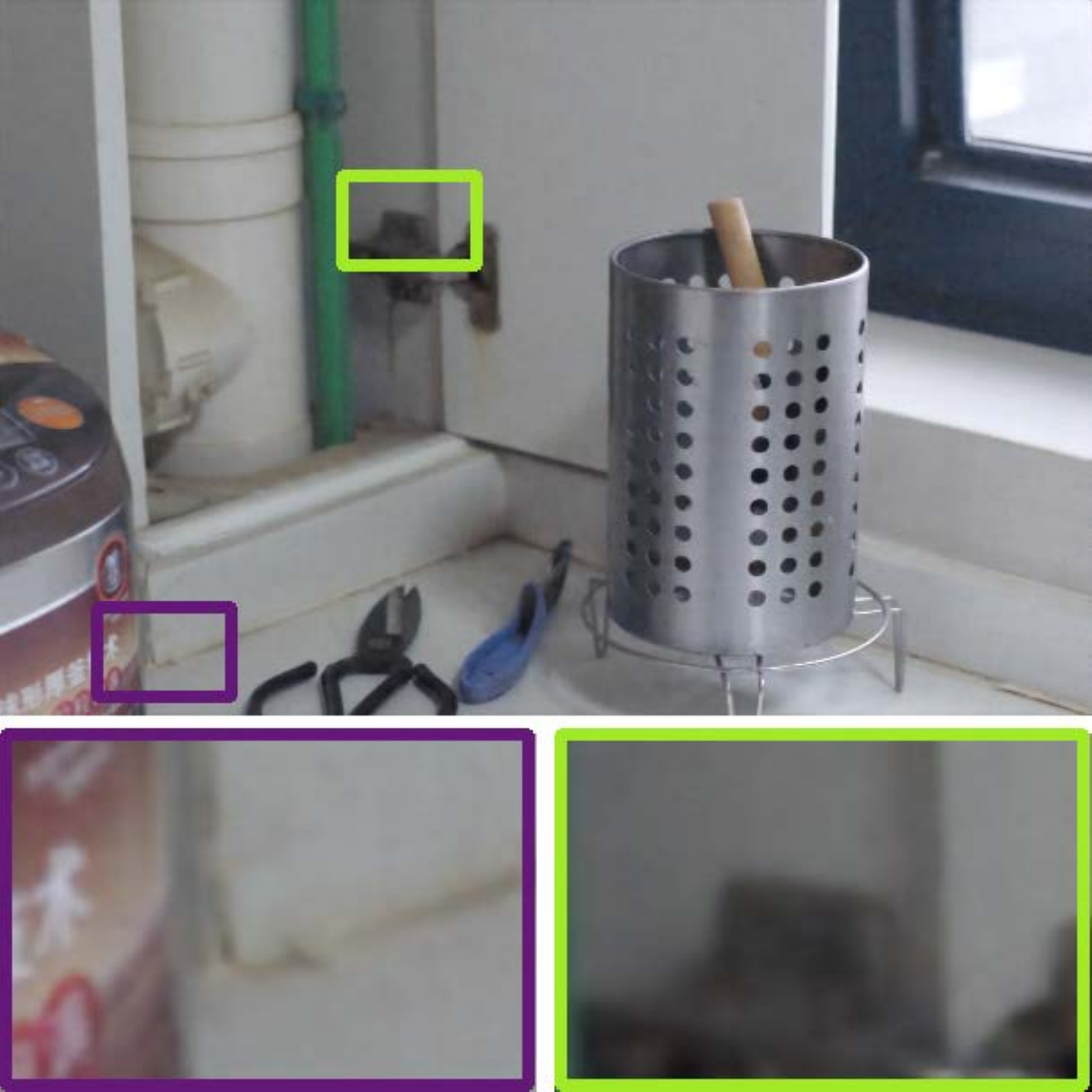}
    } \hspace{-5pt}
    \subfigure[LLFormer]{
        \includegraphics[width=0.190\linewidth]{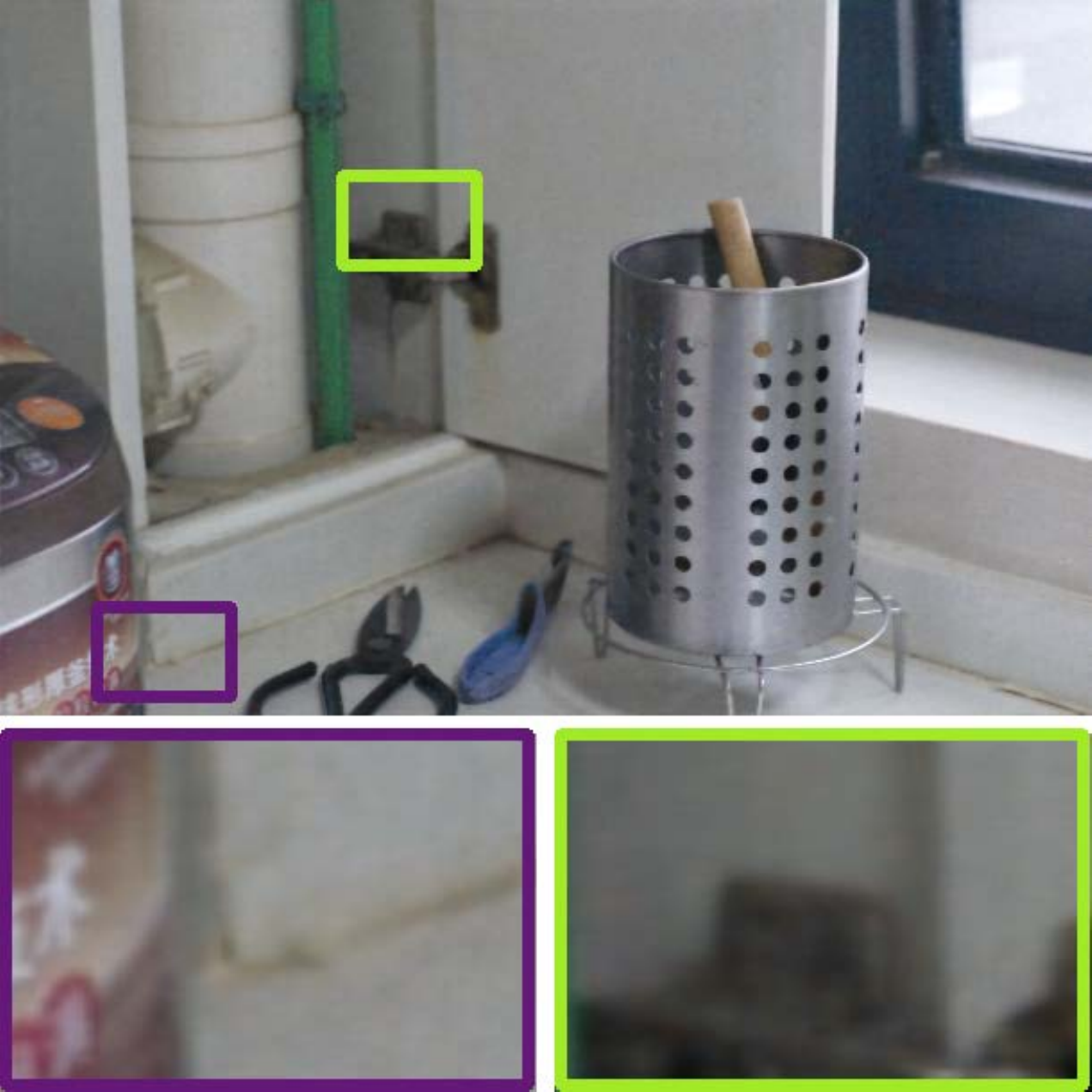}
    }\hspace{-5pt}
    \subfigure[MIRNet]{
        \includegraphics[width=0.190\linewidth]{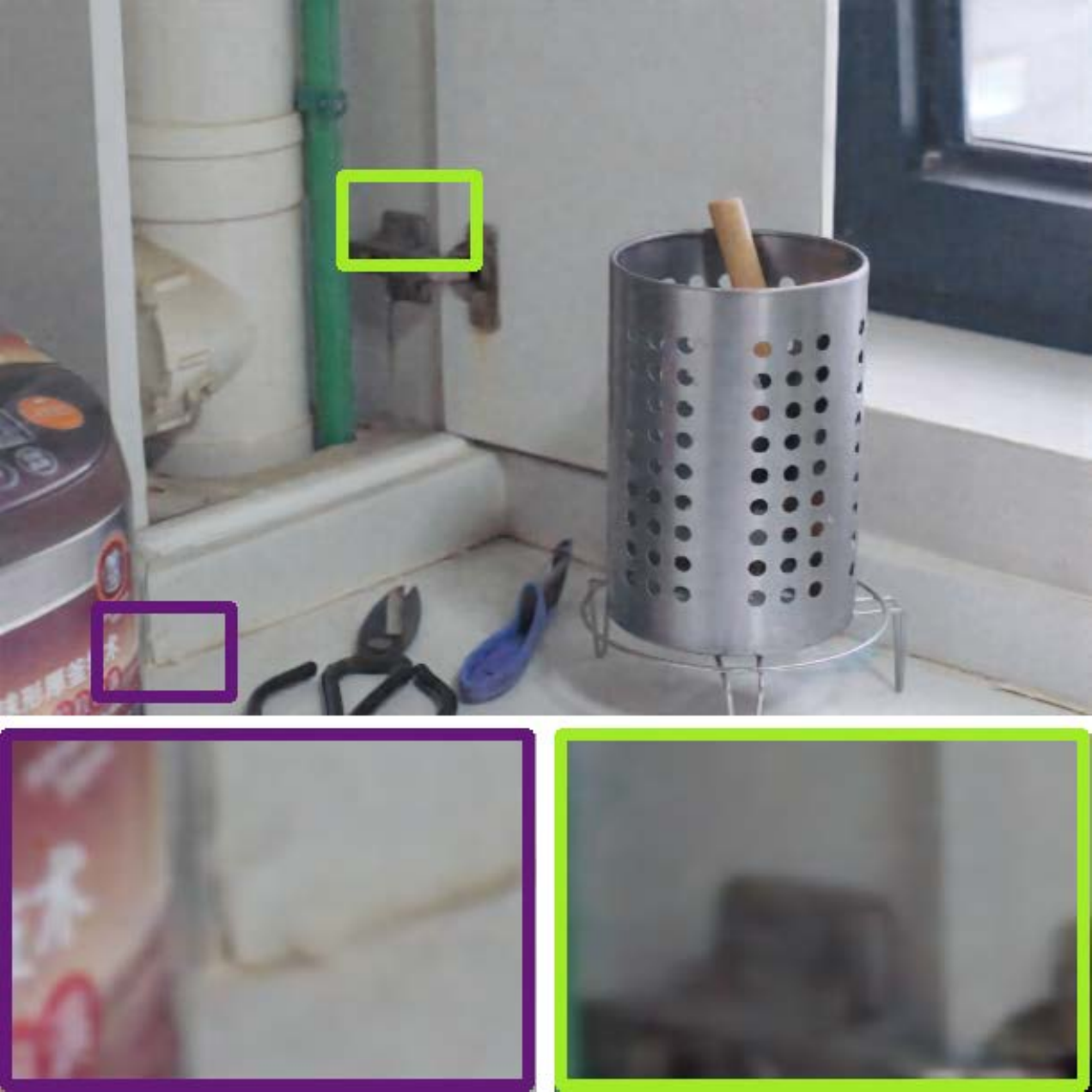}
    }\hspace{-5pt}
    \subfigure[LLDiffusion]{
        \includegraphics[width=0.190\linewidth]{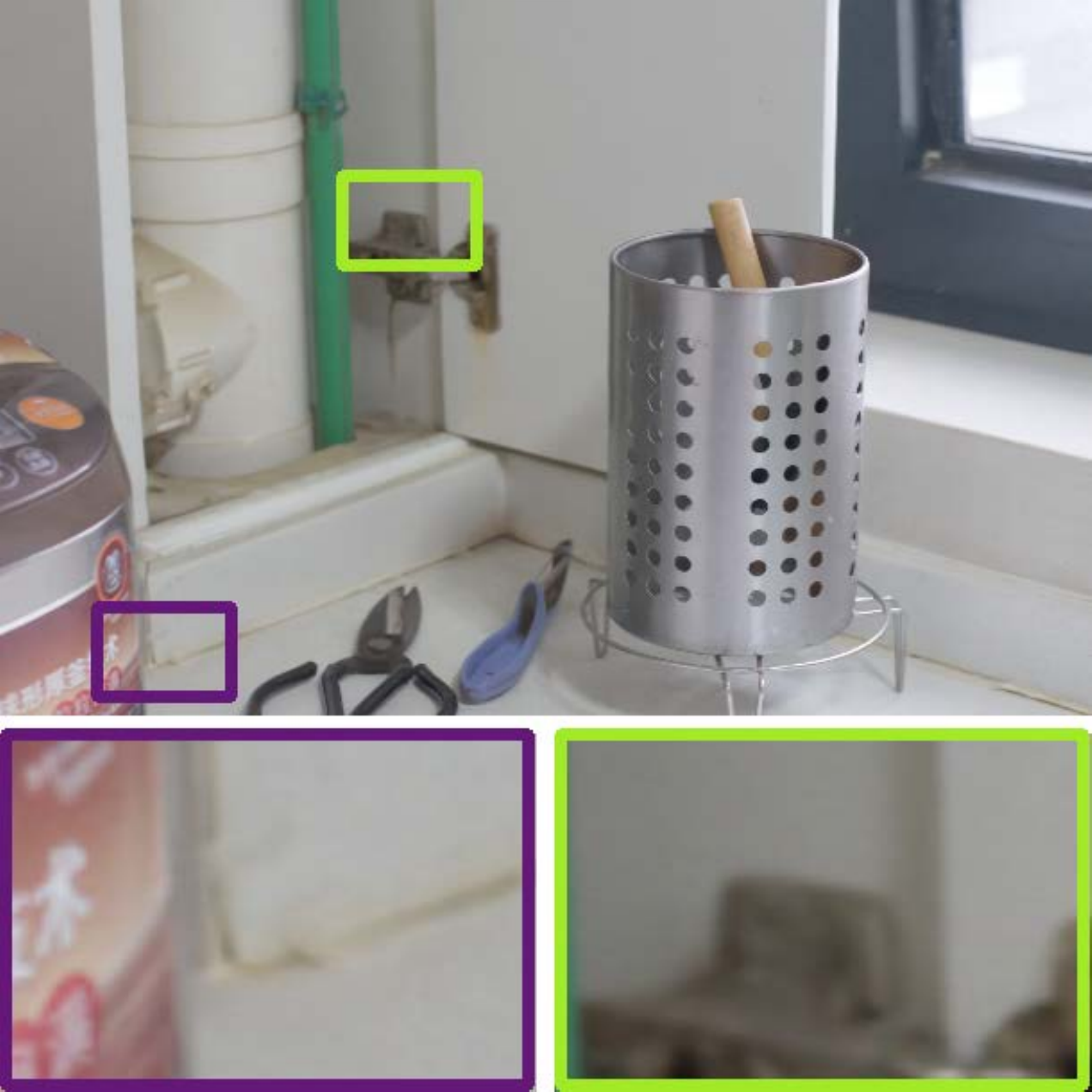}
    }\hspace{-5pt}
    \subfigure[GT]{
        \includegraphics[width=0.190\linewidth]{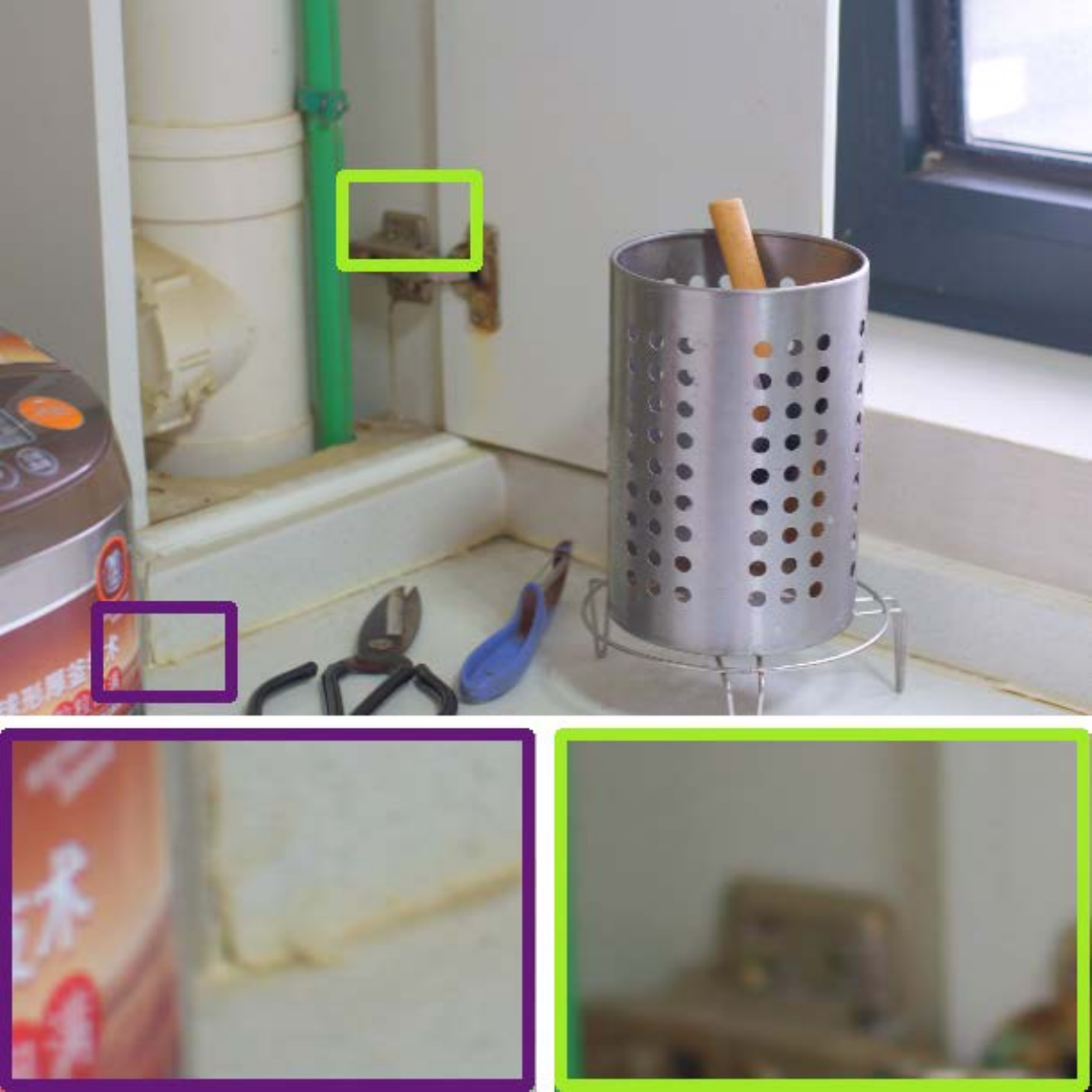}
    }
    % \vspace{-4mm}
    \caption{Qualitative results of low-light image enhancement methods on LOL~\cite{wei2018deep}. For each method, the output image is shown along with two zoomed-in regions below. The normal-light exposed images generated by our LLDiffusion exhibit less noise or artifacts, and higher color fidelity.}% \textcolor{blue}{HL: pls add some discussions here.}}
    \label{fig:lol-result}
    % \vspace{-4mm}
    % \vspace{-0.3cm}
\end{figure*}

\begin{table}[t]
    \centering
    \caption{Comparison results on the LOL dataset~\cite{wei2018deep} in terms of PSNR, SSIM and LPIPS.
    We highlight the \firstone{best} and the \secondone{second-best} values for each metric.
    }
    % \vspace{-4mm}
    \scalebox{0.95}{
    \begin{tabular}{cccc}
    \toprule
      Method  &  PSNR $\uparrow$ & SSIM $\uparrow$ & LPIPS $\downarrow$ \\
    \midrule
       BIMEF~\cite{ying2017bio}  & 13.88 & 0.595 & 0.326 \\ 
       NPE~\cite{wang2013naturalness} & 16.97 &  0.484  &  0.405 \\ 
       LIME~\cite{guo2016lime} & 16.76 & 0.445 & 0.395 \\
       MF~\cite{fu2016fusion}  & 16.97 & 0.508 &  0.380 \\
       SRIE~\cite{fu2016weighted}  & 11.86 & 0.495 & 0.257 \\
       RetinexNet~\cite{wei2018deep} & 16.77 & 0.560 & 0.474 \\
       KinD~\cite{zhang2019kindling} & 20.87 & 0.790 & 0.170 \\
       KinD++~\cite{zhang2021beyond} & 21.30 & 0.820 & 0.160 \\  
       Zero-DCE~\cite{guo2020zero} & 14.86 & 0.562 & 0.335\\ 
       RUAS~\cite{liu2021ruas} & 18.23 & 0.720 & 0.350 \\
       EnlightenGAN~\cite{jiang2021enlightengan} & 17.48 & 0.652 & 0.322 \\

        Uformer~\cite{wang2022uformer} & 18.55 & 0.721 & 0.321 \\
        Restormer~\cite{zamir2022restormer} & 22.37 & 0.816 & 0.141 \\
        LLFormer~\cite{wang2022ultra} & 23.65 & 0.816 & 0.171 \\
        MIRNet~\cite{zamir2020learning} & \secondone{24.14} & \secondone{0.830} & \secondone{0.131} \\
     LLDiffusion & \firstone{24.65} & \firstone{0.843} & \firstone{0.075} \\ 
    \bottomrule
    \end{tabular}
    }
    \label{tab:lol}
    % \vspace{-0.6cm}
     % \vspace{-4mm}
\end{table}

\def \rootrealworld {image/LOL-v2-Syn/17/}
\begin{figure*}[th]
    \centering
    \subfigure[Input]{
        \includegraphics[width=0.190\linewidth]{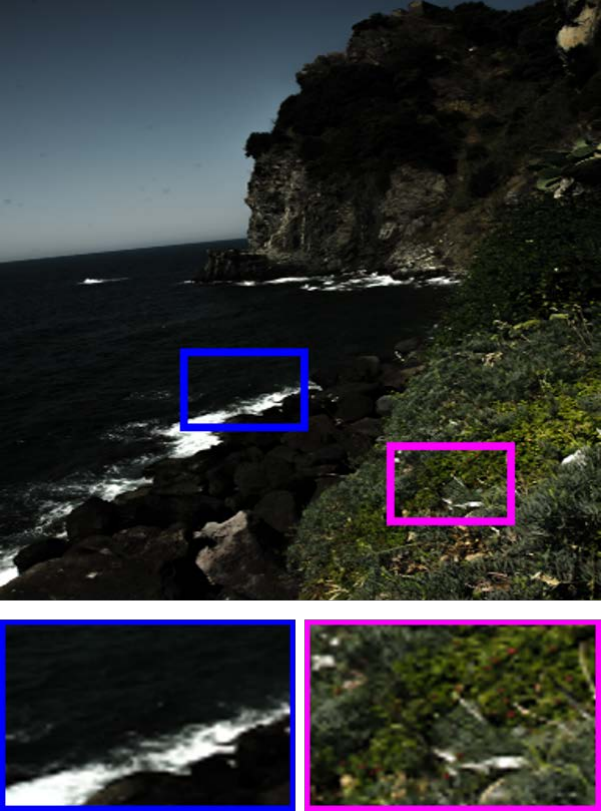}
    }\hspace{-5pt}
    \subfigure[RetinexNet]{
        \includegraphics[width=0.190\linewidth]{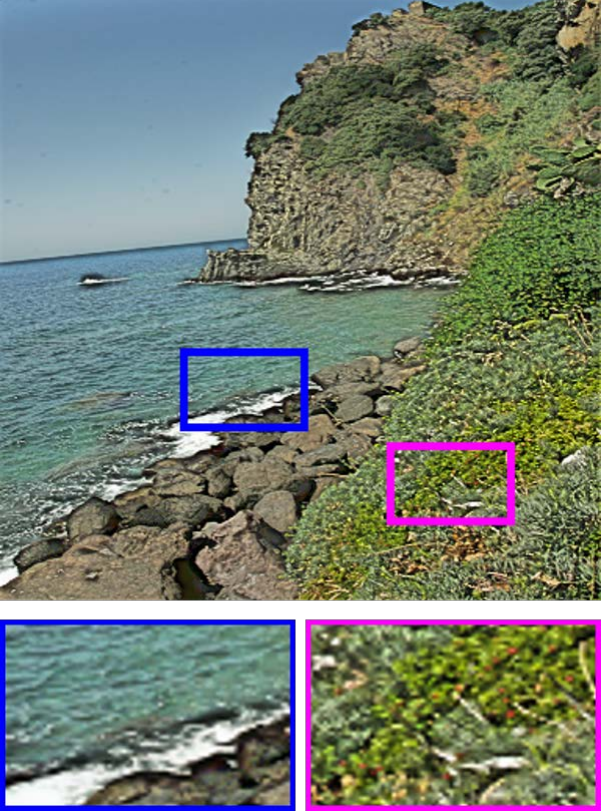}
    }\hspace{-5pt}
    \subfigure[KinD]{ 
        \includegraphics[width=0.190\linewidth]{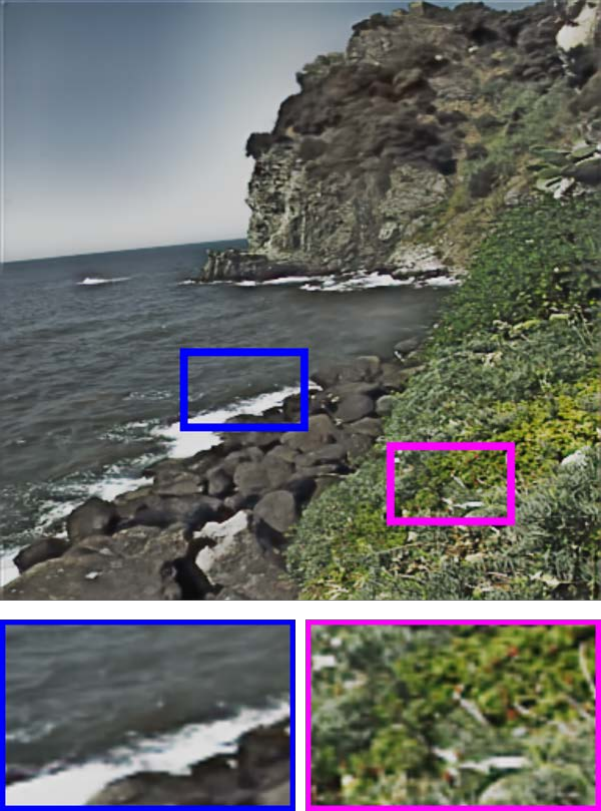}
    }\hspace{-5pt}
    \subfigure[EnlightenGAN]{
        \includegraphics[width=0.190\linewidth]{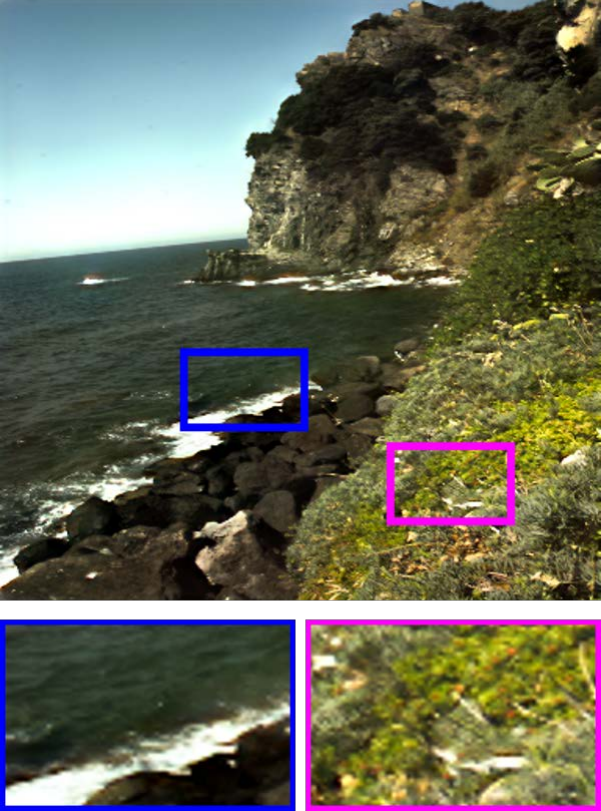}
    }\hspace{-5pt}
    \subfigure[UFormer]{
        \includegraphics[width=0.190\linewidth]{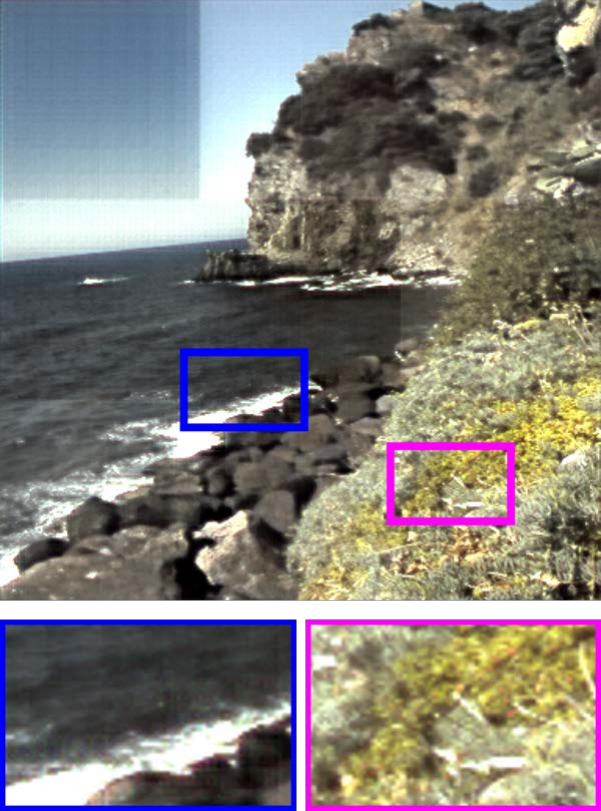}
    } 
    \subfigure[Restormer]{
        \includegraphics[width=0.190\linewidth]{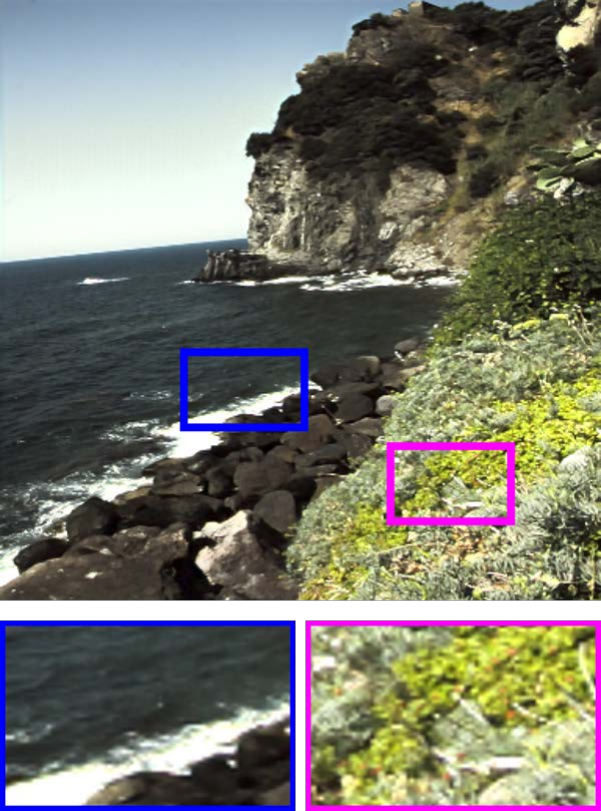}
    } \hspace{-5pt}
    \subfigure[LLFormer]{
        \includegraphics[width=0.190\linewidth]{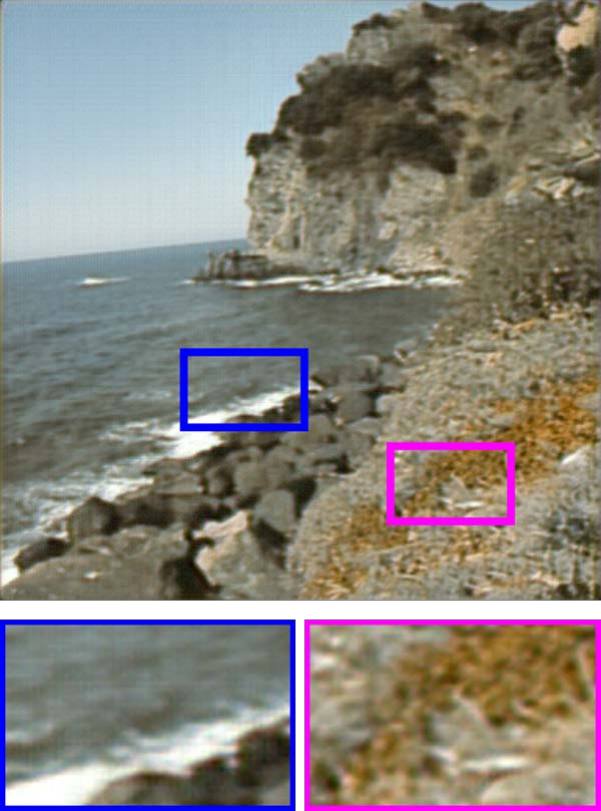}
    }\hspace{-5pt}
    \subfigure[MIRNet]{
        \includegraphics[width=0.190\linewidth]{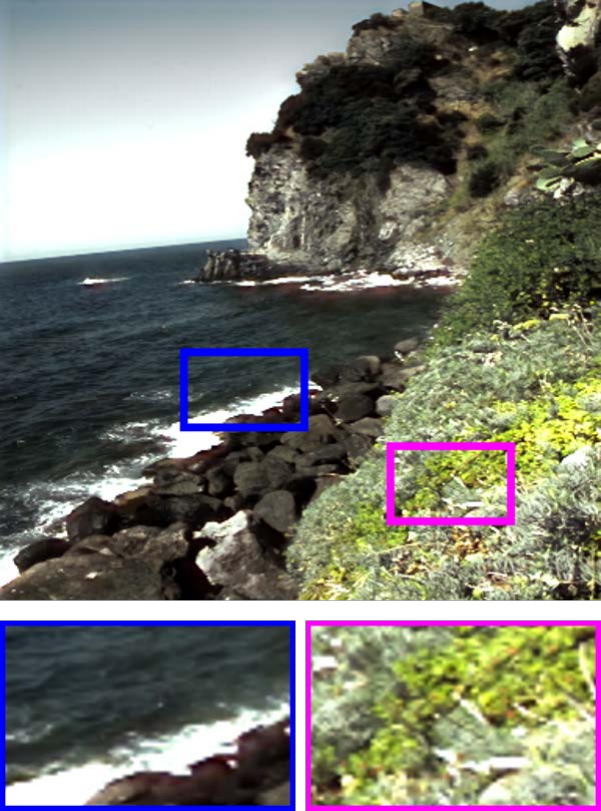}
    }\hspace{-5pt}
    \subfigure[LLDiffusion]{
        \includegraphics[width=0.190\linewidth]{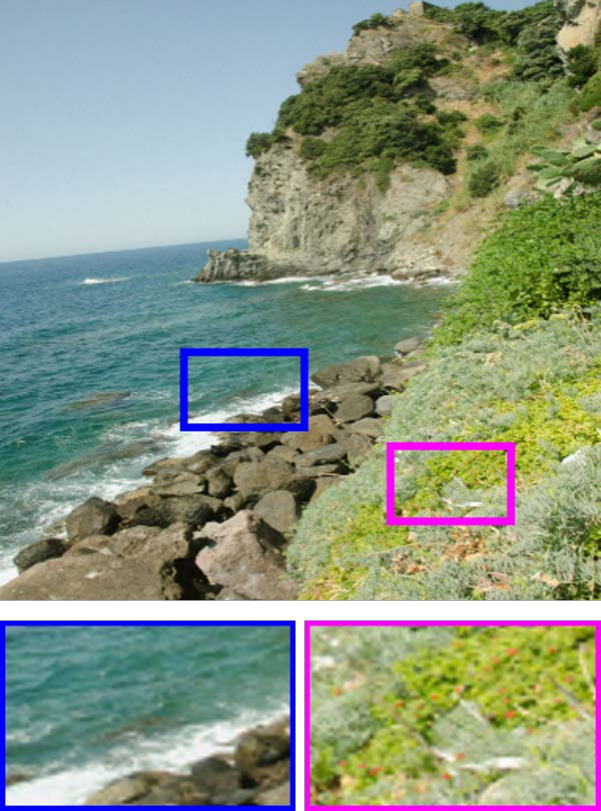}
    }\hspace{-5pt}
    \subfigure[GT]{
        \includegraphics[width=0.190\linewidth]{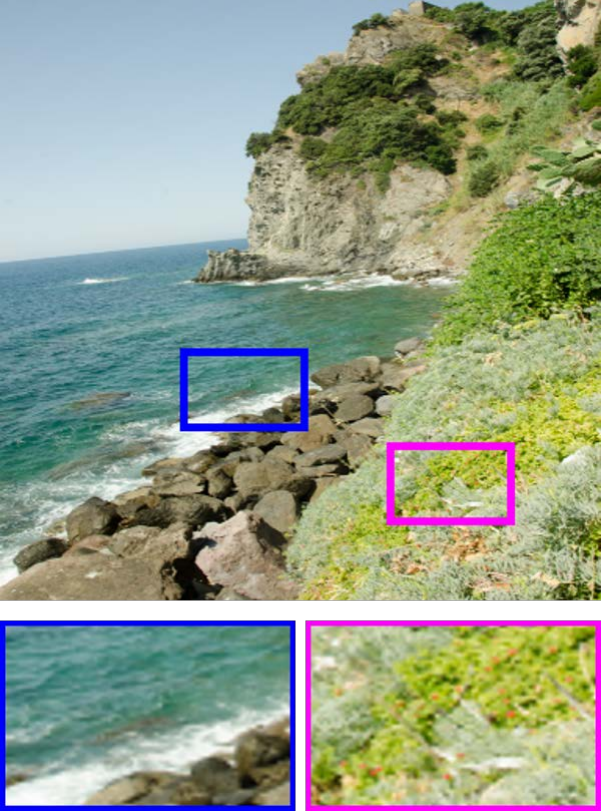}
    }
    % \vspace{-4mm}
    \caption{Visual comparison results on the subset LOL-v2-synthetic of LOL-v2~\cite{yang2021sparse} dataset. For each method, the output image is shown along with two zoomed-in regions below. The normally exposed image produced by our LLDiffusion has less artifact and better colorfulness. Zoom in for the best view.}% \textcolor{blue}{HL: pls add some discussions here.}}
    \label{fig:lol-v2-syn-result}
    % \vspace{-4mm}
    % \vspace{-0.3cm}
\end{figure*}

\textbf{Real-world unpaired datasets}. DICM~\cite{lee2012contrast}, MEF~\cite{ma2015perceptual}, and NPE~\cite{wang2013naturalness} datasets are commonly used real-world datasets for the LLIE task, which contain $69$, $17$, and $8$ real-world low-light images, respectively. However, these public datasets include only limited scenes. Therefore, we collect a new real-world dataset called Real-world Test (RWT) for testing. RWT contains $120$ low-light images including various indoor and outdoor scenes. Specifically, we crawl $120$ low-light images in the real world from the Internet. These $120$ images have complex and diverse low-light degradation. Some images of our collected RWT are illustrated in Fig.~\ref{fig:rwt_dataset}. We use these public datasets and our RWT to evaluate the generalization of the model in real-world scenery.

\textbf{Experimental details}. The proposed LLDiffusion is trained with the Adam optimizer~\cite{kingma2014adam}, in which $\beta_{1} = 0.9$, $\beta_{1} = 0.999$ for $6 \times 10^{5}$ iterations. The initial learning rate is set to $0.001$. The learning rate is decreased with a factor of $0.5$ at $1\times 10^{3}$, $1\times 10^{4}$, $6 \times 10^{5}$ iterations. We use the Kaiming initialization technique \cite{he2015delving} to initialize the weights of the proposed model and use $0.999$ Exponential Moving Average (EMA) for all of our experiments. In the training stage, the batch size and patch size are set as 8 and $128\times128$. We sample $30$ time steps for evaluation using Denoising Diffusion Implicit Model~\cite{JiamingSong2020DenoisingDI}. We first train the proposed model for $4 \times 10^{5}$ iterations of joint learning and set $\alpha$ to $0.1$. We freeze the pre-trained encoder $E$ and retrain only the image enhancement diffusion module for $2 \times 10^{4}$ iterations. The implementation is in PyTorch and we use NVIDIA Tesla V100 GPUs. For evaluation, we use referenced and no-reference metrics for paired and unpaired datasets, respectively. Specifically, for paired datasets, we adopt commonly-used PSNR and SSIM~\cite{wang2004image} metrics. In addition, we also use LPIPS~\cite{zhang2018unreasonable} to evaluate perceptual quality. For unpaired datasets, we use PI~\cite{blau20182018} and ILNIQE~\cite{zhang2015feature} to evaluate the performance. Some existing methods like LLFlow~\cite{wang2022low} compute metrics by adjusting the overall brightness according to the reference images, which may introduce biases and potential fairness issues. Thus, following existing methods~\cite{zamir2020learning,wang2022ultra}, we do not use any reference information when computing metrics.

\subsection{Comparison with Existing Methods}
\textbf{Performance on the LOL dataset}. We compare the performance of LLDiffusion with state-of-the-art LLIE methods on the LOL dataset~\cite{wei2018deep} quantitatively and qualitatively. For the quantitative comparison, we choose fifteen representative low-light image enhancement methods. BIMEF~\cite{ying2017bio}, NPE~\cite{wang2013naturalness}, LIME~\cite{guo2016lime}, MF~\cite{fu2016fusion}, and SRIE~\cite{fu2016weighted} are traditional methods using hand-craft priors to achieve enhancement. RetinexNet~\cite{wei2018deep}, KinD~\cite{zhang2019kindling}, KinD++~\cite{zhang2021beyond}, Zero-DCE~\cite{guo2020zero}, RUAS~\cite{liu2021ruas}, EnlightenGAN~\cite{jiang2021enlightengan} and MIRNet~\cite{zamir2020learning} are deep CNN based LLIE methods that mainly optimize distortion metrics. Uformer~\cite{wang2022uformer}, Restormer~\cite{zamir2022restormer}, and LLFormer~\cite{wang2022ultra} are recent transformer-based methods. Quantitative results of different state-of-the-art methods are reported in Table~\ref{tab:lol}. LLDiffusion significantly outperforms other methods on all metrics, including current SOTA methods MIRNet~\cite{zamir2020learning} and LLFormer~\cite{wang2022ultra} in terms of PSNR (SSIM) by 0.51 dB (0.013) and 1.00 dB (0.027) on average. These values show that LLDiffusion not only performs well in restoring color information but also preserves the structural information of images. In addition, LLDiffusion also obtains the best LPIPS value, indicating that it generates images closer to human visual perception. A qualitative comparison is shown in Fig.~\ref{fig:lol-result}. Among all the comparison methods, the second-ranked method MIRNet and the third-ranked method LLFormer exhibit limitations in correctly restoring the image brightness, as observed in the patch regions shown in Figure 4. These methods fail to accurately enhance the brightness of the images, leading to a loss of details and a darker appearance. On the other hand, other comparison methods demonstrate a common issue of color distortion in the restored images, where the colors appear unnatural and deviate from the original scene.
In contrast, LLDiffusion achieves the best visual results among the compared methods. The enhanced images produced by LLDiffusion exhibit clearer details, as evidenced by the improved texture in the patch regions. Furthermore, the images enhanced by LLDiffusion closely resemble the reference image, demonstrating a high level of fidelity and maintaining a more accurate representation of the scene.

\def \rootrealworld {image/LOL-v2-Real/786/}
\begin{figure*}[th]
    \centering
    \subfigure[Input]{
        \includegraphics[width=0.190\linewidth]{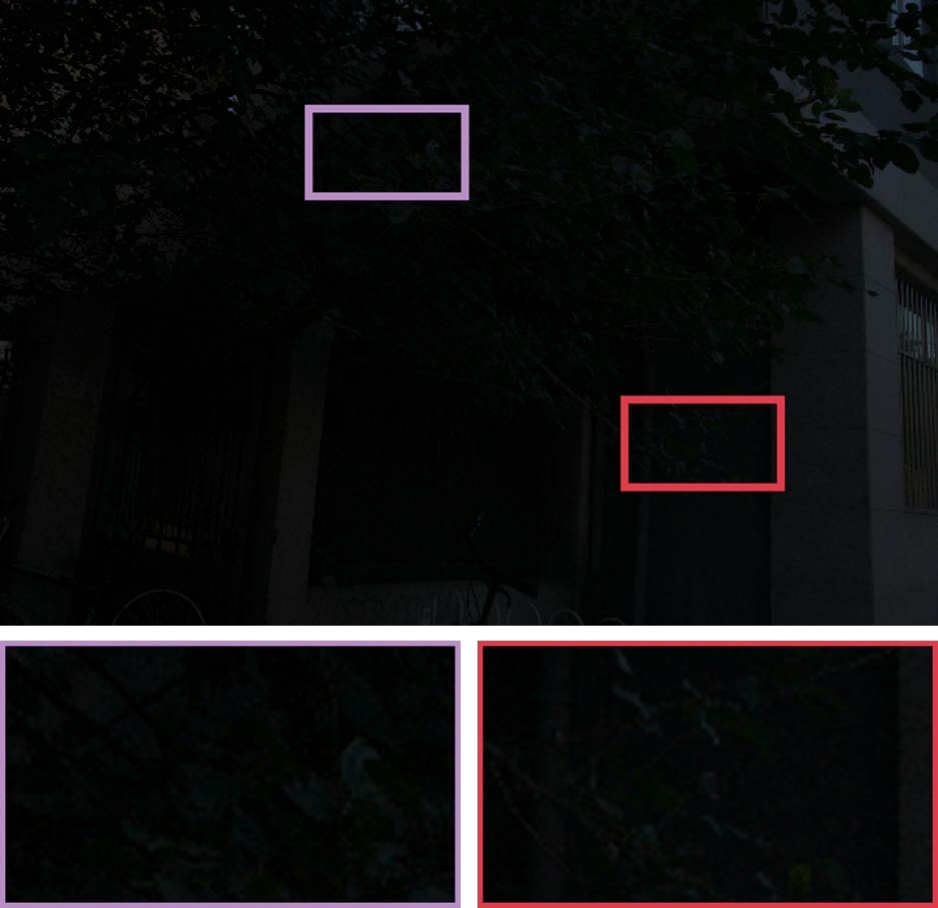}
    }\hspace{-5pt}
    \subfigure[RetinexNet]{
        \includegraphics[width=0.190\linewidth]{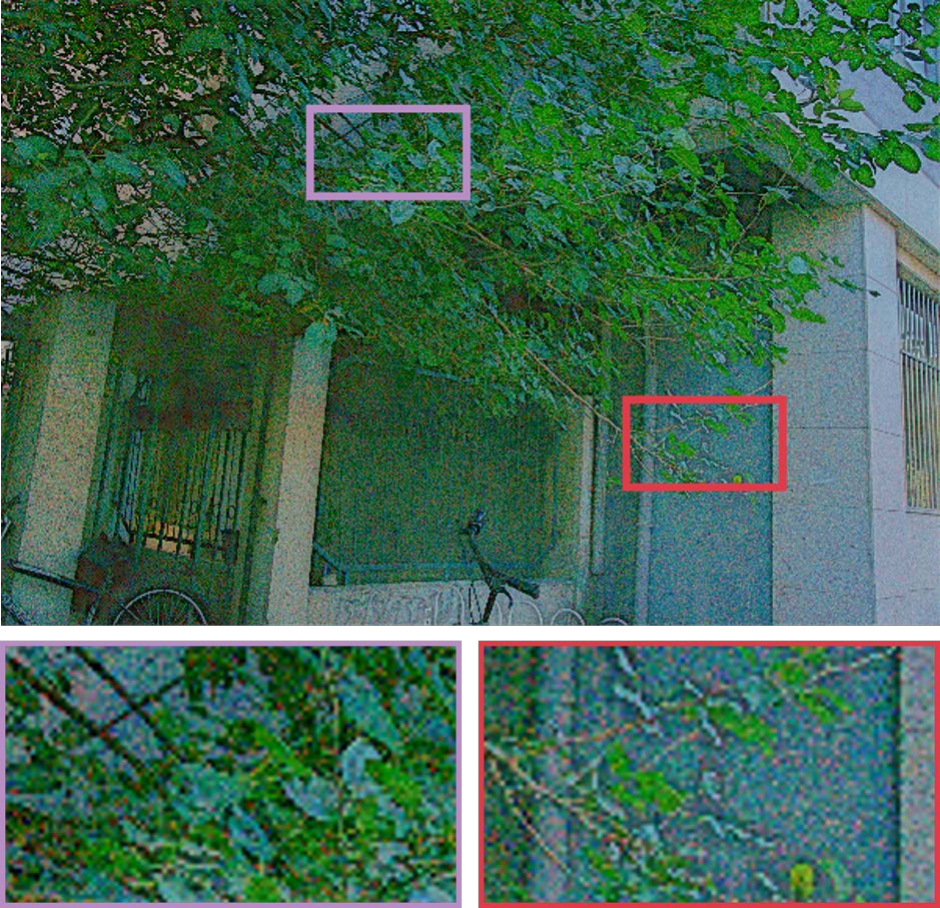}
    }\hspace{-5pt}
    \subfigure[KinD]{ 
        \includegraphics[width=0.190\linewidth]{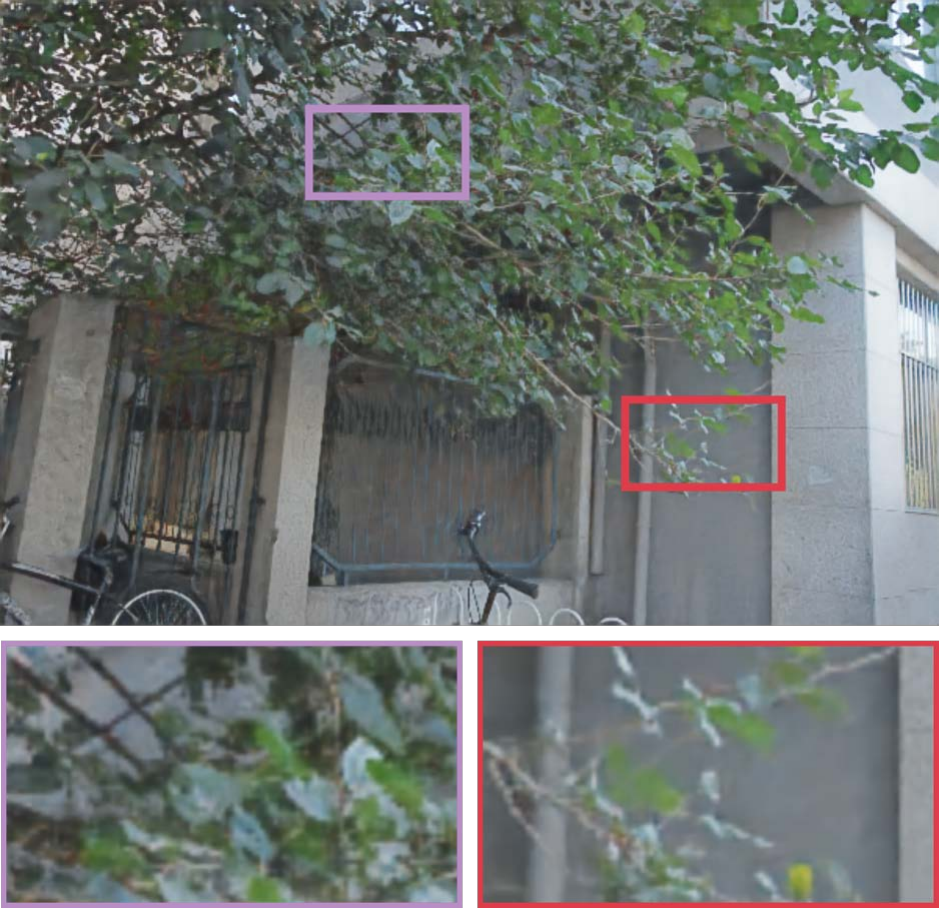}
    }\hspace{-5pt}
    \subfigure[EnlightenGAN]{
        \includegraphics[width=0.190\linewidth]{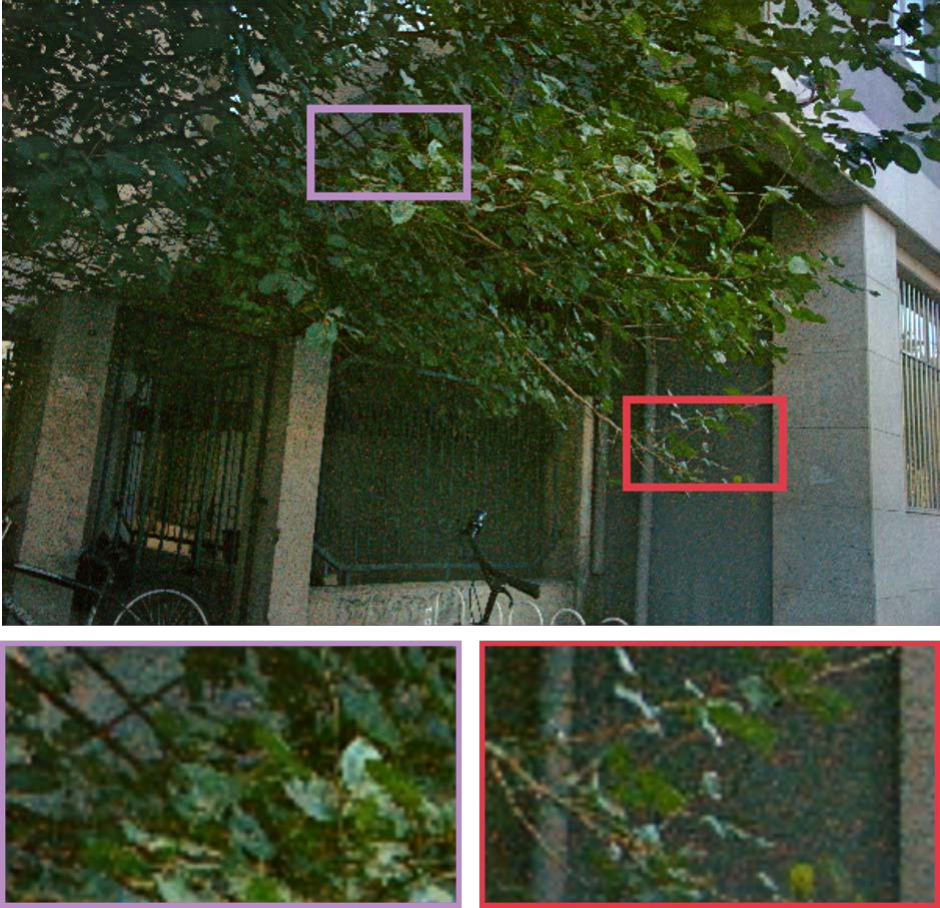}
    }\hspace{-5pt}
    \subfigure[UFormer]{
        \includegraphics[width=0.190\linewidth]{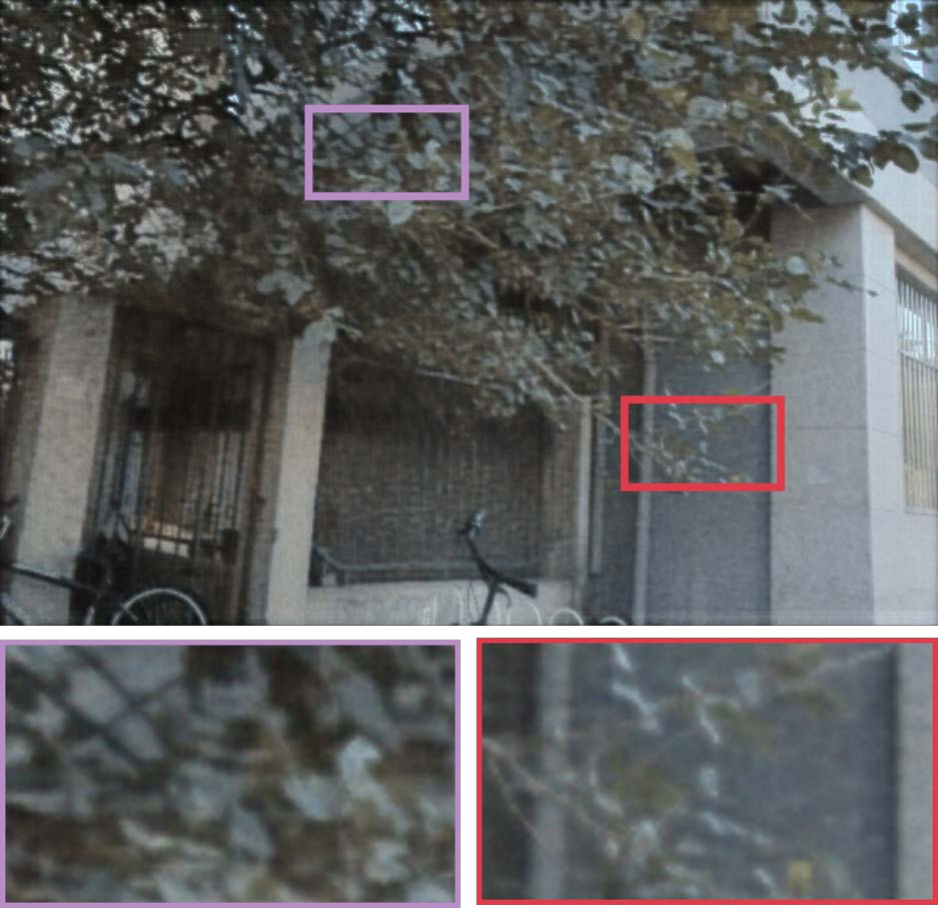}
    } 
    \subfigure[Restormer]{
        \includegraphics[width=0.190\linewidth]{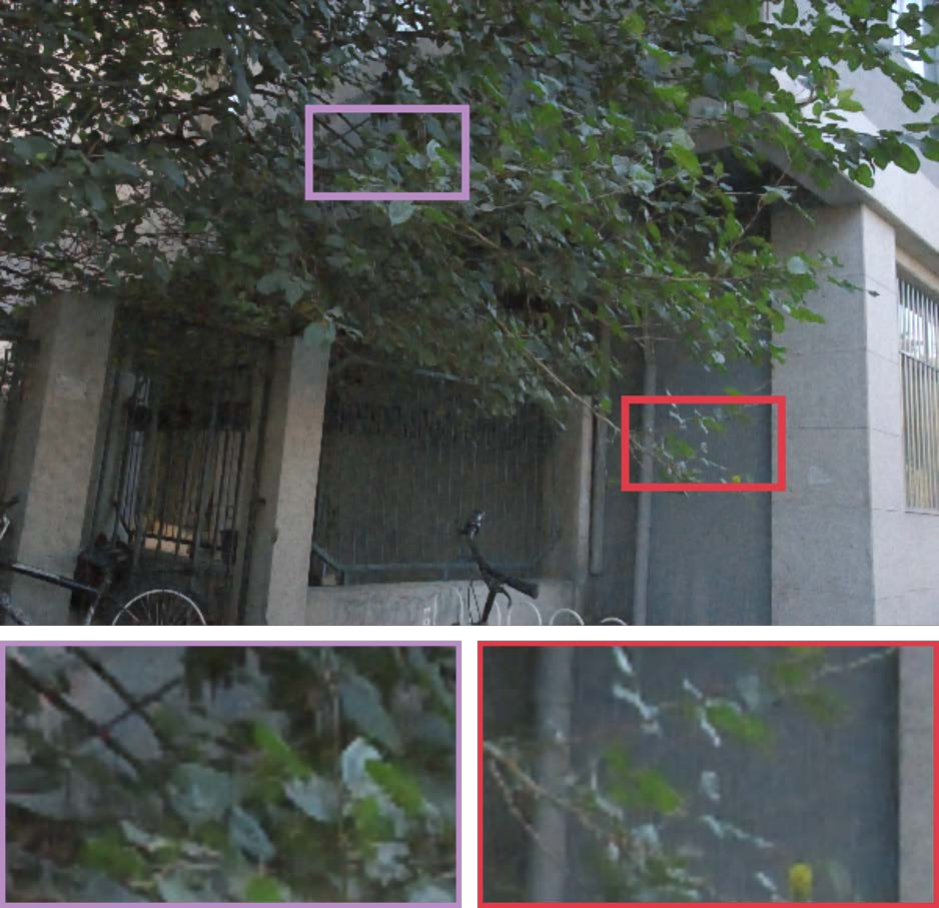}
    } \hspace{-5pt}
    \subfigure[LLFormer]{
        \includegraphics[width=0.190\linewidth]{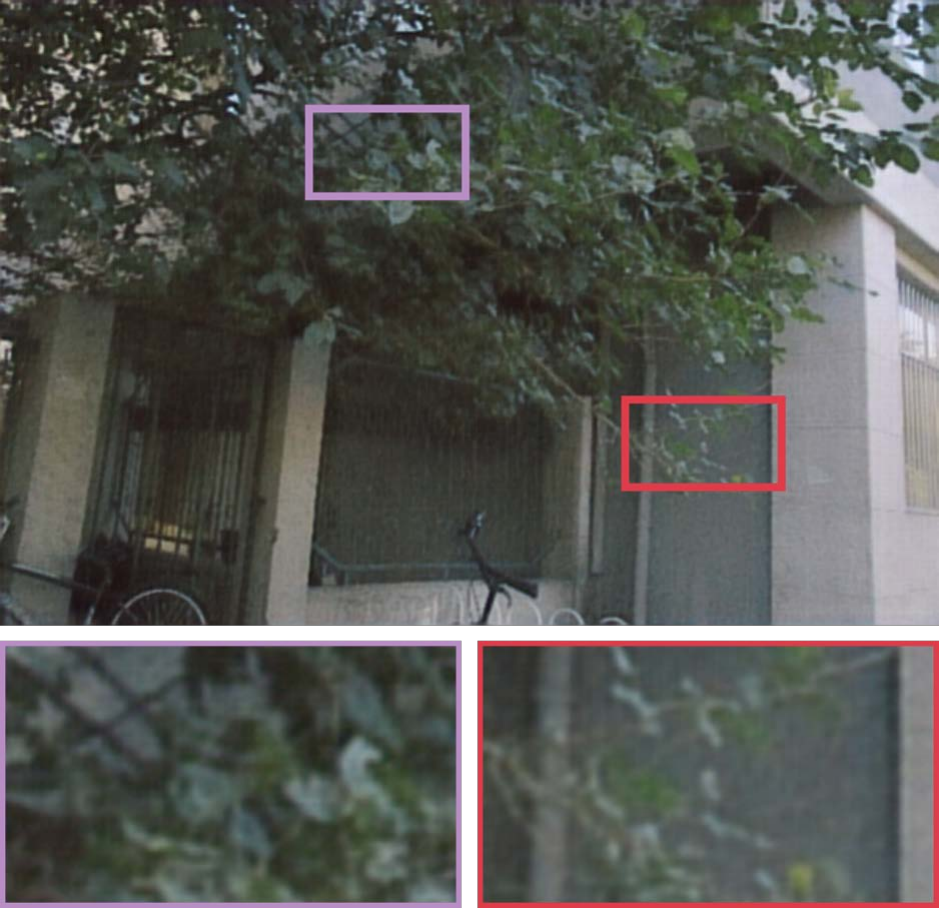}
    }\hspace{-5pt}
    \subfigure[MIRNet]{
        \includegraphics[width=0.190\linewidth]{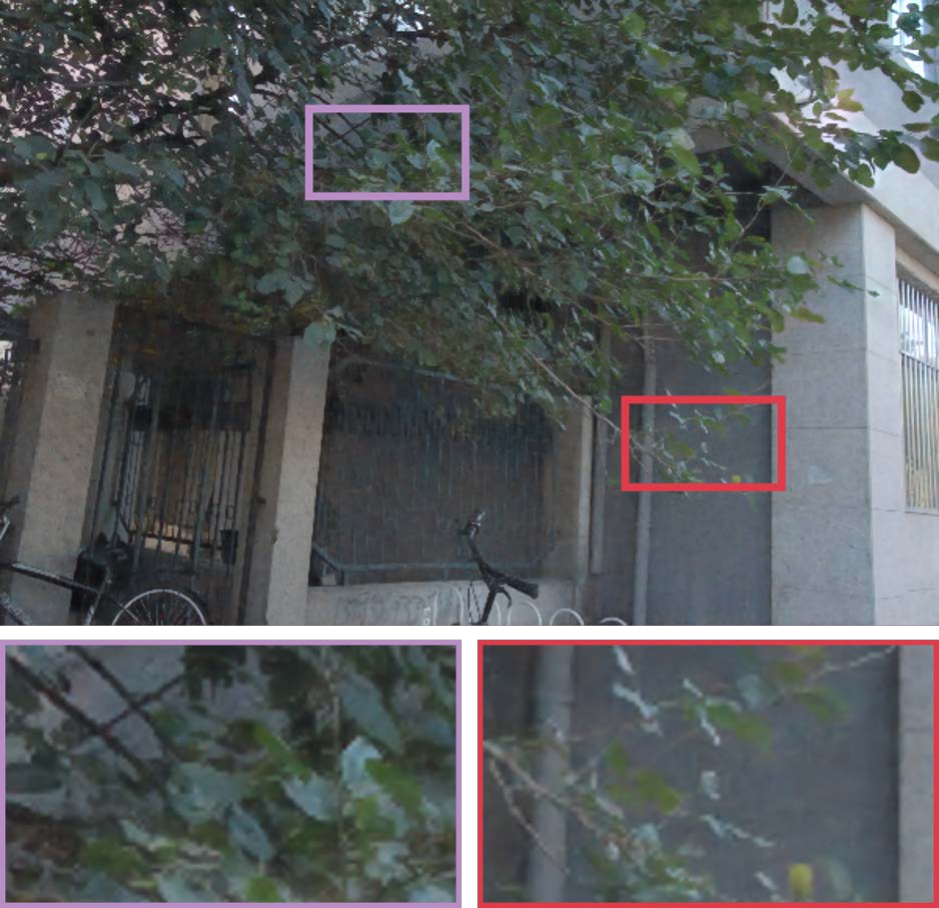}
    }\hspace{-5pt}
    \subfigure[LLDiffusion]{
        \includegraphics[width=0.190\linewidth]{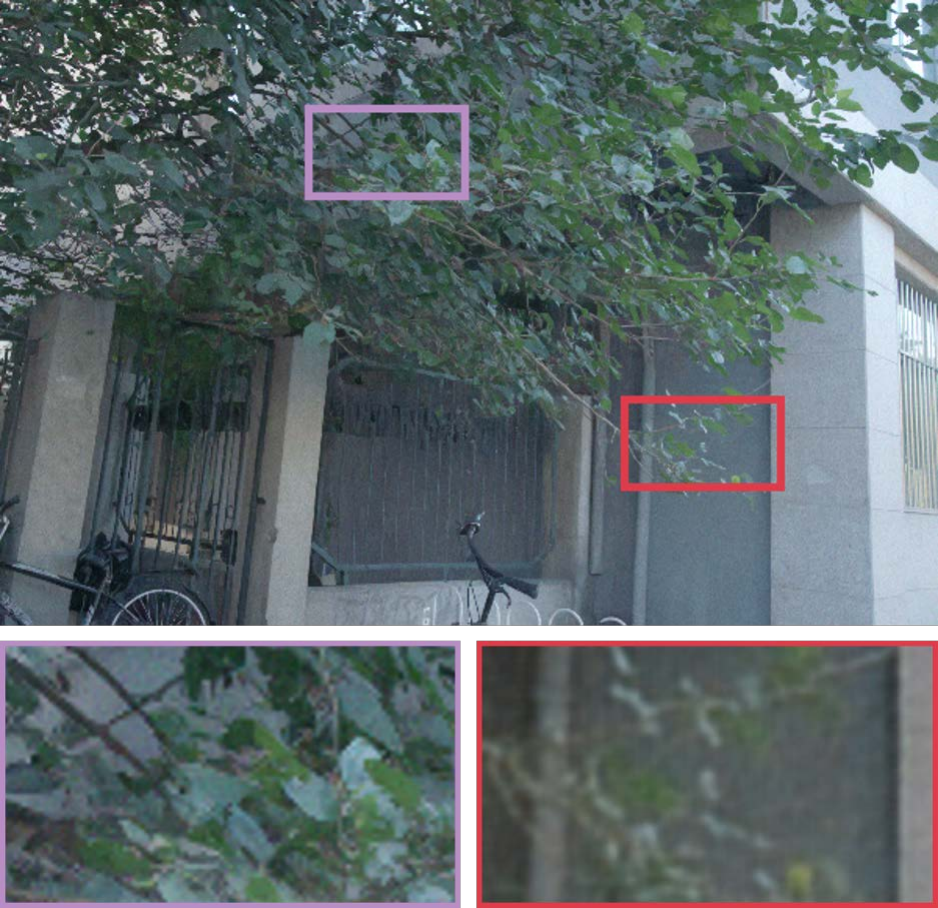}
    }\hspace{-5pt}
    \subfigure[GT]{
        \includegraphics[width=0.190\linewidth]{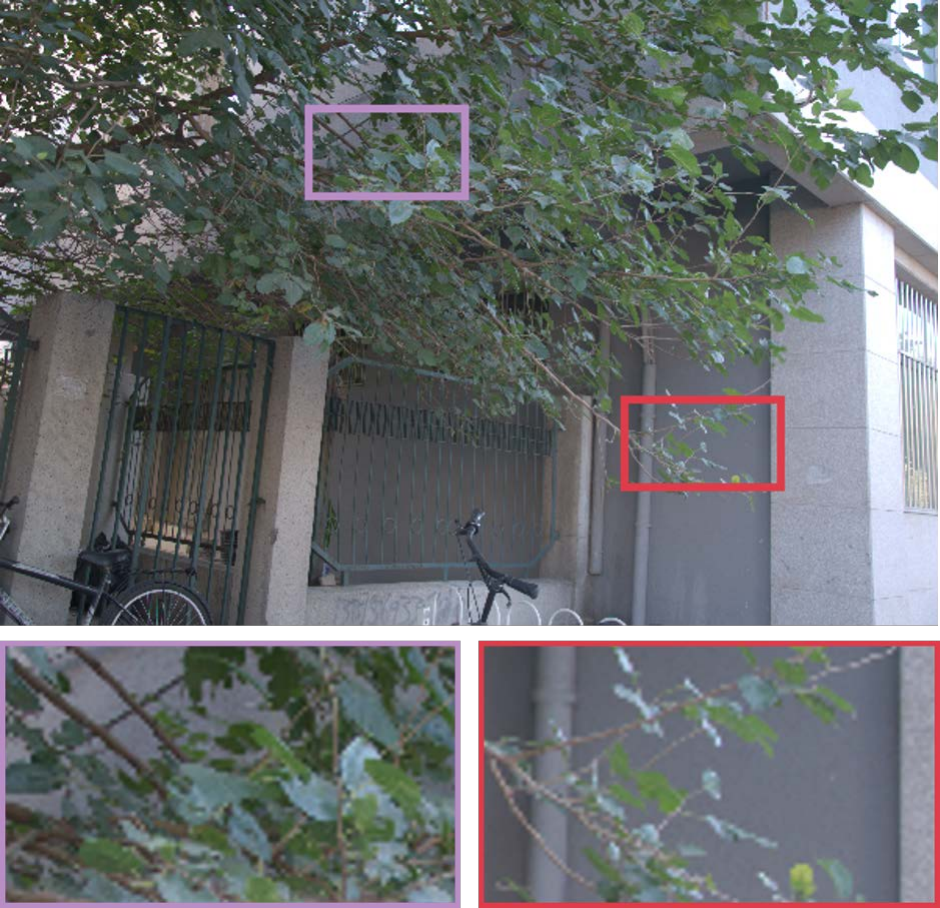}
    }
    % \vspace{-4mm}
    \caption{Visual comparison with state-of-the-art low-light image enhancement methods on the subset LOL-v2-real of LOL-v2~\cite{yang2021sparse} dataset. For each method, the output image is shown along with two zoomed-in regions below. The normal-light exposed images generated by our LLDiffusion exhibit fewer artifacts.}% 
    \label{fig:lol-v2-real-result}
\end{figure*}

%%%% The comparison results on VE-LOL benchmark. 
\def \rootrealworld {image/VE-LOL/697/}
\begin{figure*}[t]
    \centering
    \subfigure[Input]{
        \includegraphics[width=0.190\linewidth]{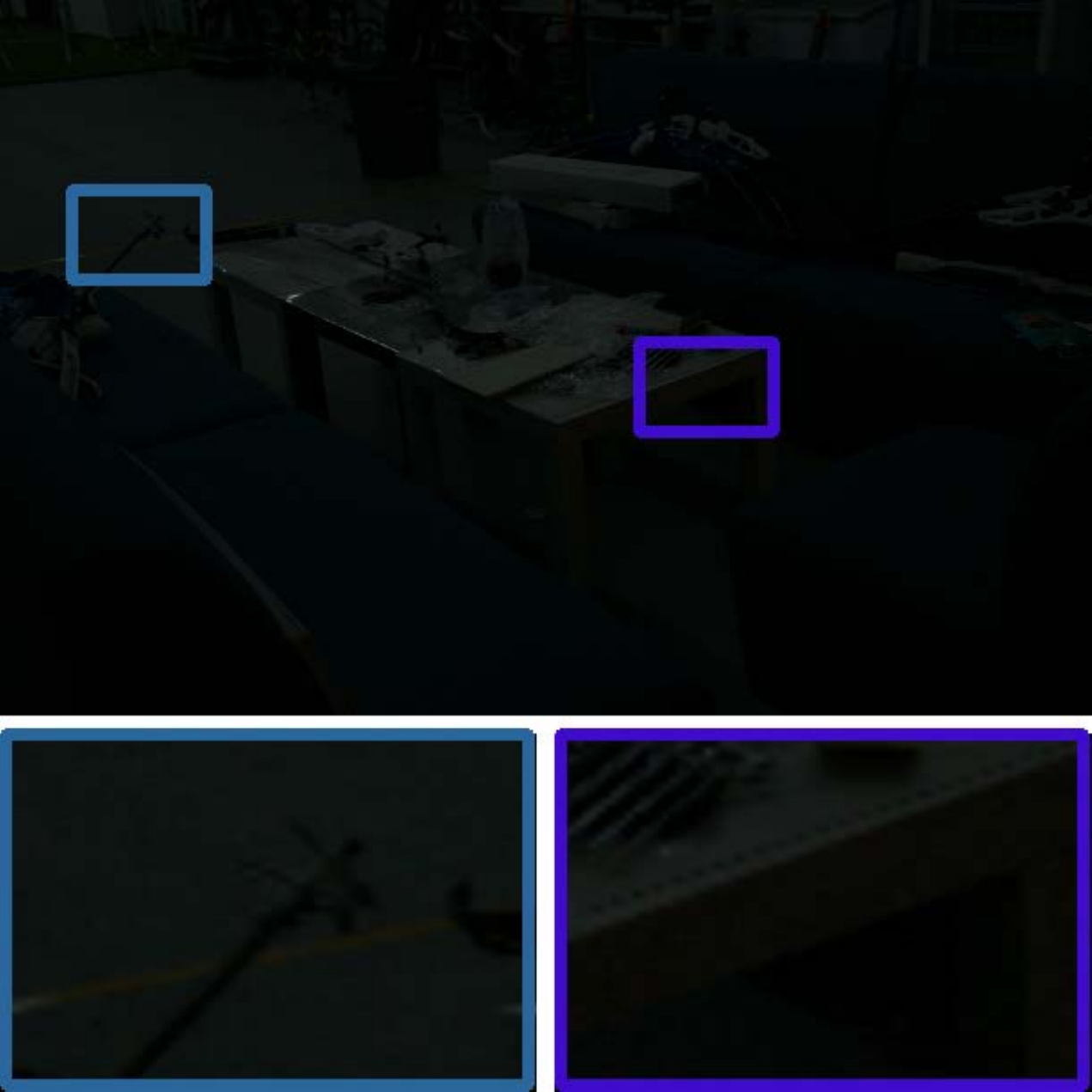}
    }\hspace{-5pt}
    \subfigure[RetinexNet]{
        \includegraphics[width=0.190\linewidth]{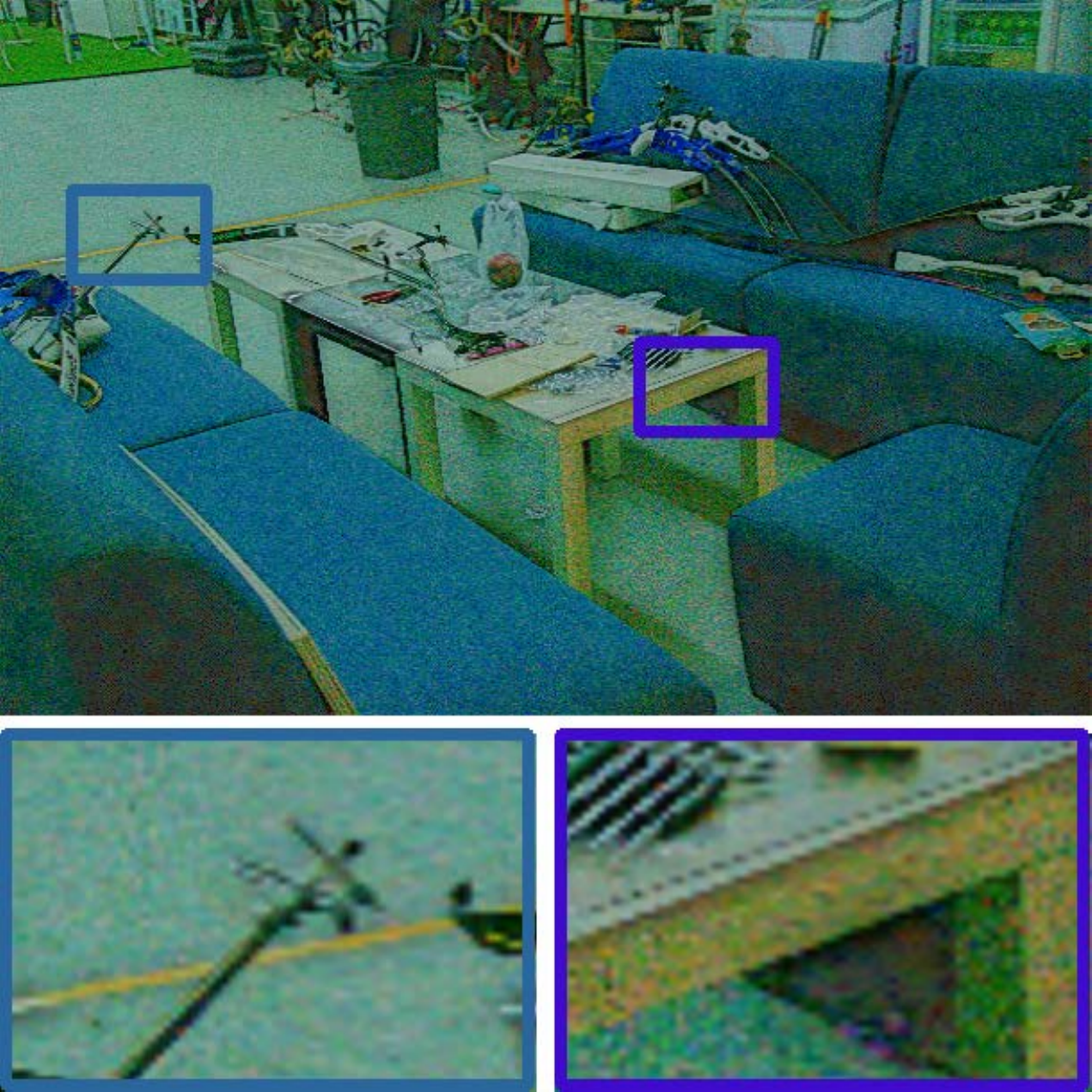}
    }\hspace{-5pt}
    \subfigure[KinD]{ 
        \includegraphics[width=0.190\linewidth]{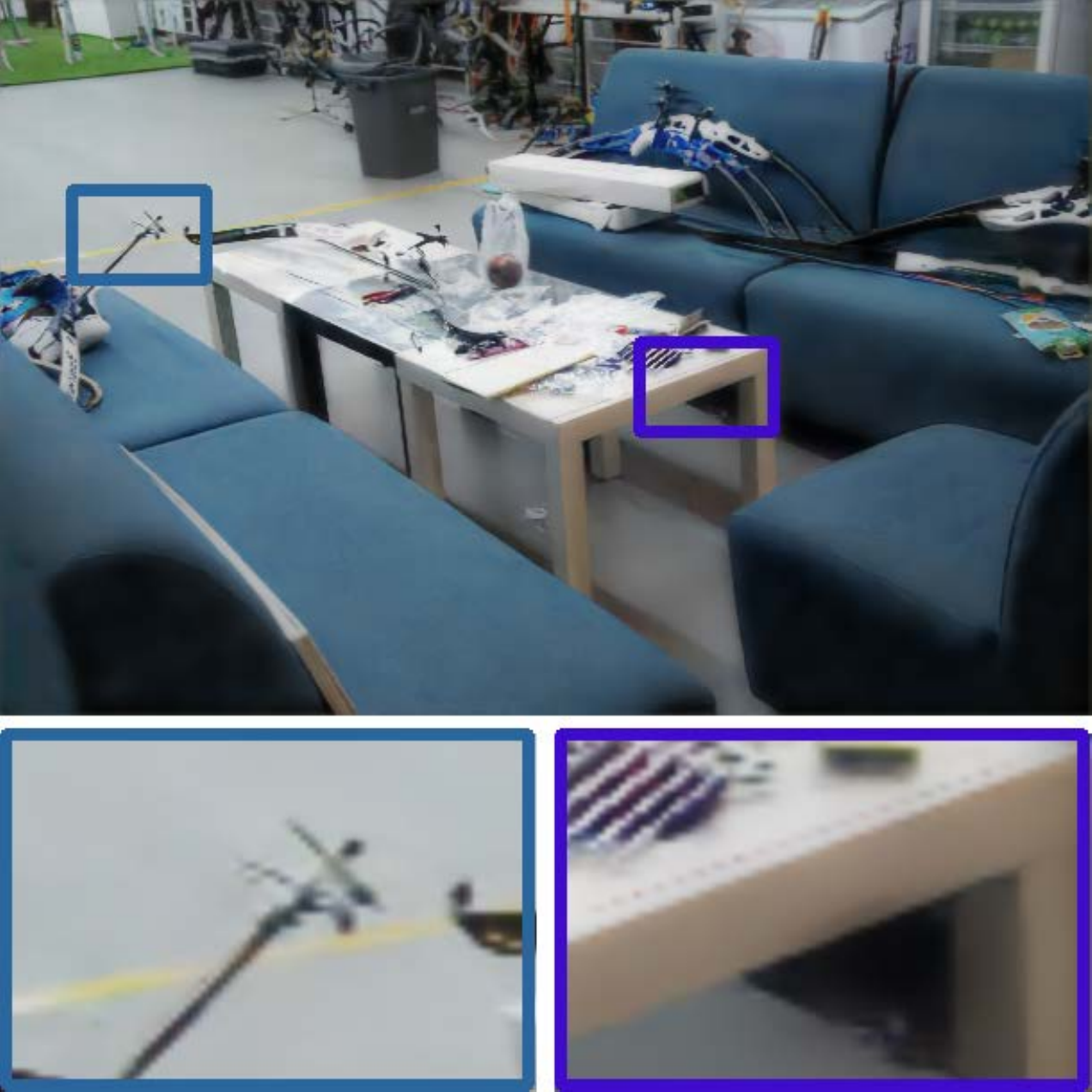}
    }\hspace{-5pt}
    \subfigure[KinD++]{
        \includegraphics[width=0.190\linewidth]{\rootrealworld KinD.pdf}
    }\hspace{-5pt}
    \subfigure[UFormer]{
        \includegraphics[width=0.190\linewidth]{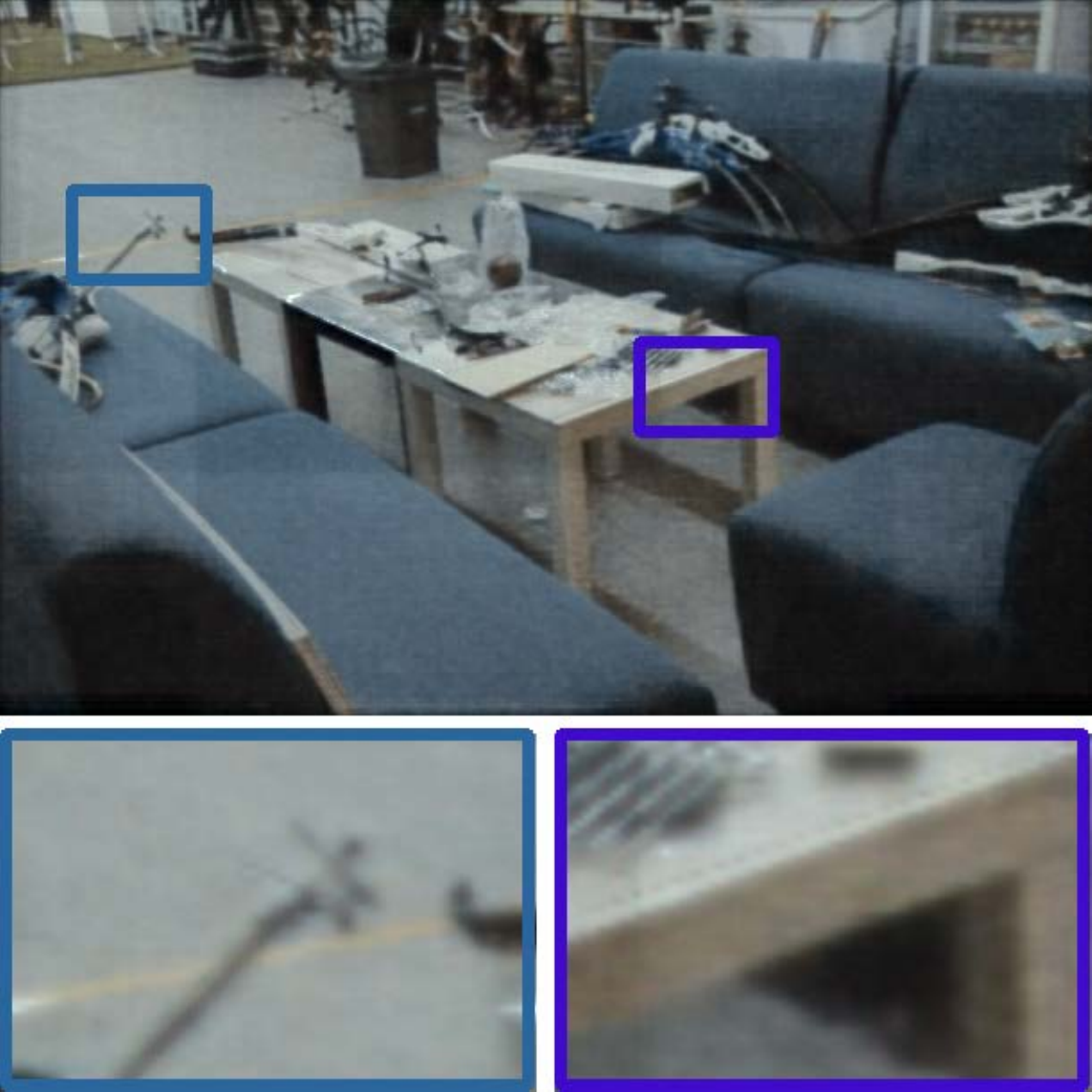}
    }\vspace{-1.0mm}
    \subfigure[Restormer]{
        \includegraphics[width=0.190\linewidth]{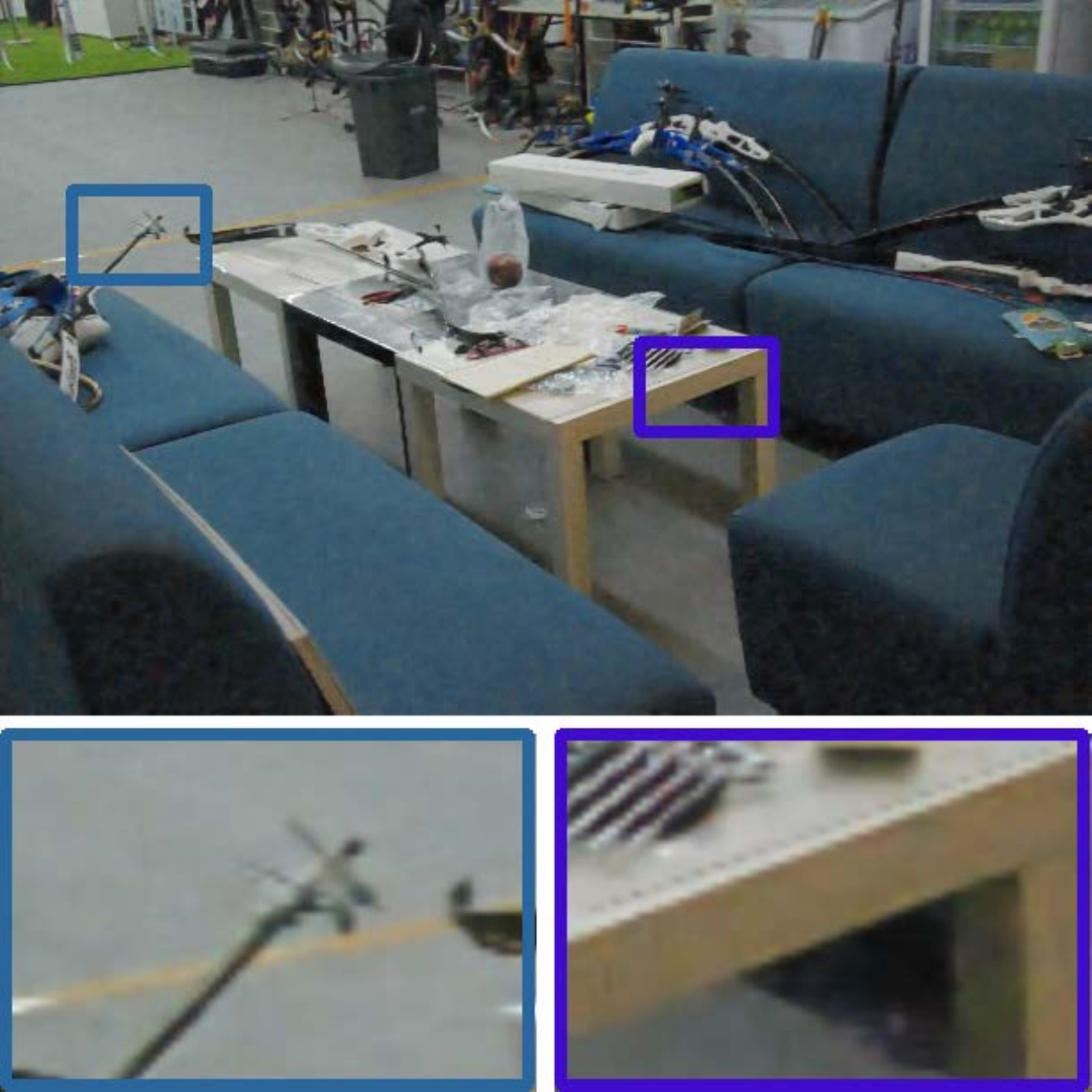}
    }\hspace{-5pt}
    \subfigure[LLFormer]{
        \includegraphics[width=0.190\linewidth]{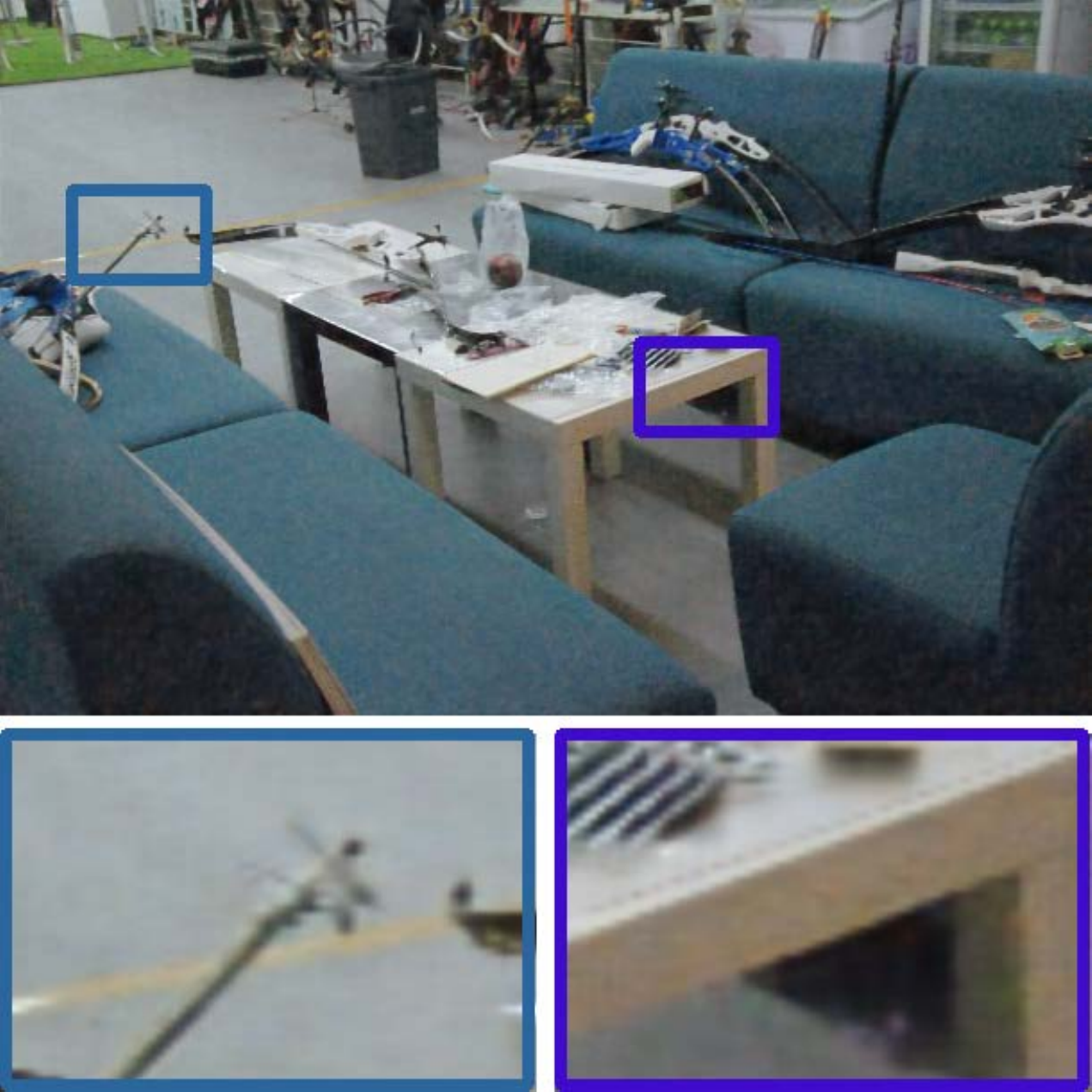}
    }\hspace{-5pt}
    \subfigure[MIRNet]{
        \includegraphics[width=0.190\linewidth]{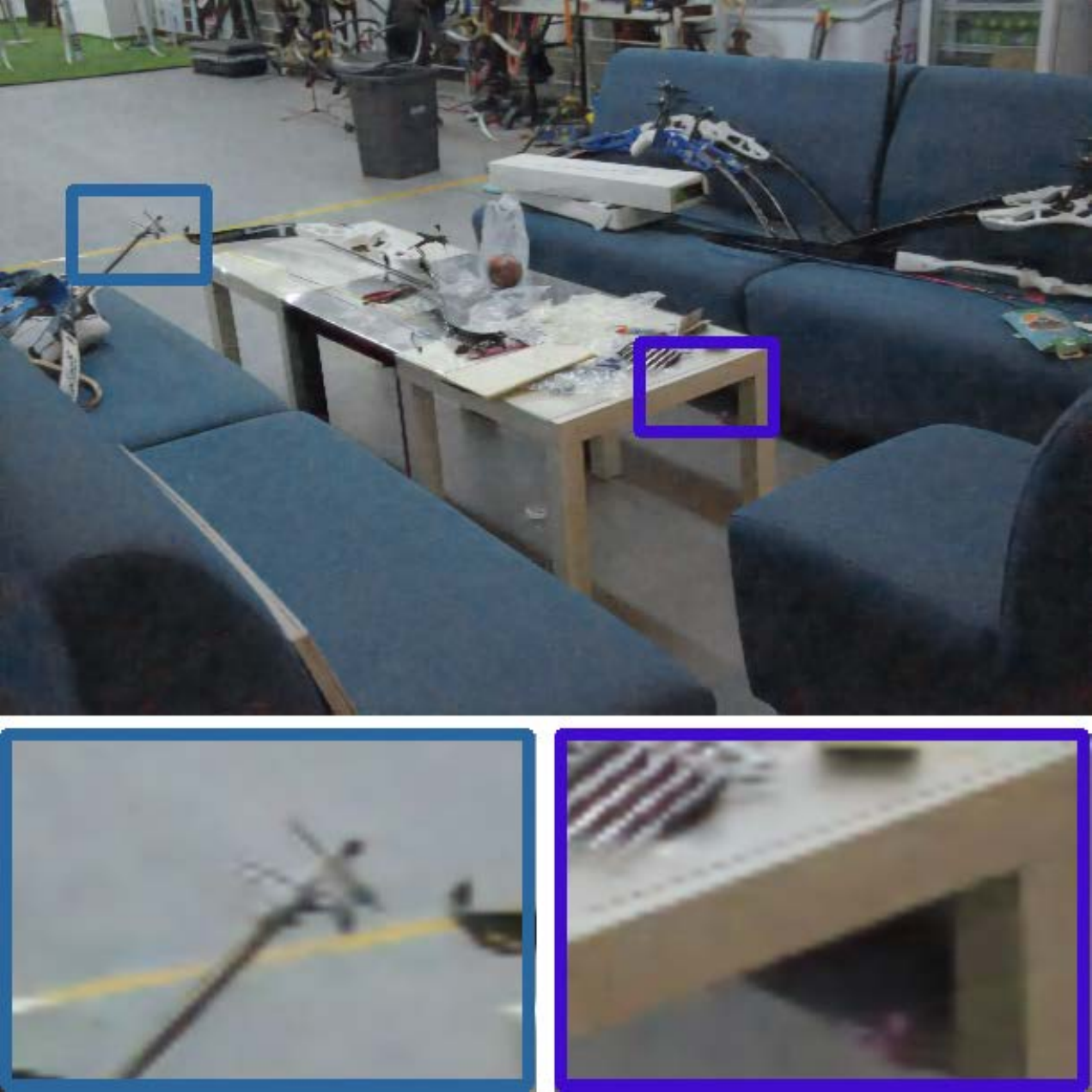}
    }\hspace{-5pt}
    \subfigure[LLDiffusion]{
        \includegraphics[width=0.190\linewidth]{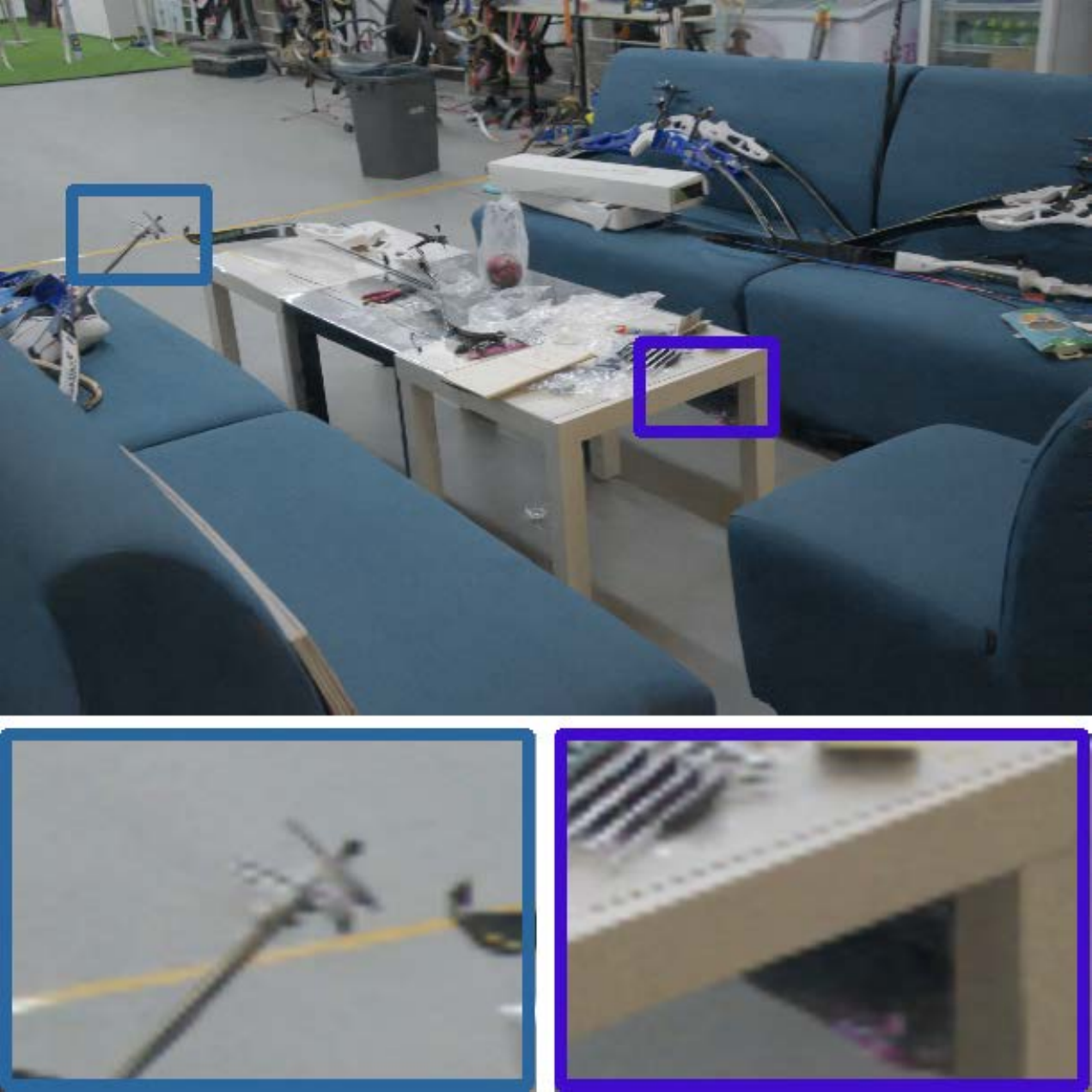}
    }\hspace{-5pt}
    \subfigure[GT]{
        \includegraphics[width=0.190\linewidth]{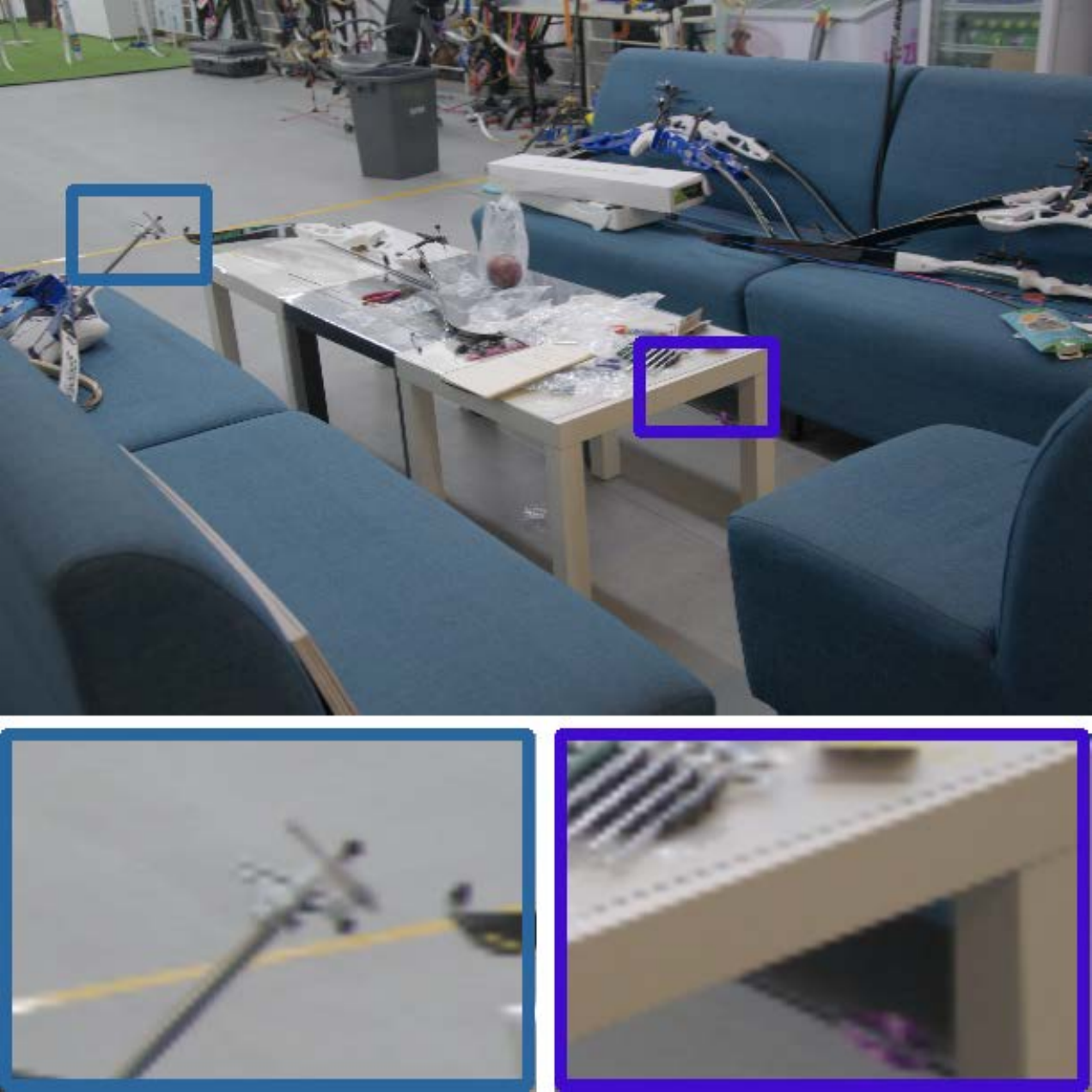}
    }\hspace{-5pt}
    % \vspace{-4mm}
    \caption{Qualitative results on VE-LOL~\cite{liu2021benchmarking}. Comparison with state-of-the-art low-light image enhancement methods on the VE-LOL dataset. Our LLDiffusion better recovers image details and recovers image contrast closer to the ground-truth image.}% \textcolor{blue}{HL: pls add some discussions here.}}
    % \vspace{-4mm}
    \label{fig:velol-result}
\end{figure*}

%%%% The comparison results on VE-LOL benchmark.
\begin{table}[t]
    \centering
    \caption{Evaluation on the LOL-v2 dataset~\cite{yang2021sparse} in terms of PSNR and SSIM. %The top two results are marked in \textcolor{red}{red} and \textcolor{blue}{blue}, respectively. 
    We highlight the \firstone{best} and the \secondone{second-best} values for each metric.}
        % \vspace{-4mm}
    \scalebox{0.92}{
    \begin{tabular}{c|cc|cc}
    \toprule
    % \multirow{2}{*}{LOL-v2-syn} & \multirow{2}{*}{LOL-v2-real}\\
    \multirow{2}{*}{Methods}  & \multicolumn{2}{c|}{LOL-v2-syn} & \multicolumn{2}{c}{LOL-v2-real} \\ 
     &  PSNR $\uparrow$ & SSIM $\uparrow$ & PSNR $\uparrow$ & SSIM $\uparrow$ \\
    \midrule
    SID~\cite{chen2019seeing}  &15.04 &0.610 &13.24 &0.442\\
    3DLUT~\cite{zeng2020learning}   &18.04  & 0.800  & 17.59 &0.721\\
    DeepUPE~\cite{wang2019underexposed}  &15.08 &0.623 & 13.27 & 0.452\\
    RF~\cite{kosugi2020unpaired}  &15.97 &0.632 & 14.05 &0.458 \\
    DeepLPE~\cite{moran2020deeplpf}  &16.02 &0.587  & 14.10 &0.480\\
    IPT~\cite{chen2021pre}  &18.30 &0.811 & 19.80 &0.813 \\
    Uformer~\cite{wang2022uformer}  &19.66 &0.871 &18.82 &0.771\\
    RetinexNet~\cite{wei2018deep}  &17.13 & 0.798 & 15.47 &0.567 \\
    Sparse~\cite{yang2021sparse}  &22.05 &0.905 &20.06 &0.816\\
    EnlightenGAN~\cite{jiang2021enlightengan}  &16.57 &0.734 &18.23 &0.617 \\
    RUAS~\cite{liu2021ruas}  &16.55& 0.652 &18.37& 0.723\\
    FIDE~\cite{xu2020learning}  &15.20 &0.612 &16.85 &0.678\\
    DRBN~\cite{xu2020learning}  &23.22 &\secondone{0.927} &20.29 &\secondone{0.831}\\
    KinD~\cite{zhang2019kindling}  &13.29 &0.578 &14.74 &0.641\\
    Restormer~\cite{zamir2022restormer}  &21.41 &0.830 &19.94 &0.827\\
    MIRNet~\cite{zamir2020learning}  &21.94 &0.876 &20.02 &0.820\\
    LLFormer~\cite{wang2022ultra} &\secondone{23.55} & 0.905&\secondone{20.48} &0.787 \\
    LLDiffusion & \firstone{25.99} &\firstone{0.948} &\firstone{23.16} & \firstone{0.842}\\
    \bottomrule
    \end{tabular}
    }
    \label{tab:lol-v2}
\end{table}

\textbf{Performance on the LOL-v2 dataset}. We also compare LLDiffusion with a wide range of low-light enhancement algorithms on the LOV-v2 dataset. Table~\ref{tab:lol-v2} shows the comparison results. LLDiffusion achieves the best performance in terms of PSNR and SSIM on both LOL-v2-synthetic and LOL-v2-real subsets. Specifically, (1)
compared with the recent state-of-the-art method LLFormer~\cite{wang2022ultra}, LLDiffusion acquires $2.44$ dB, $2.68$ dB improvements in terms of PSNR on LOL-v2-synthetic and LOL-v2-real respectively. (2) Compared to other retinex-based methods such as DeepUPE~\cite{wang2019underexposed}, RetinexNet~\cite{wei2018deep}, RUAS~\cite{liu2021ruas}, and KinD~\cite{zhang2019kindling}, LLDiffusion demonstrates superior performance. Specifically, LLDiffusion improves PSNR and SSIM by 8.86 dB and 0.150, respectively, on LOL-v2-synthetic compared to RetinexNet, and by 4.79 dB and 0.119, respectively, on LOL-v2-real compared to RUAS. (3) Compared with recent Transformer-based methods, \ie IPT~\cite{chen2021pre}, Uformer~\cite{wang2022uformer}, Restormer~\cite{zamir2022restormer}, and LLFormer~\cite{wang2022ultra}, LLDiffusion is still superior to these methods in terms of PSNR and SSIM.     
Fig.~\ref{fig:lol-v2-syn-result} and Fig.~\ref{fig:lol-v2-real-result} describe a qualitative comparison of two subsets from the LOL-v2 dataset under different low-light conditions (different exposures and degenerations). The results show that our LLDiffusion can handle all low-light situations, producing results with clear and trustworthy detail, as well as better fidelity and naturalness. For example, from the "sea" and "plant" patches in Fig.~\ref{fig:lol-v2-syn-result}, it can be seen that only LLDiffusion recovered credible image detail and brought the contrast closer to the ground-truth image, while its competitors failed to address the degradation problem and the result presented a significant color deviation.

\begin{table}[t]
    \centering
    \caption{Evaluation on the VE-LOL dataset~\cite{liu2021benchmarking} in terms of PSNR, SSIM and LPIPS. 
    We highlight the \firstone{best} and the \secondone{second-best} values for each metric.
    }
    \begin{tabular}{cccc}
    \toprule
     Method  &  PSNR $\uparrow$ & SSIM $\uparrow$ & LPIPS $\downarrow$ \\
    \midrule
    BIMEF~\cite{ying2017bio} & 15.95 & 0.639 & 0.457 \\
    LIME~\cite{guo2016lime} & 14.07 & 0.527 & 0.402 \\  
    SRIE~\cite{fu2016weighted} & 13.66 & 0.551 & 0.458\\
    LLNet~\cite{lore2017llnet} & 17.57 & 0.739 & 0.402 \\
    RetinexNet~\cite{wei2018deep} & 14.68 & 0.525 & 0.642\\
    DeepUPE~\cite{wang2019underexposed} & 13.19 & 0.490 & 0.463 \\
    % JED \cite{ren2018joint} & 16.73 & 0.6817 & 0.3899 \\
    SICE~\cite{cai2018learning} & 18.06 & 0.709 & 0.508 \\
    KinD~\cite{zhang2019kindling} & 18.42 & 0.766 & 0.288\\
    KinD++~\cite{zhang2021beyond} & 17.63 & 0.800 & 0.226\\
    % MIRNET \cite{zamir2020learning} & 21.93 & 0.8307 & 0.1811 \\
    Zero-DCE~\cite{guo2020zero} & 21.12 & 0.771 & 0.248 \\
    EnlightenGAN~\cite{jiang2021enlightengan} & 20.43 & 0.792 & 0.242 \\
    Uformer~\cite{wang2022uformer} &17.94 &0.743 &0.325 \\
    Restormer~\cite{zamir2022restormer} &24.48 &0.857 & 0.1357\\
    LLFormer~\cite{wang2022ultra} &27.75 & 0.860&0.143 \\
    MIRNet~\cite{zamir2020learning} &\secondone{28.10}&\secondone{0.889}&\secondone{0.106}\\
    LLDiffusion & \firstone{31.77} & \firstone{0.902} & \firstone{0.040} \\ 
    \bottomrule
    \end{tabular}
    % }
    \label{tab:ve-lol}
     % \vspace{-5.2mm}
\end{table}
\textbf{Performance on the VE-LOL dataset}. To extensively evaluate the performance and generalization of LLDiffusion, we further test the method on the VE-LOL dataset~\cite{liu2021benchmarking}, which is more challenging than the LOL dataset. In this dataset, the comparison methods are first trained on LOL~\cite{wei2018deep}, then the trained models are directly evaluated on the VE-LOL test set~\cite{liu2021benchmarking}. Table~\ref{tab:ve-lol} 
shows the comparison results. LLDiffusion obtains the highest PSNR and SSIM values, indicating that among all the comparison methods, the enhancement result of LLDiffusion is closer to the real ground truth image. Specifically, compared with the recent best MIRNet method, our method improves the PSNR from $28.10$~dB to $31.77$~dB. For the SSIM metric, LLDiffusion is superior to other methods, demonstrating that LLDiffusion can better restore the structures of images. For the LPIPS metric, LLDiffusion ranks first place, \ie LLDiffusion can produce more visually realistic results. In addition, we also show the visual comparison results, which are shown in Fig.~\ref{fig:velol-result}. LLDiffusion successfully estimates normally exposed images with fewer artifacts or noise.

%%%% The comparison results on the real-world benchmark.
\begin{table*}[t]
\setlength{\tabcolsep}{3pt}
    \centering
    \caption{Evaluation on the real-world unpaired datasets including DICM~\cite{liu2021benchmarking}, MEF~\cite{ma2015perceptual}, NPE~\cite{wang2013naturalness}, and our collected RWT.  
    We highlight the \firstone{best} and the \secondone{second-best} values for each metric.
    }
     % \vspace{-2mm}
    % \scalebox{0.96}{
    \begin{tabular}{c|cc|cc|cc|cc}
    \toprule
    \multirow{2}{*}{Methods} & \multicolumn{2}{c|}{DICM~\cite{lee2012contrast}}  & \multicolumn{2}{c|}{MEF~\cite{ma2015perceptual}} & \multicolumn{2}{c}{NPE~\cite{wang2013naturalness}} & \multicolumn{2}{c}{RWT} \\ 
     & PI $\downarrow$  & ILNIQE $\downarrow$   & PI $\downarrow$  & ILNIQE $\downarrow$ & PI $\downarrow$  & ILNIQE $\downarrow$ & PI $\downarrow$  & ILNIQE $\downarrow$\\ 
    \midrule
    Input  &3.5262 &28.2519   &3.3732 &31.6624  &2.9492 &27.3469 & 4.6148 & 36.1893\\
    RetinexNet~\cite{wei2018deep} 	&\secondone{3.2494}		&27.9314   &3.0174 &\secondone{23.5726}  &2.9581 &28.6988 & 3.3800 & 26.9690\\
    KinD~\cite{zhang2019kindling}  &3.5141 &26.1600  &3.1475 &28.8427  &3.3326 &29.6344 & 3.4665&28.4822\\
    KinD++~\cite{zhang2021beyond}  & 3.3473 &25.1380   &2.8237 &25.8564  &3.2880 &28.8251&3.3295&27.4456\\
    Uformer~\cite{wang2022uformer} & 3.2551 & 24.0142   &3.3722 &27.8308  &4.8211 &30.0771 &3.6507&25.3465\\
    Restormer~\cite{zamir2022restormer} & 3.5799  & 24.3929 &\secondone{2.9476} &24.8303  &\secondone{2.8784} &27.2111 & \secondone{3.2296}&25.0798 \\
    MIRNet~\cite{zamir2020learning}  & 3.7686  & 25.6681   &2.9779 &27.0997  &3.2213 &29.7564 &3.4564&29.2997 \\
    LLFormer~\cite{wang2022ultra}  & 3.4251 & \secondone{23.3539}  & 3.0783 & 24.3855  & 2.9379 & \firstone{25.3209}&3.3670&24.1830 \\
    LLDiffusion  & \firstone{3.1284} & \firstone{22.4767}  & \firstone{2.7602}
 & \firstone{22.0691}  & \firstone{2.8713} & \secondone{27.0381} & \firstone{2.9713}&\firstone{23.5172}\\
    \bottomrule
    \end{tabular}
    % }
    \label{tab:real-word}
       % \vspace{-3mm}
\end{table*}

\def \rootdicm {image/image-real/DICM/05/}
\def \rootmef {image/image-real/MEF/Candle/}
\def \rootnpe {image/image-real/NPE/Parking/}
\def \rootrealworld {image/image-real/RWT/022/}

\renewcommand{\tabcolsep}{.6pt}  
\begin{figure*}
% \vspace{-0.2cm}
% \begin{minipage}{\textwidth}
\begin{center}
\begin{tabular}{ccccccccc}
    \includegraphics[width=\swten]{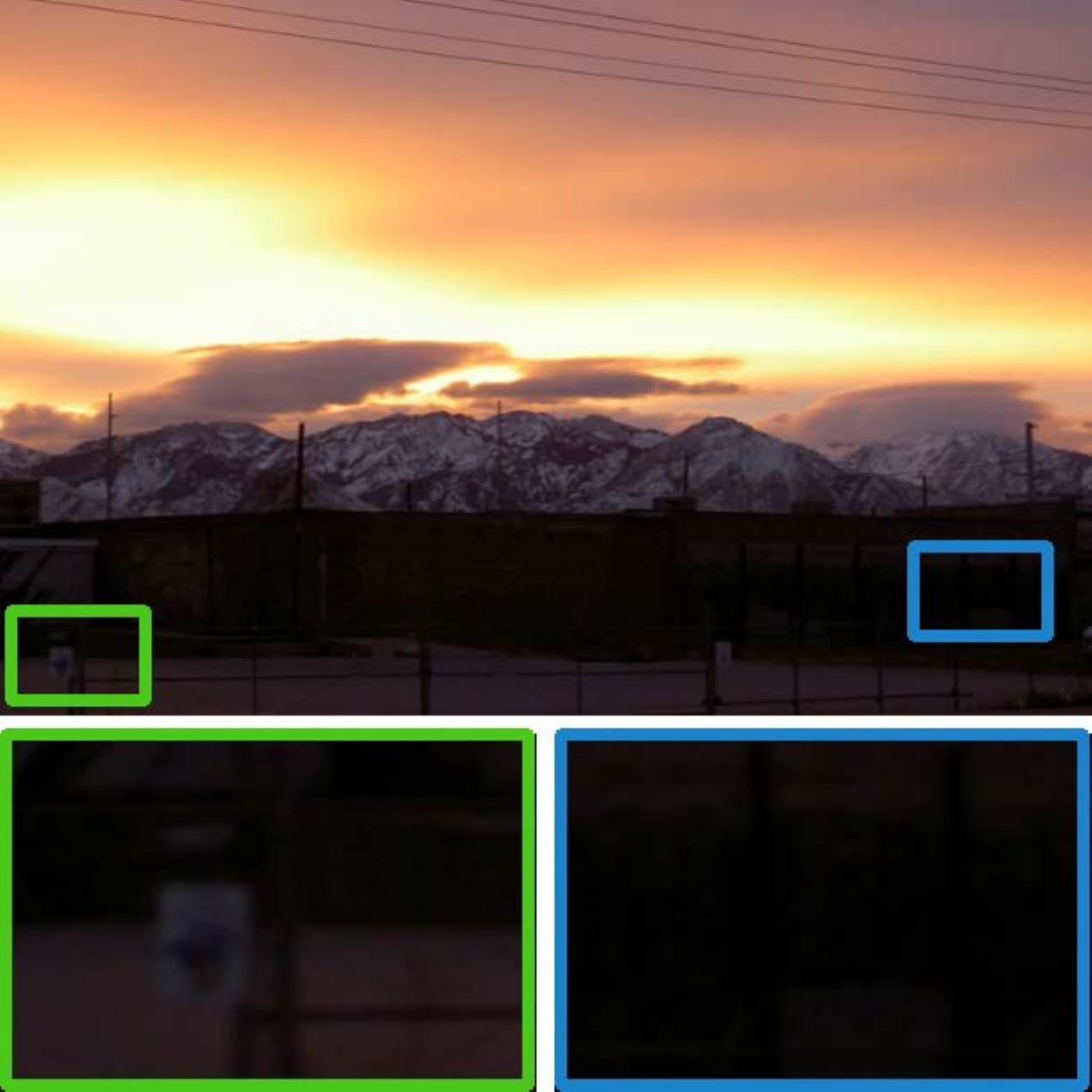}           &
    \includegraphics[width=\swten]{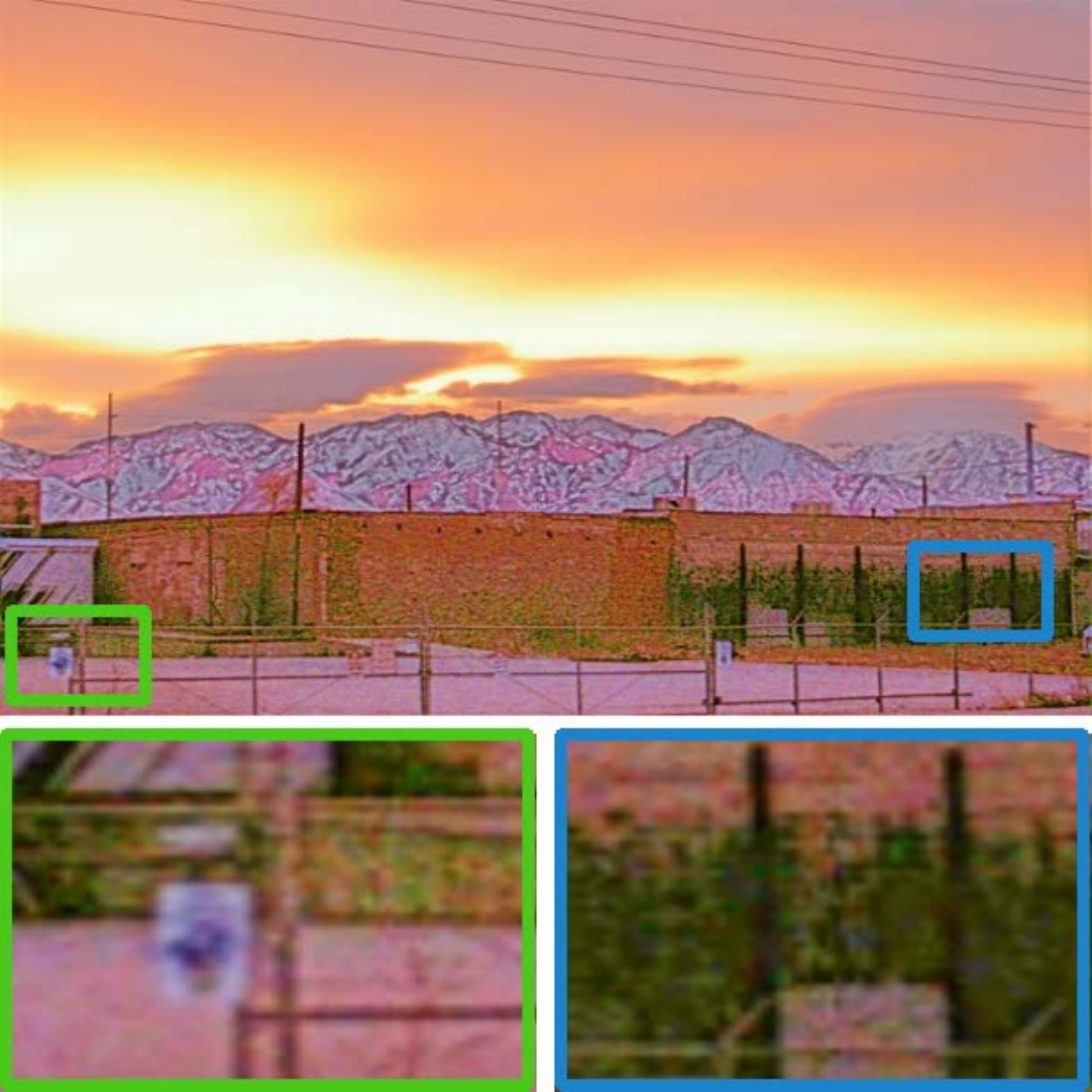}      &
    \includegraphics[width=\swten]{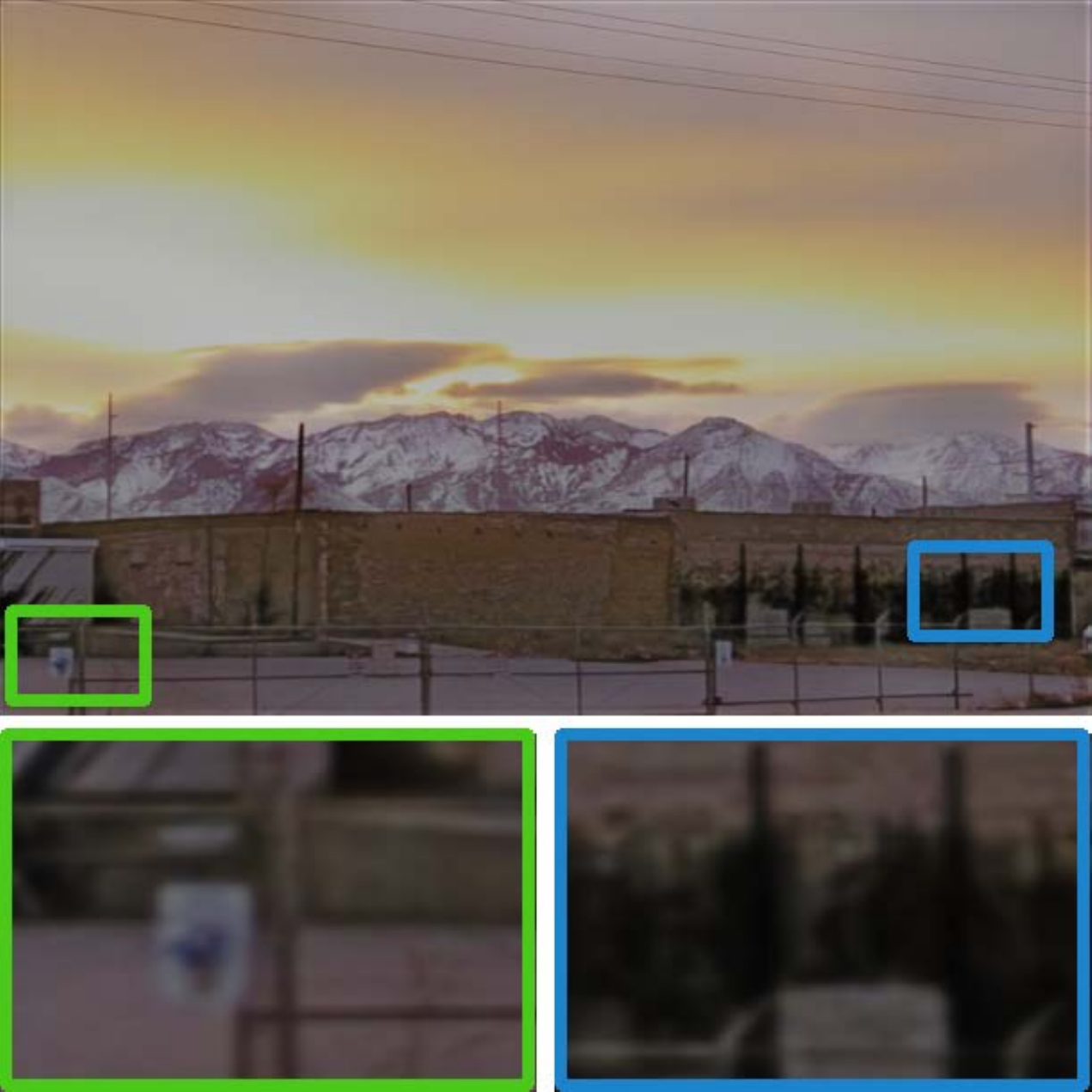}            &
    \includegraphics[width=\swten]{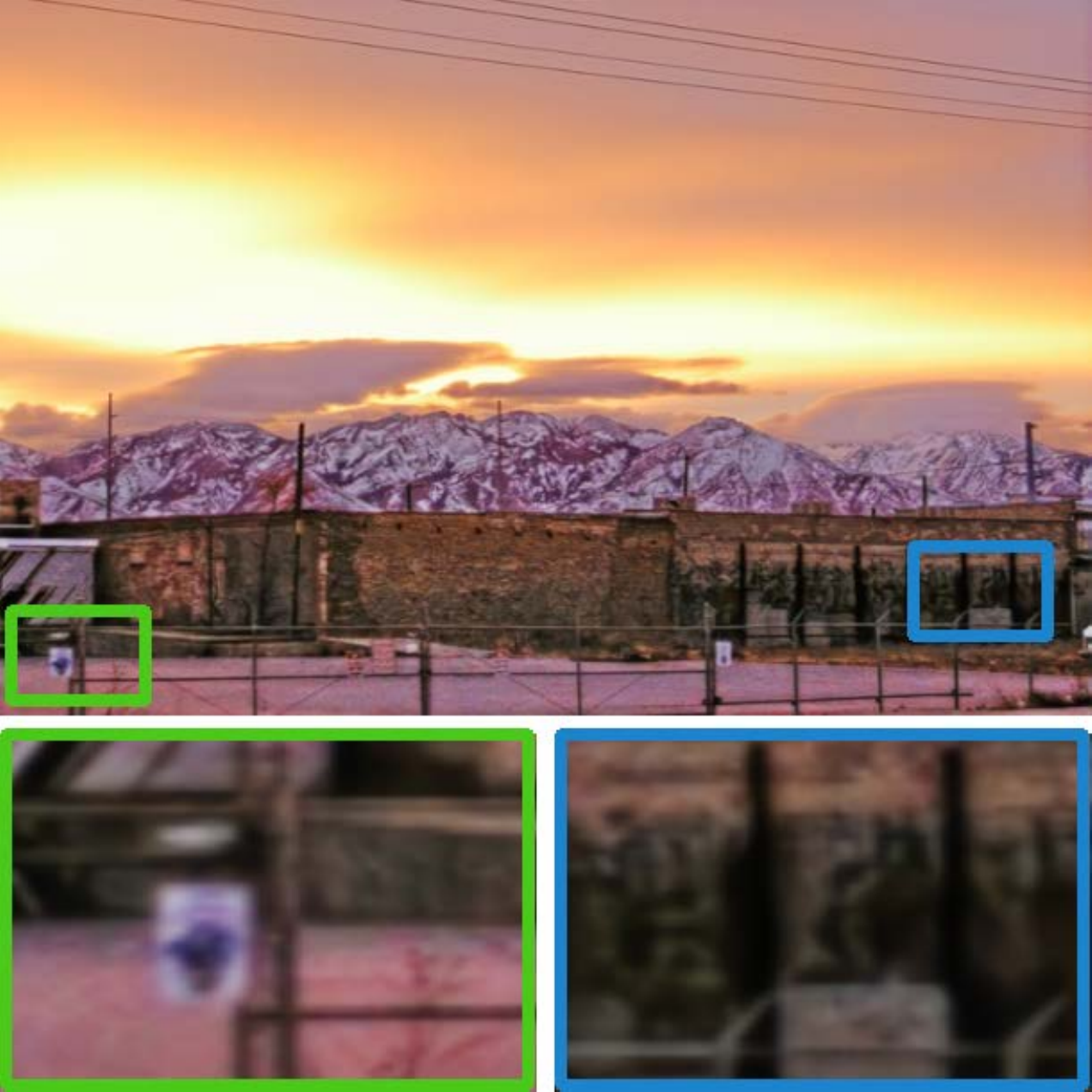}          &
    \includegraphics[width=\swten]{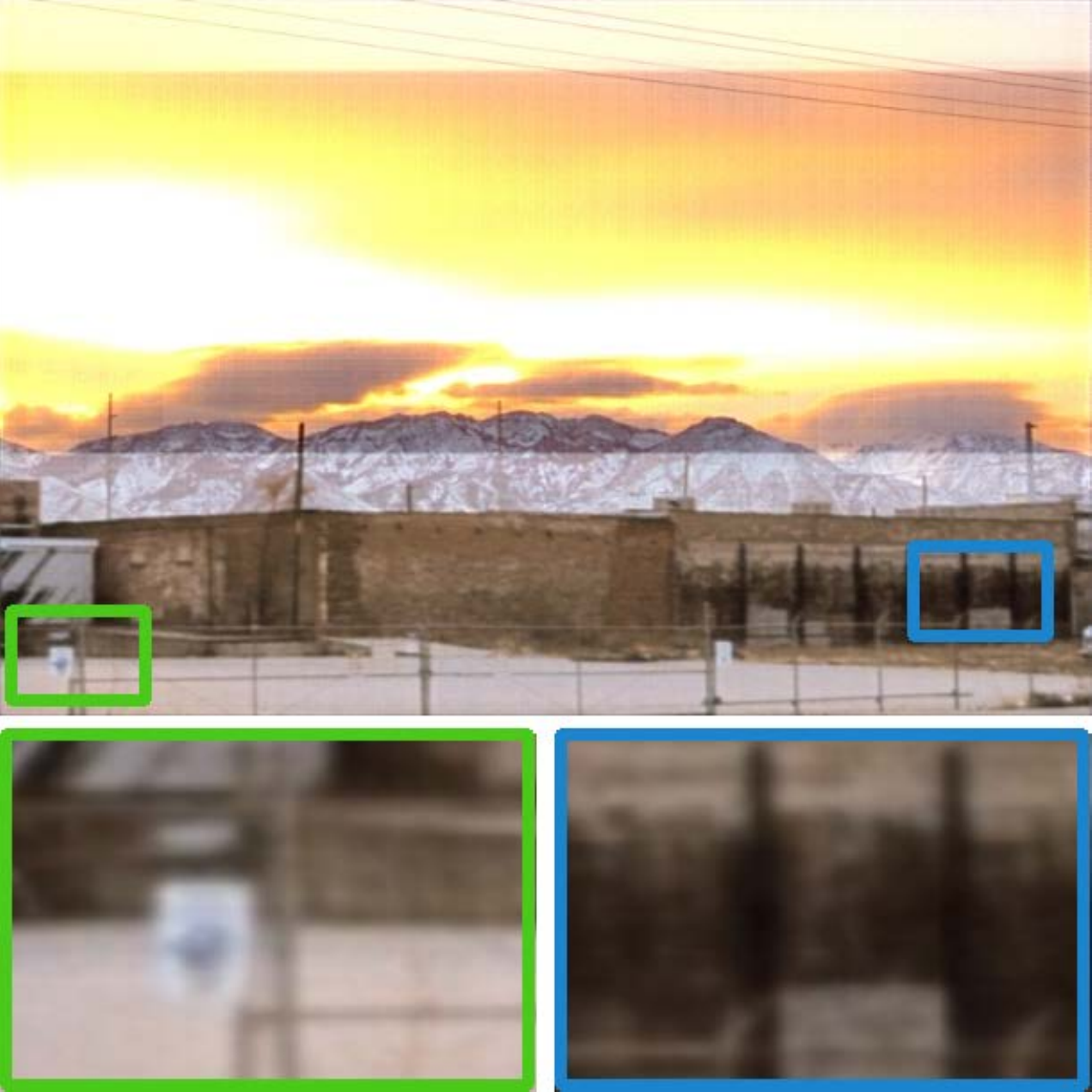}         &
    \includegraphics[width=\swten]{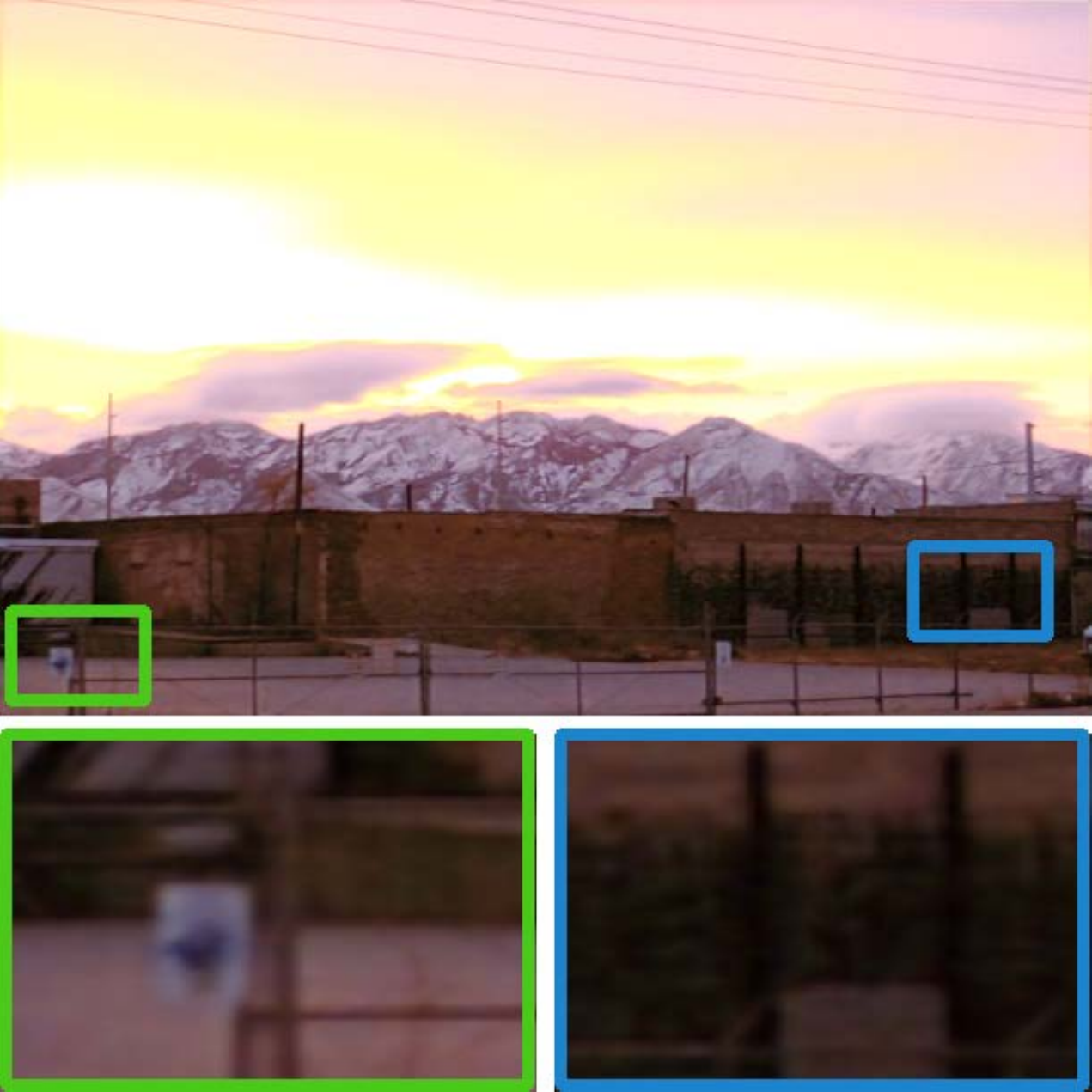}       &
    \includegraphics[width=\swten]{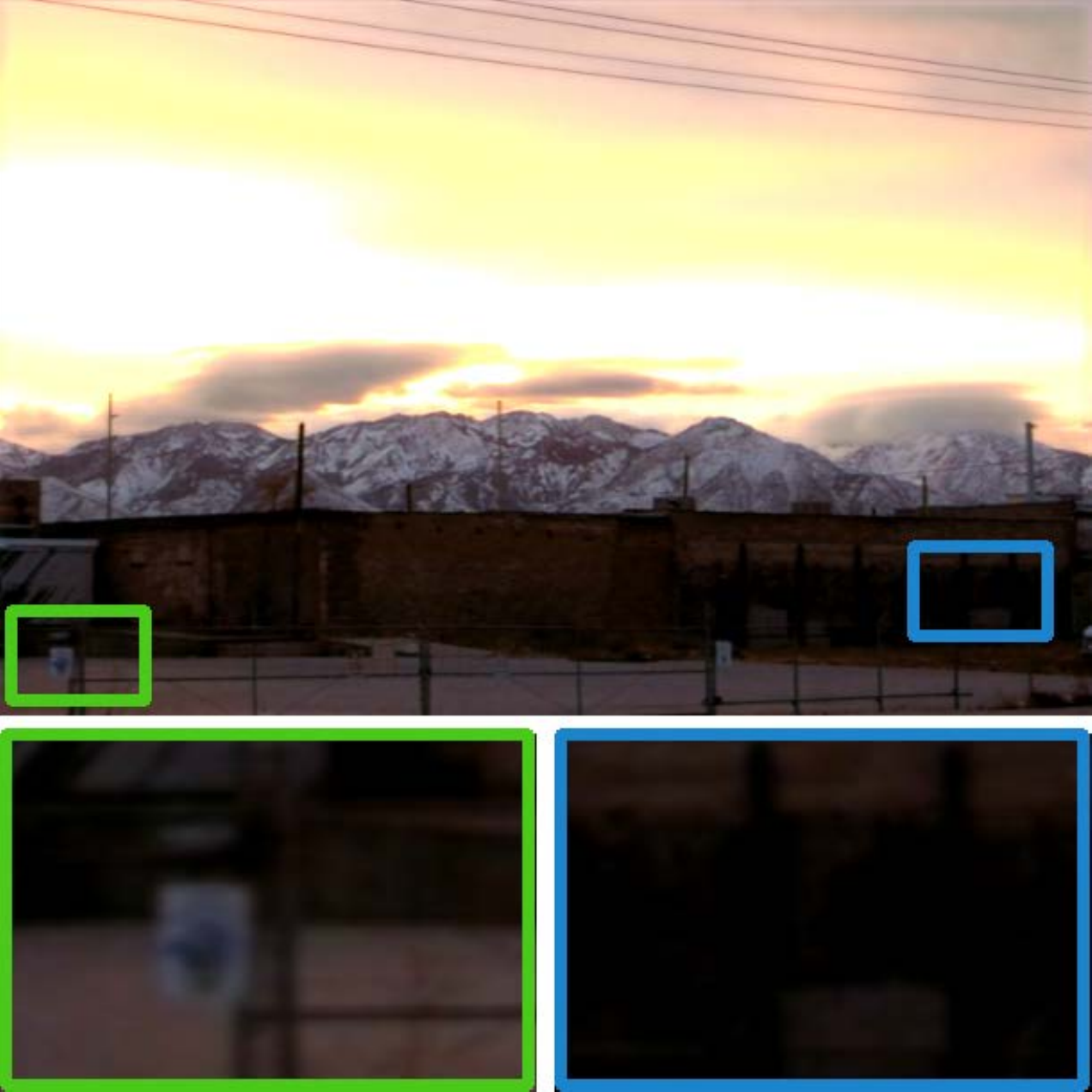}          &
    \includegraphics[width=\swten]{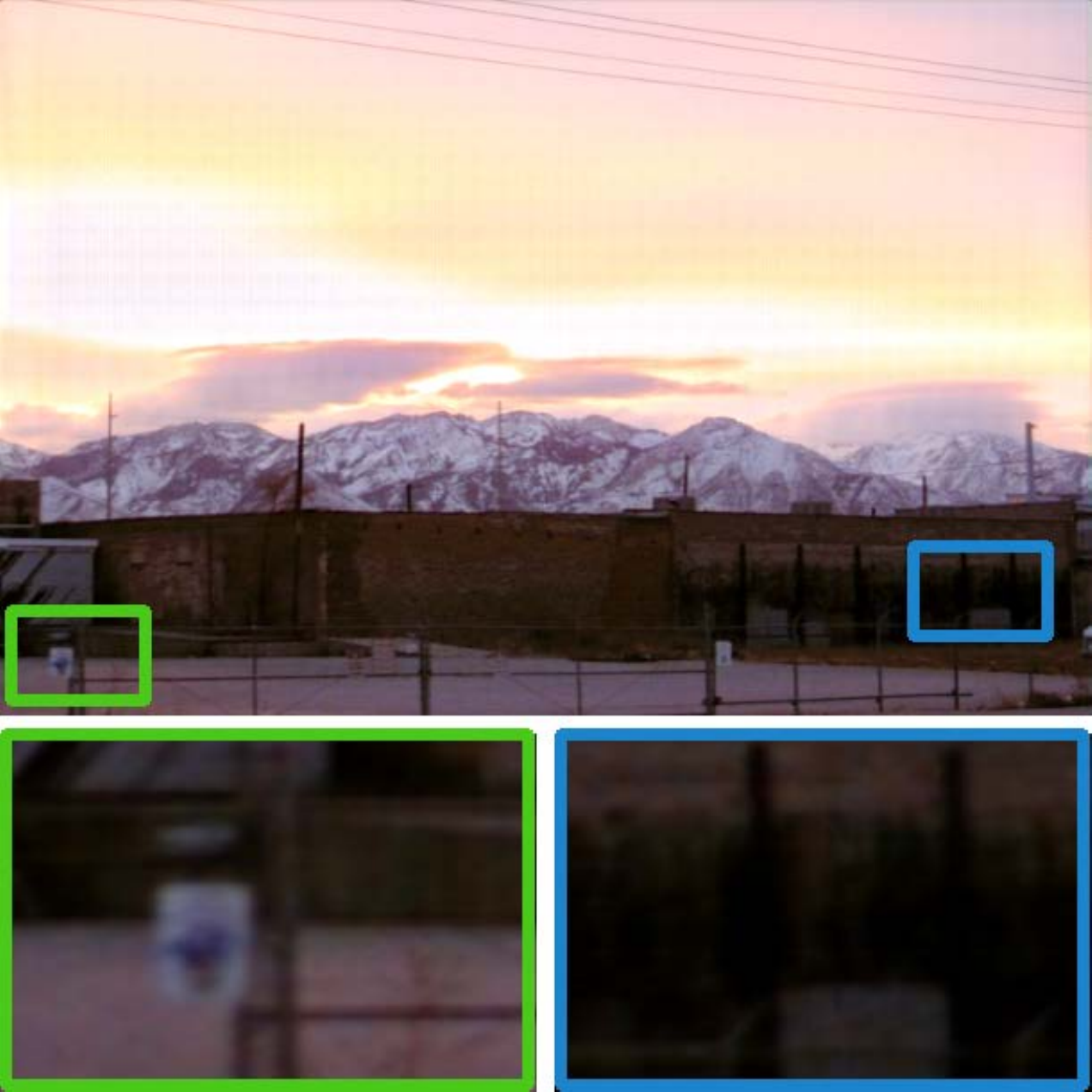}        &
    \includegraphics[width=\swten]{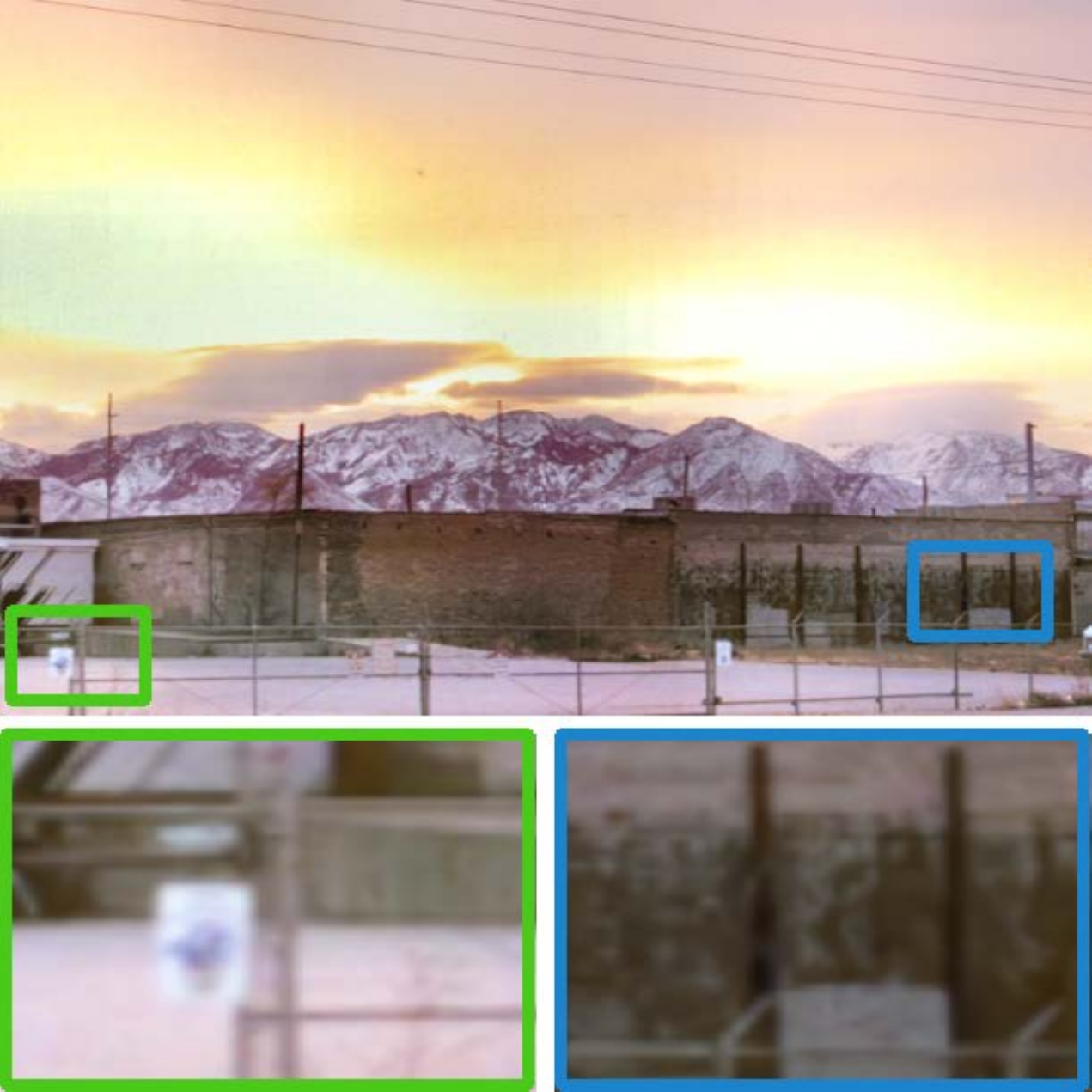}      \\
    \includegraphics[width=\swten]{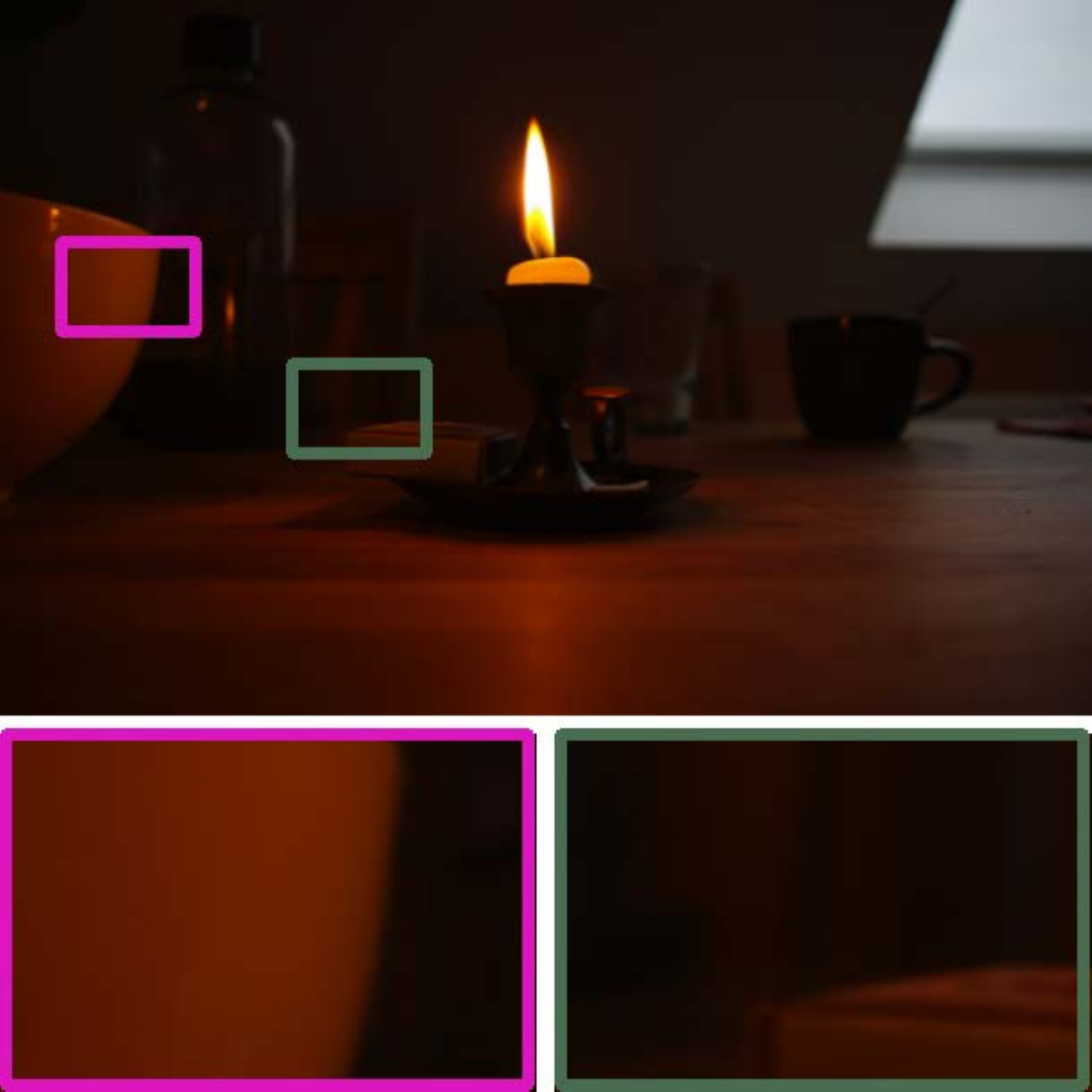}            &
    \includegraphics[width=\swten]{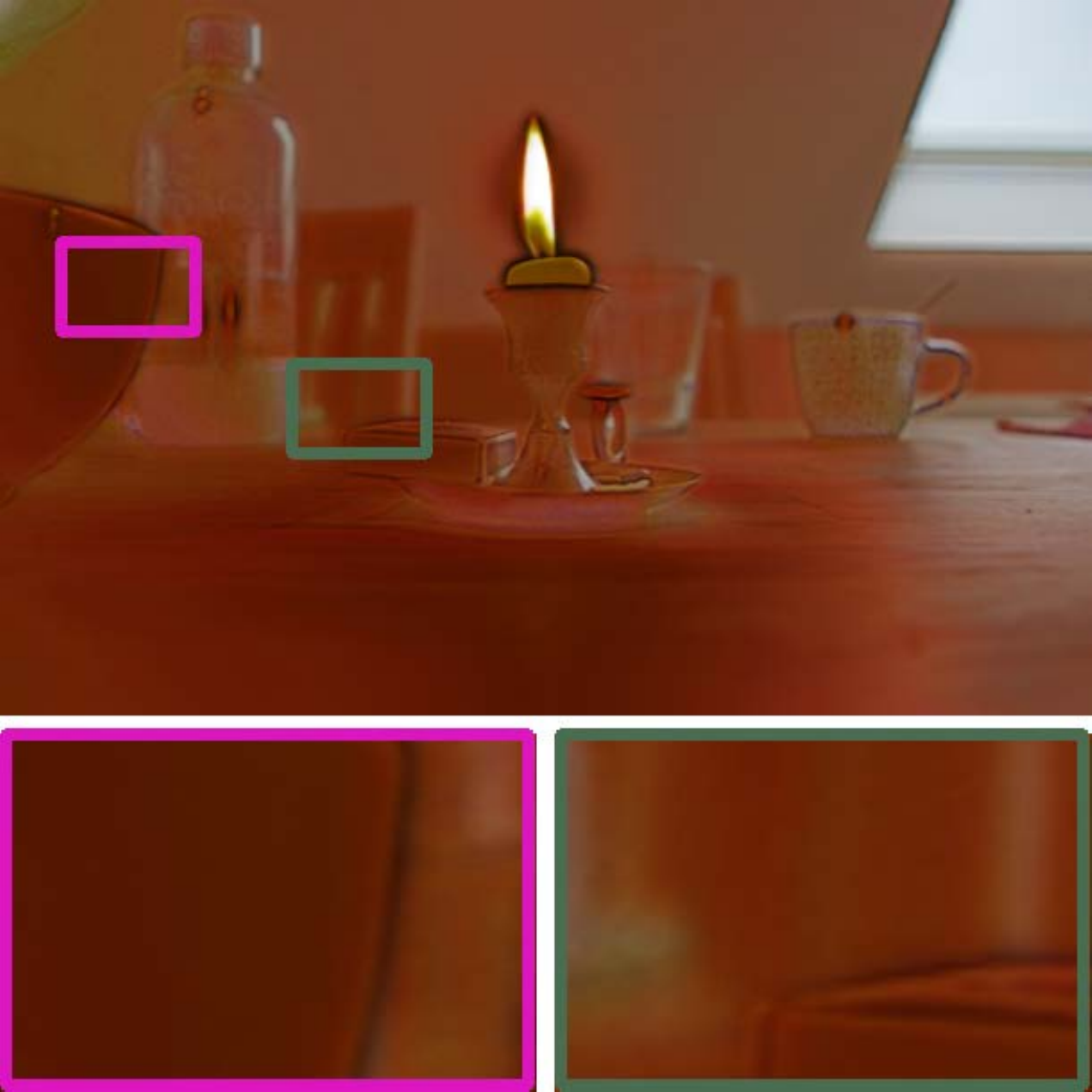}       &
    \includegraphics[width=\swten]{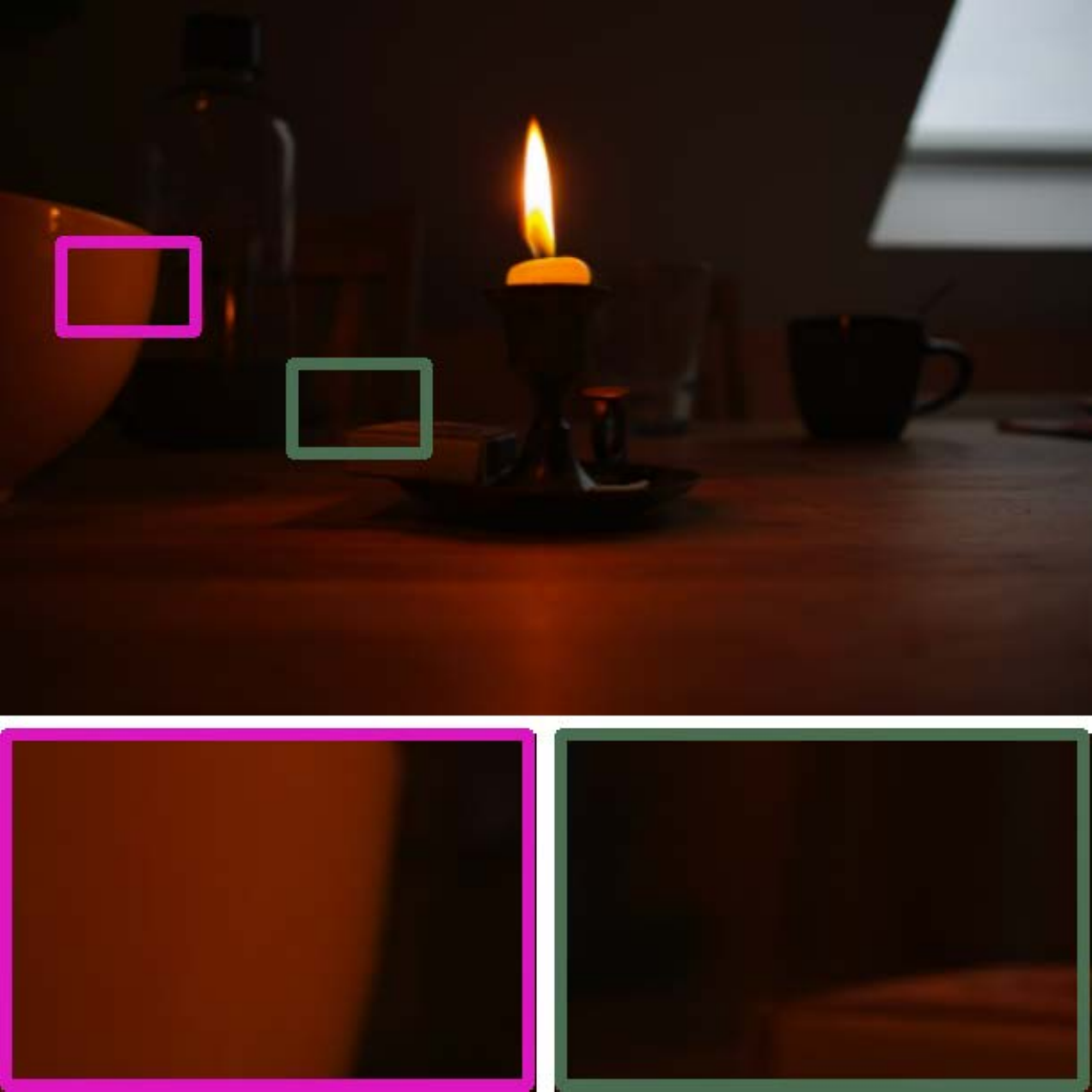}             &
    \includegraphics[width=\swten]{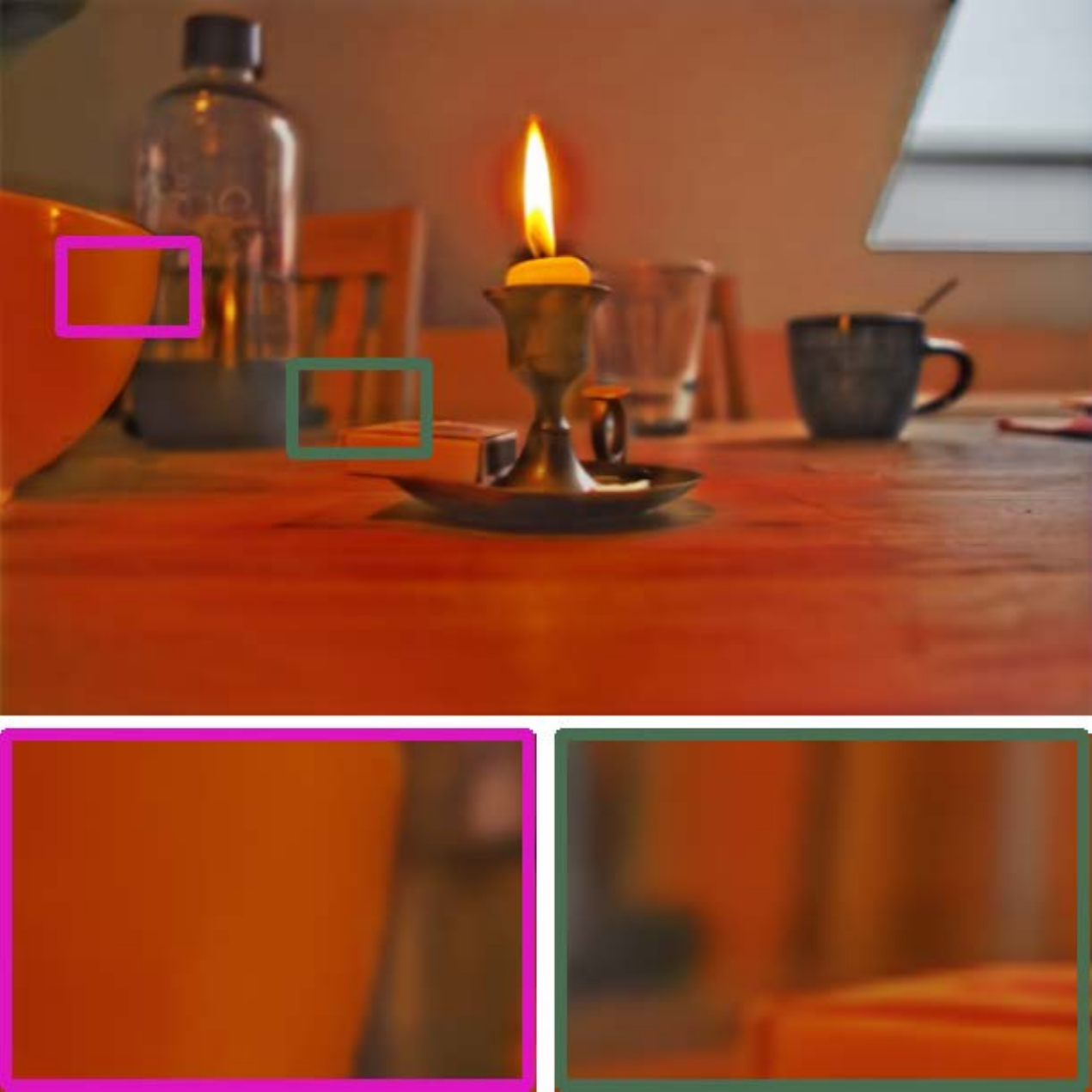}           &
    \includegraphics[width=\swten]{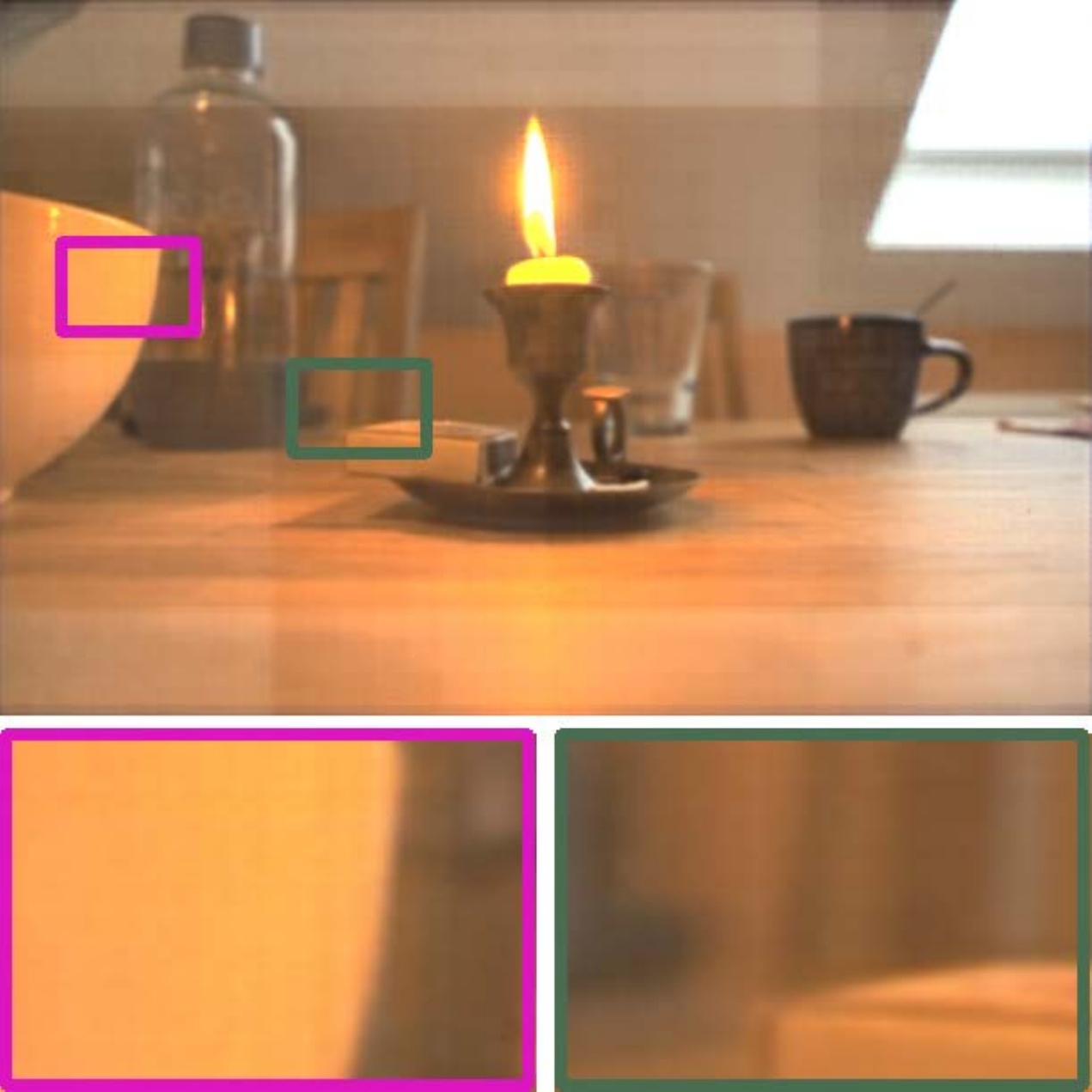}          &
    \includegraphics[width=\swten]{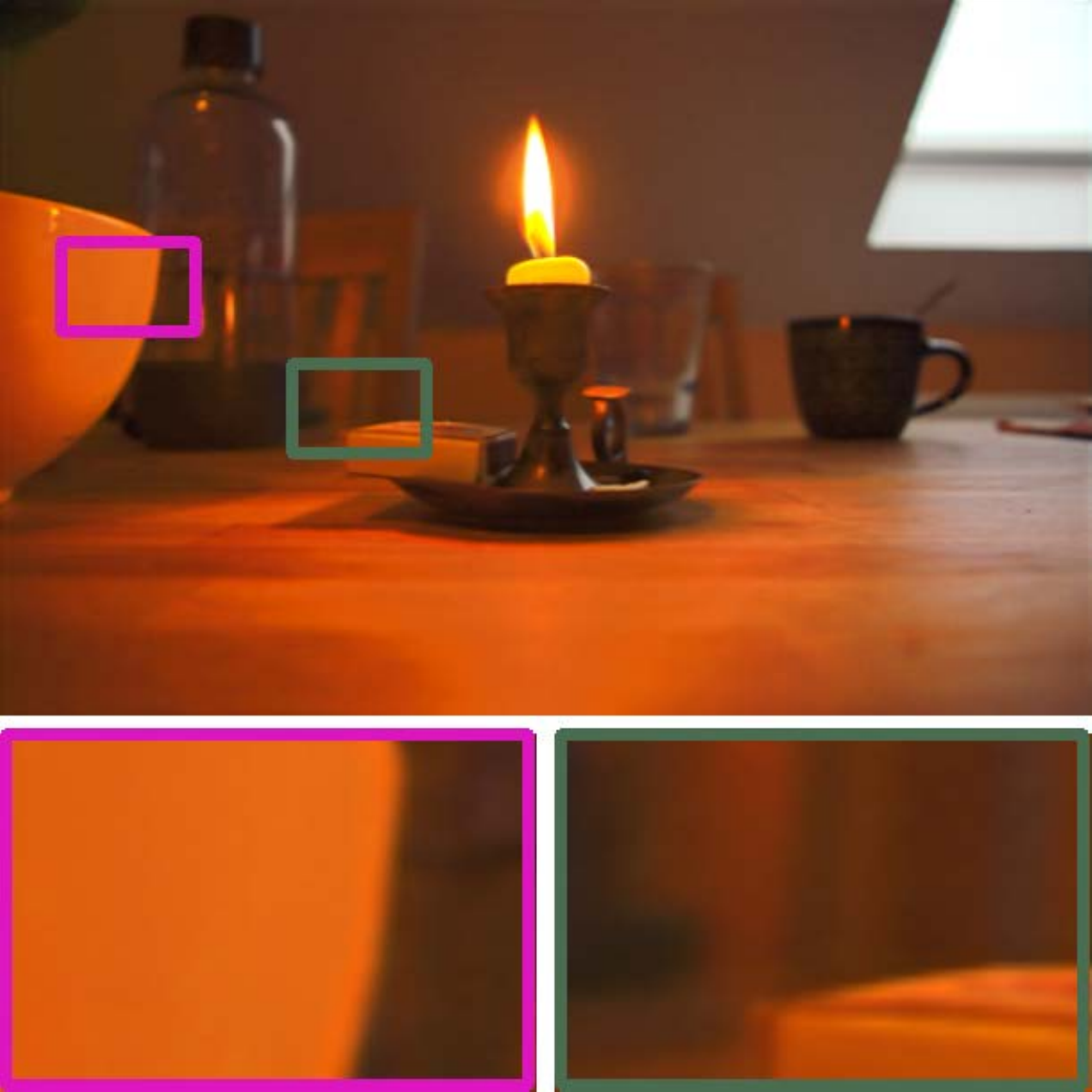}        &
    \includegraphics[width=\swten]{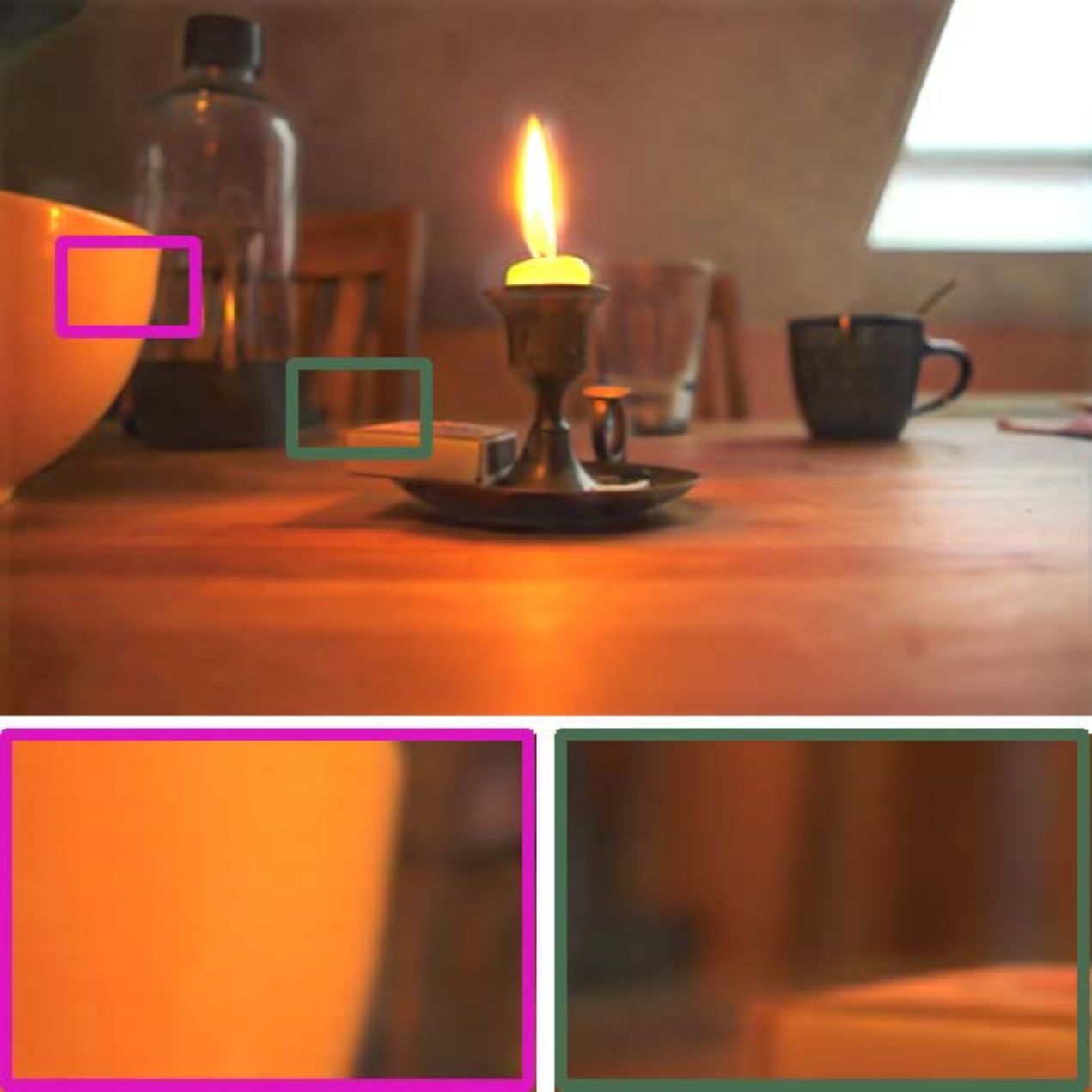}           &
    \includegraphics[width=\swten]{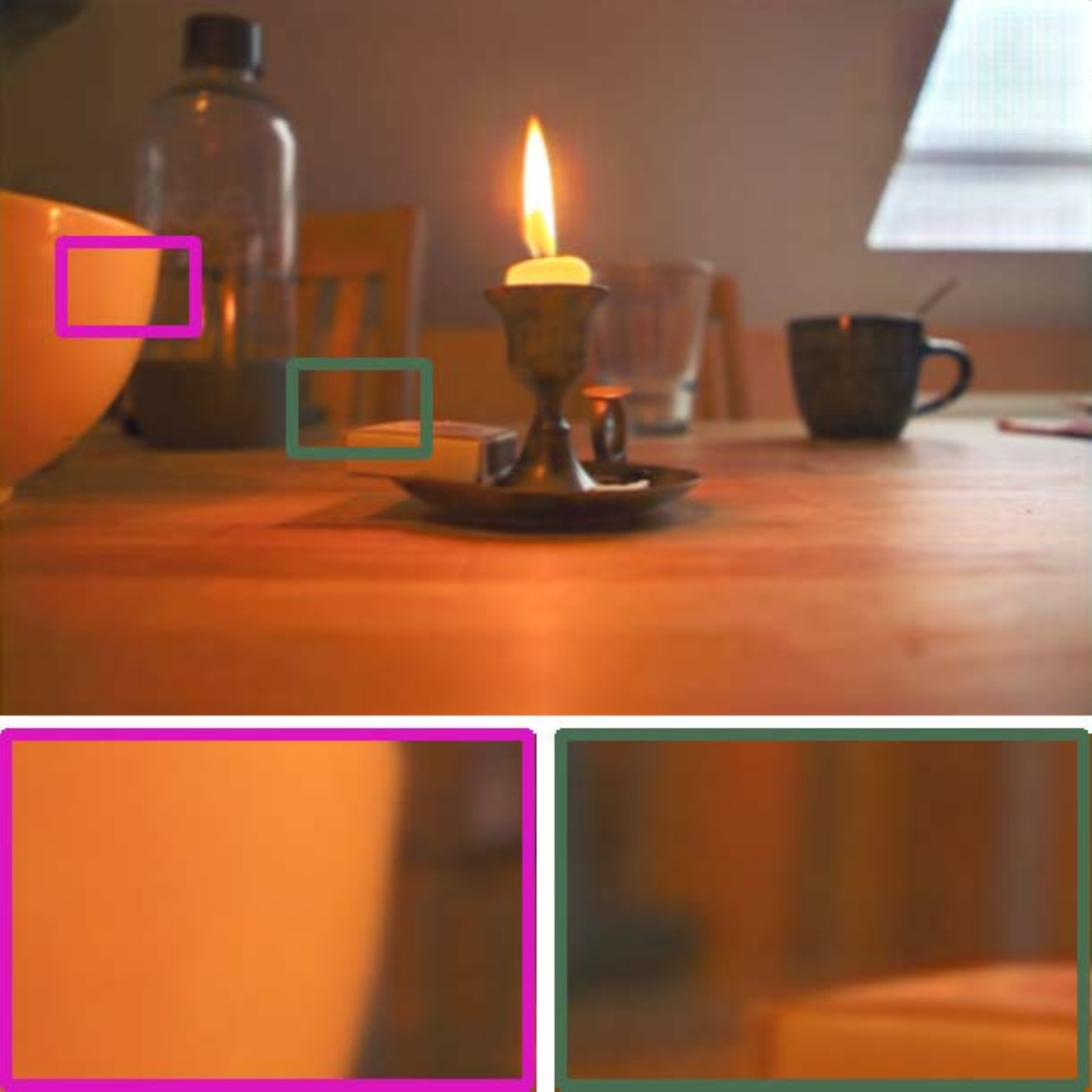}         &
    \includegraphics[width=\swten]{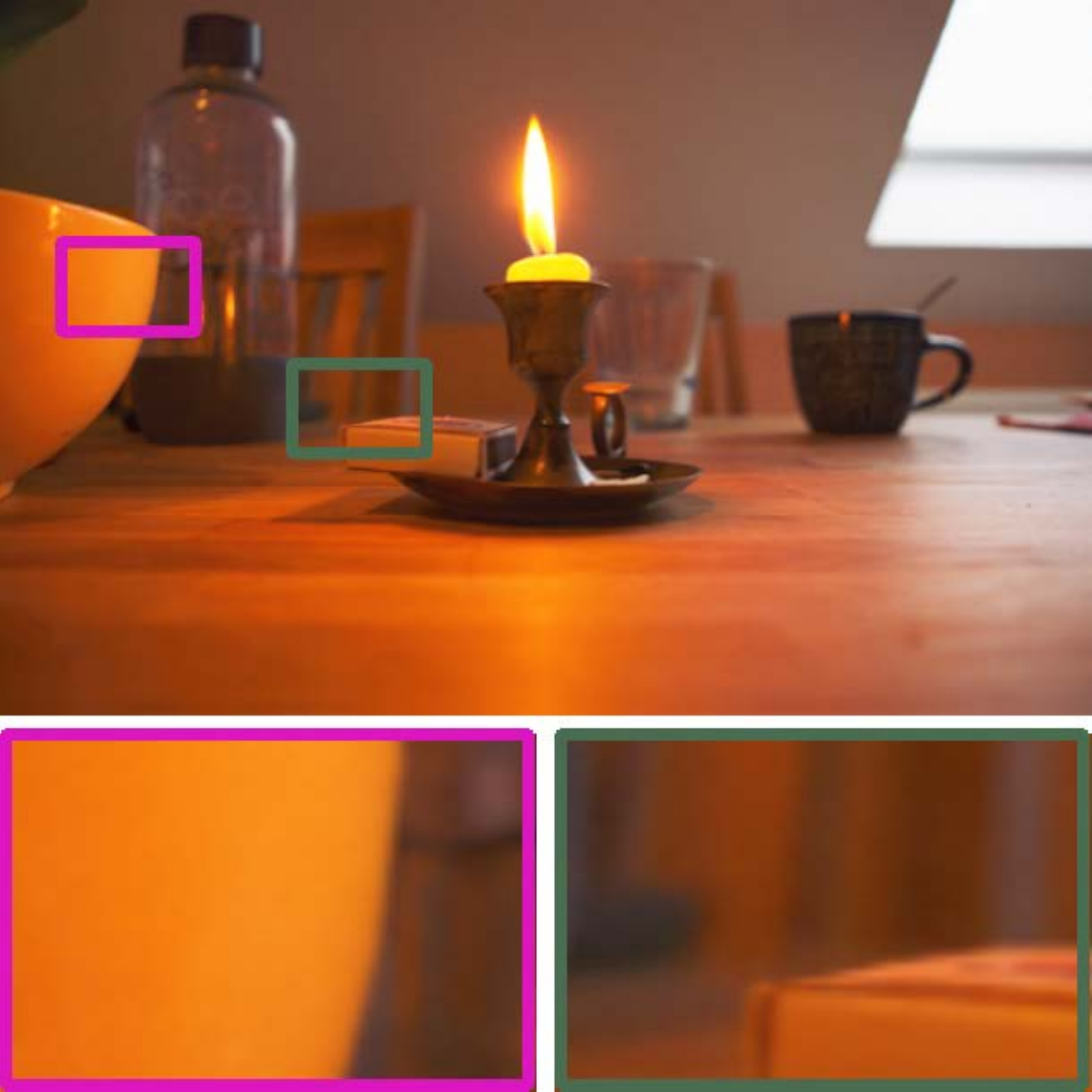}     \\
    \includegraphics[width=\swten]{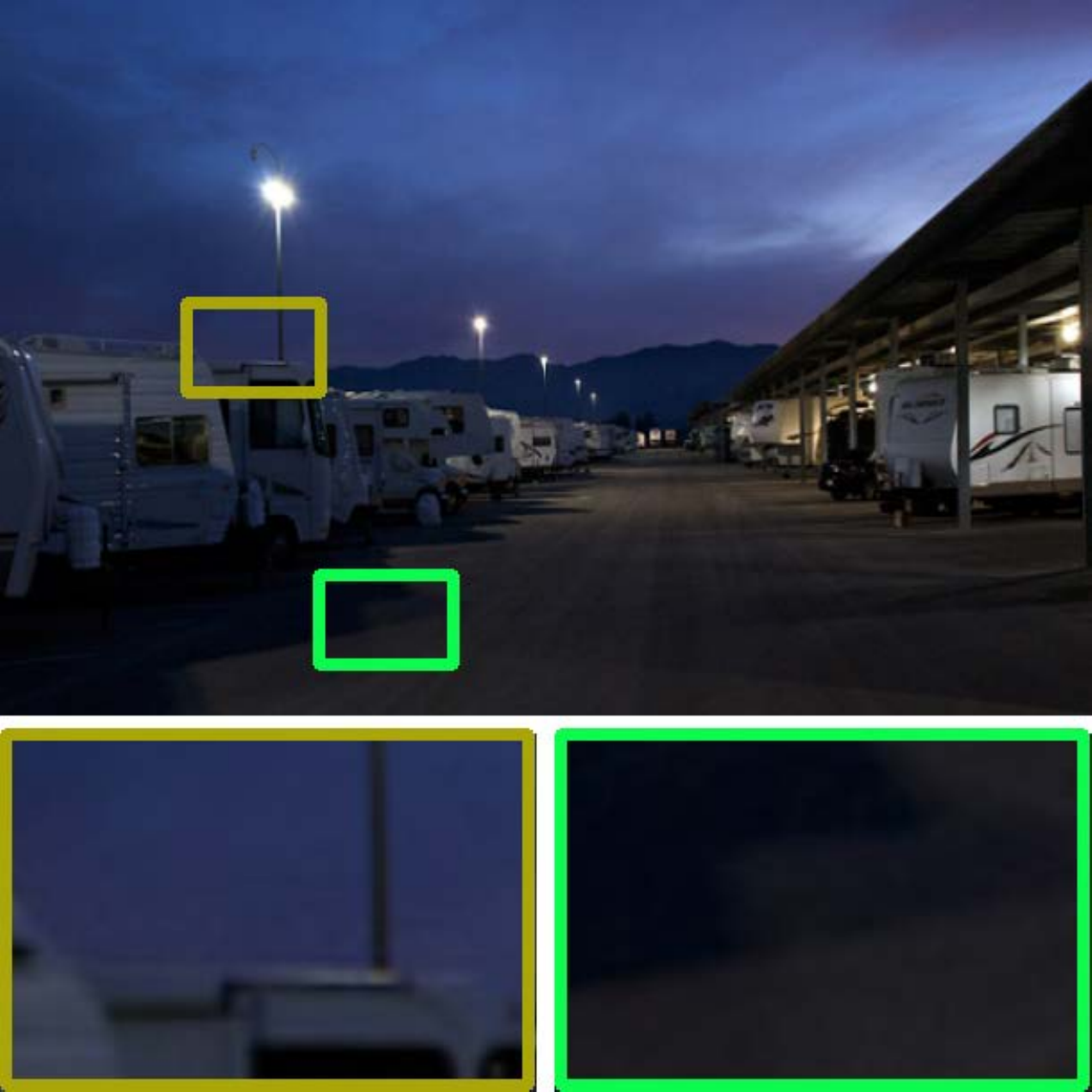}            &
    \includegraphics[width=\swten]{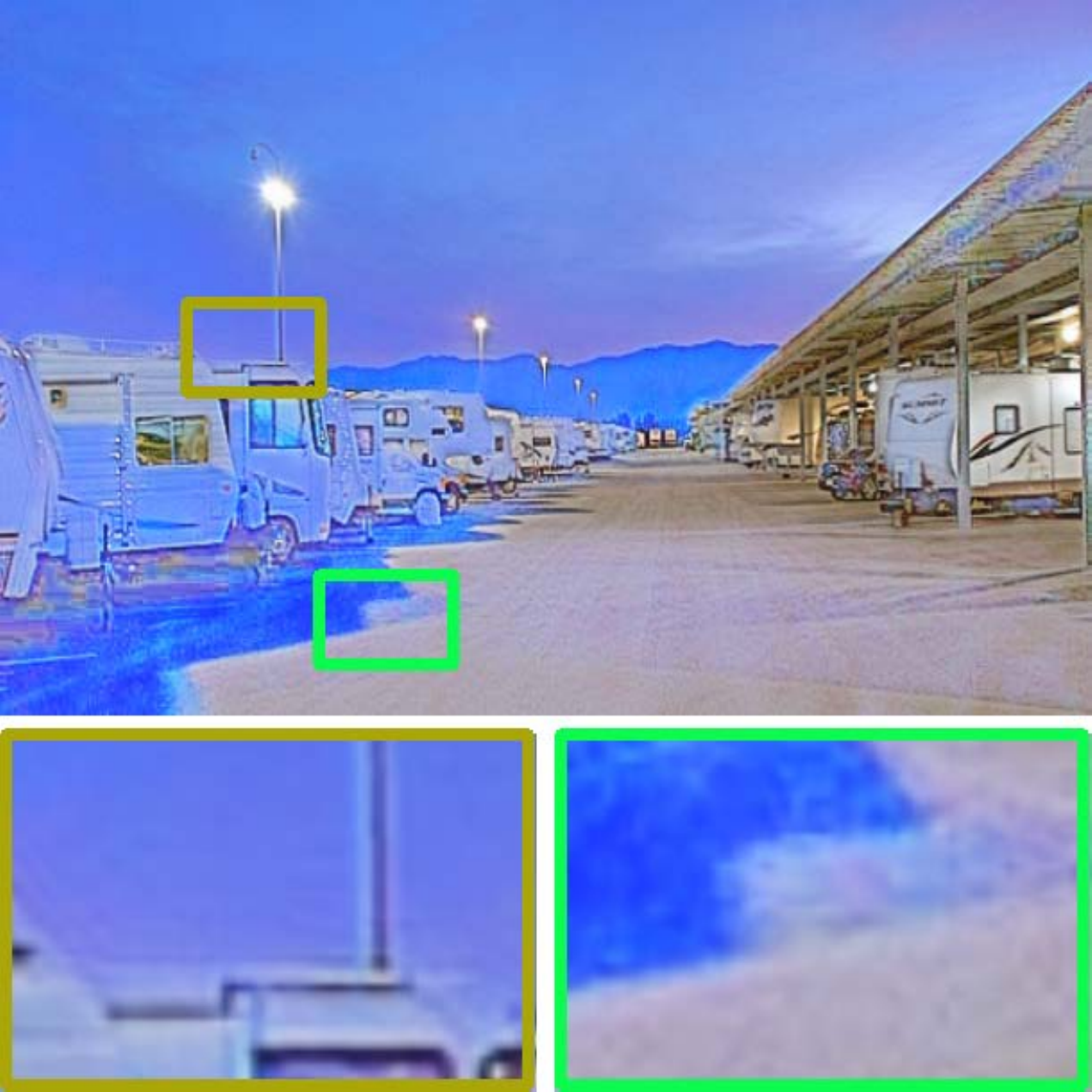}       &
    \includegraphics[width=\swten]{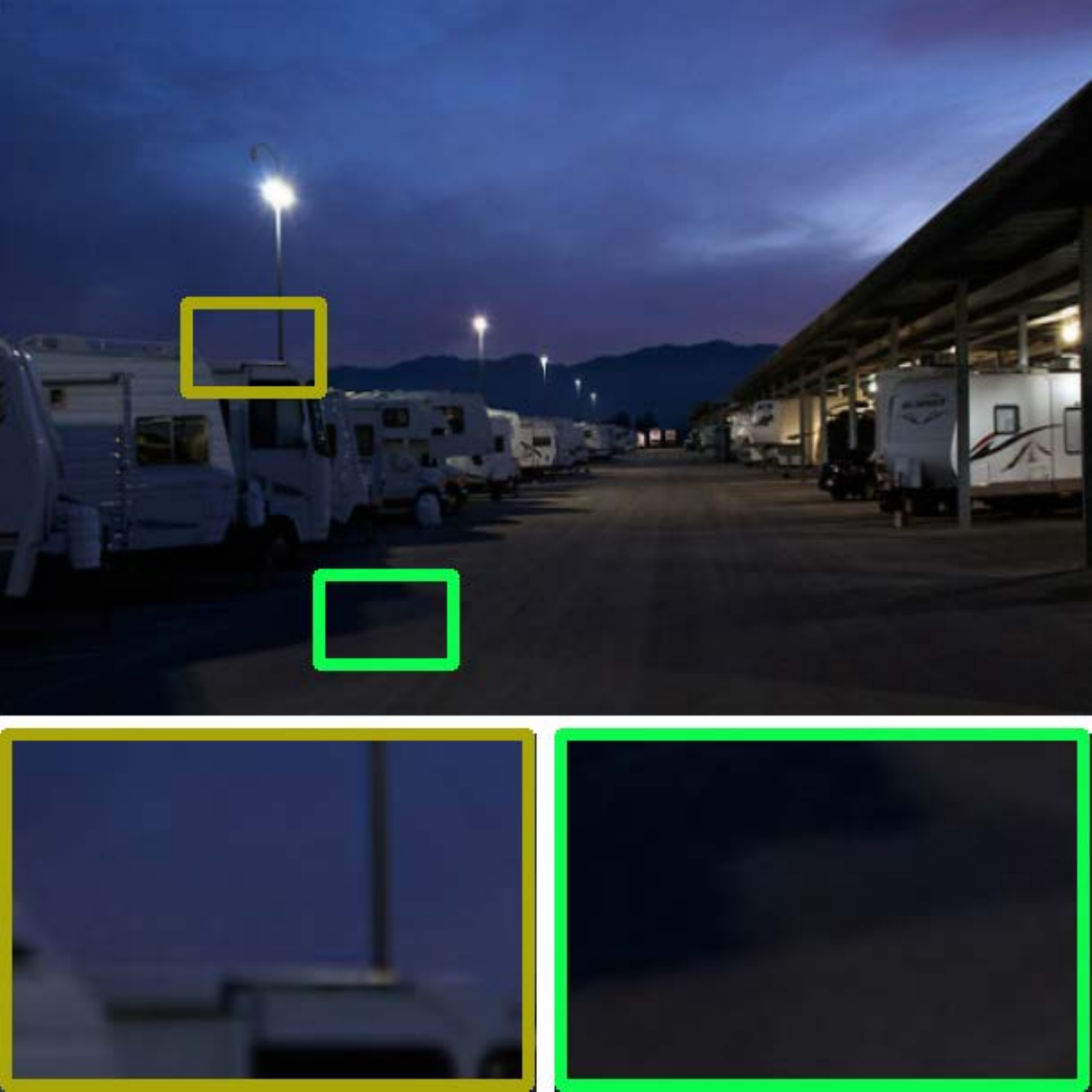}             &
    \includegraphics[width=\swten]{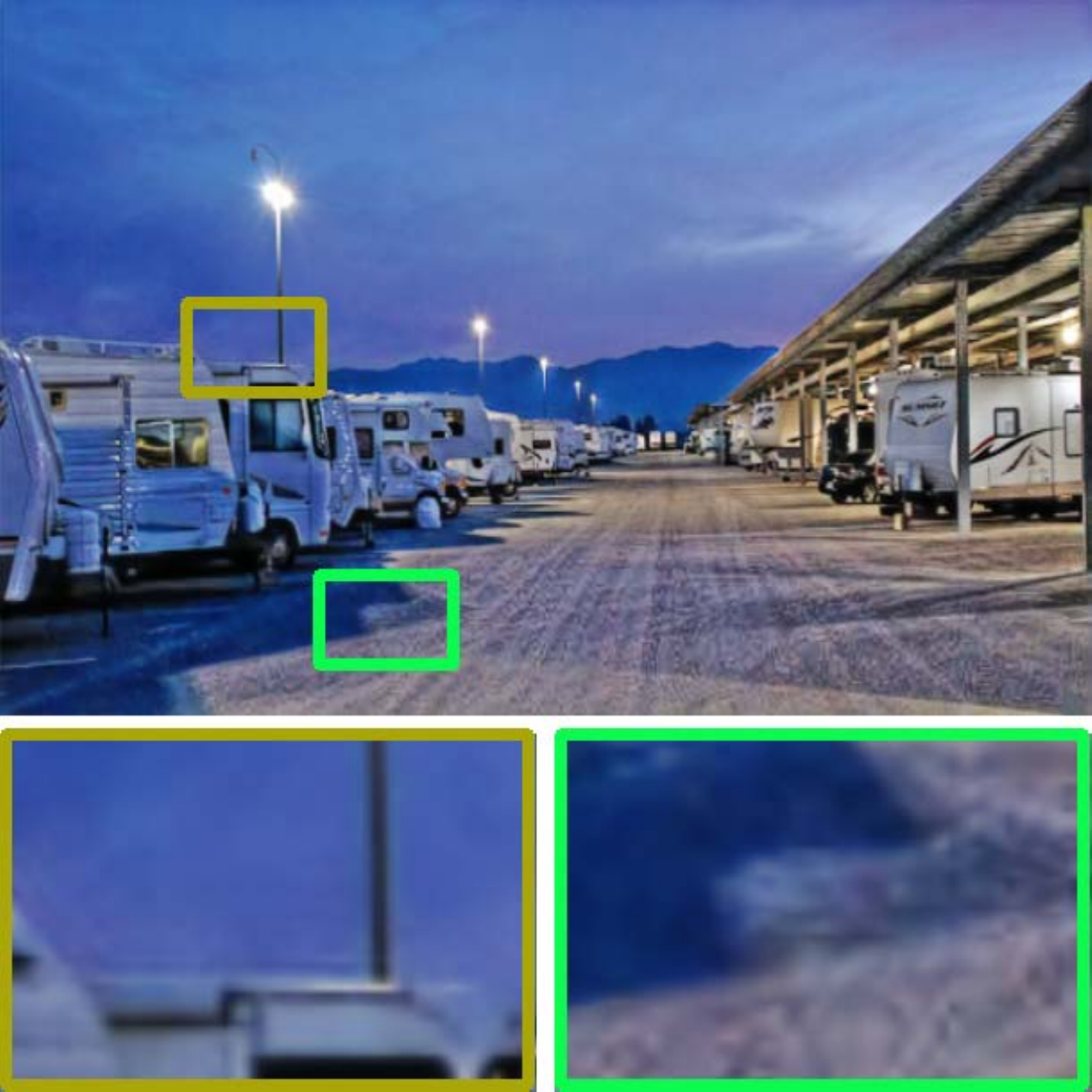}           &
    \includegraphics[width=\swten]{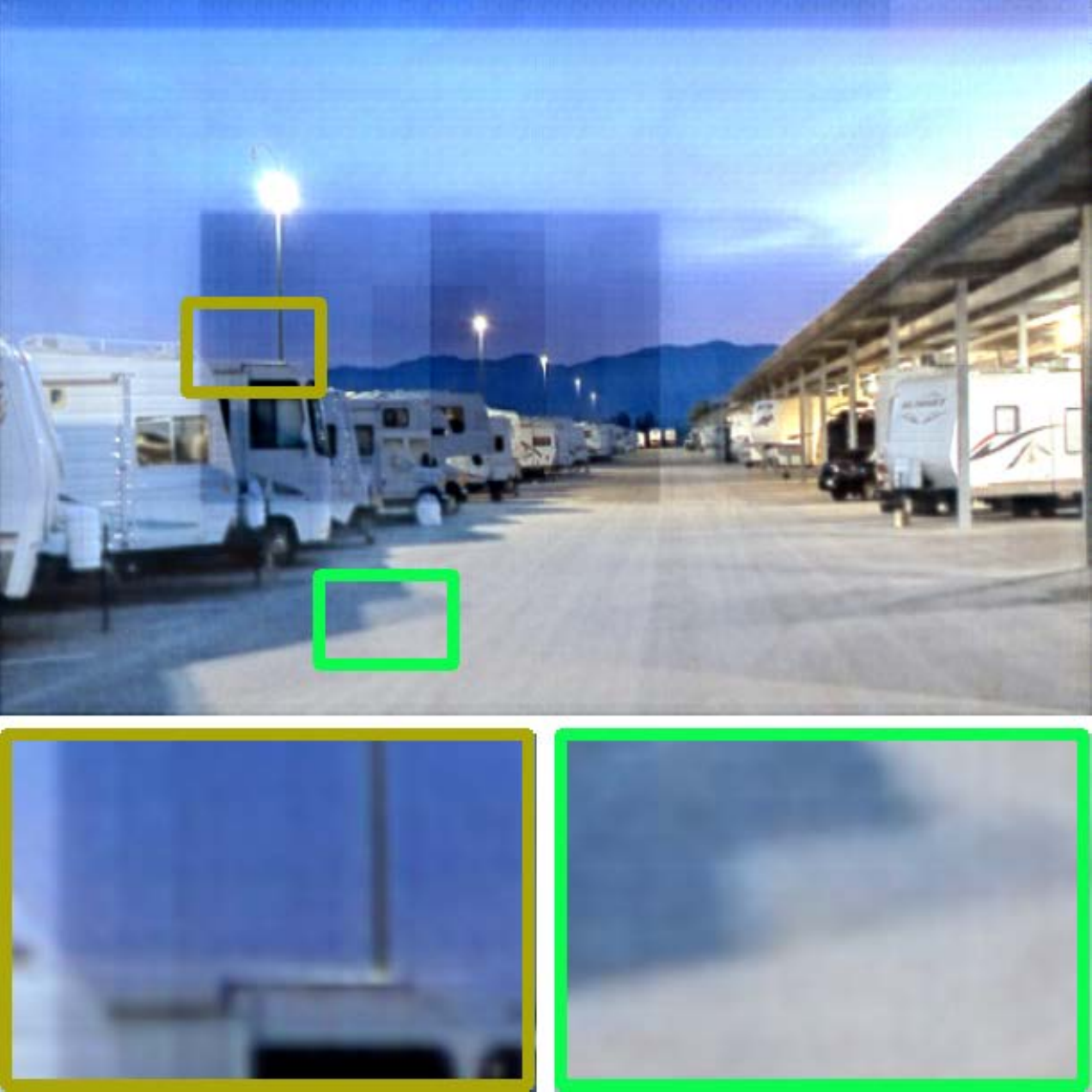}          &
    \includegraphics[width=\swten]{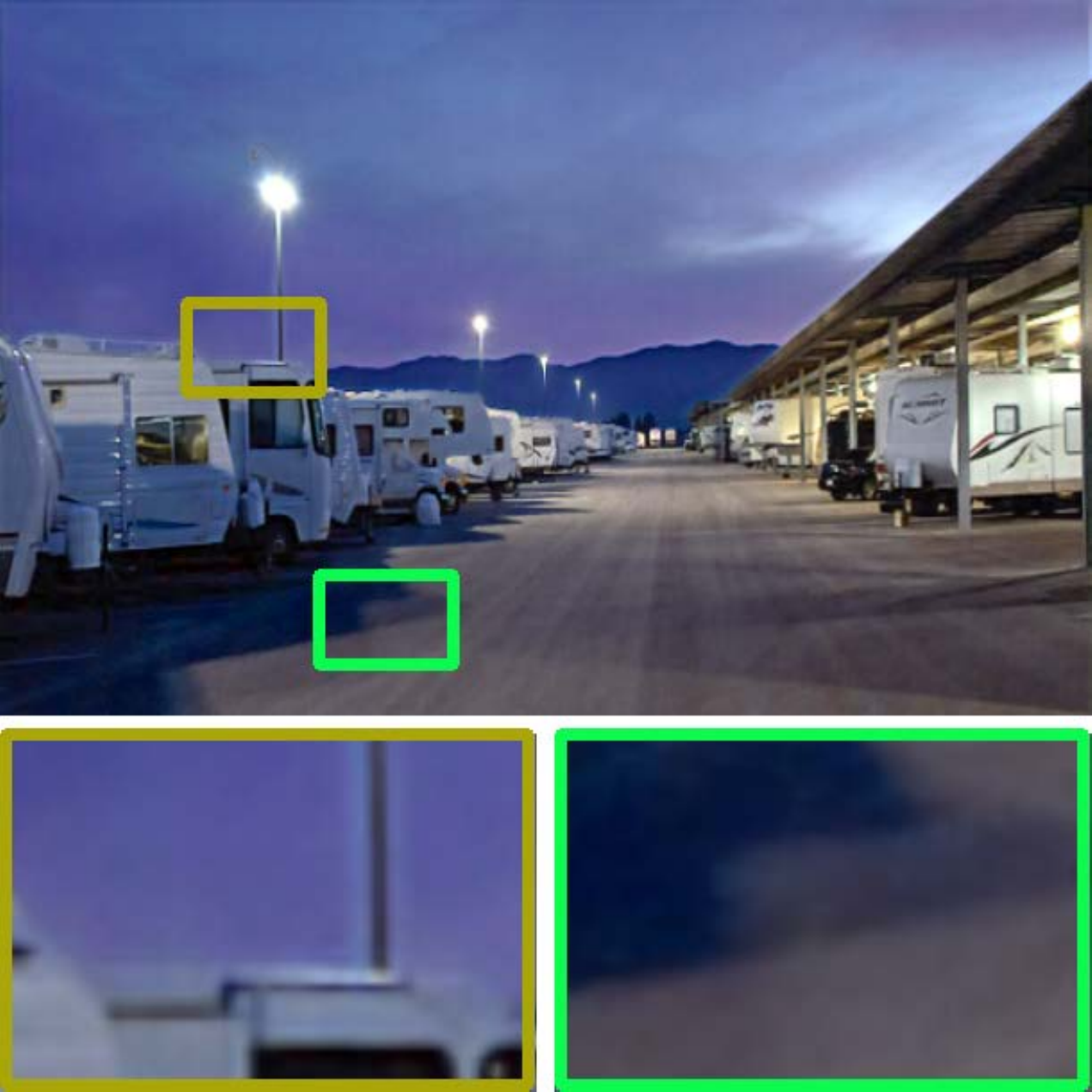}        &
    \includegraphics[width=\swten]{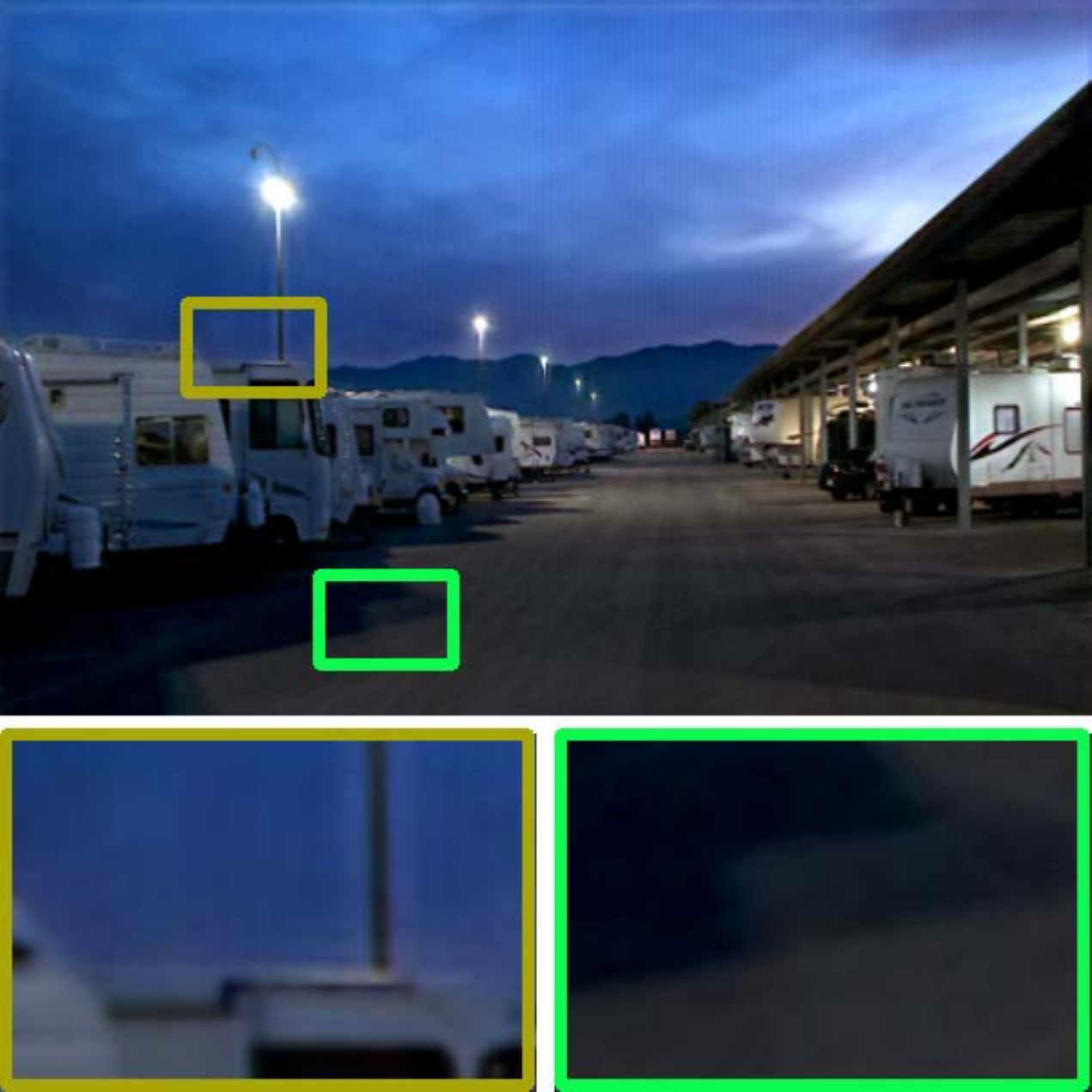}           &
    \includegraphics[width=\swten]{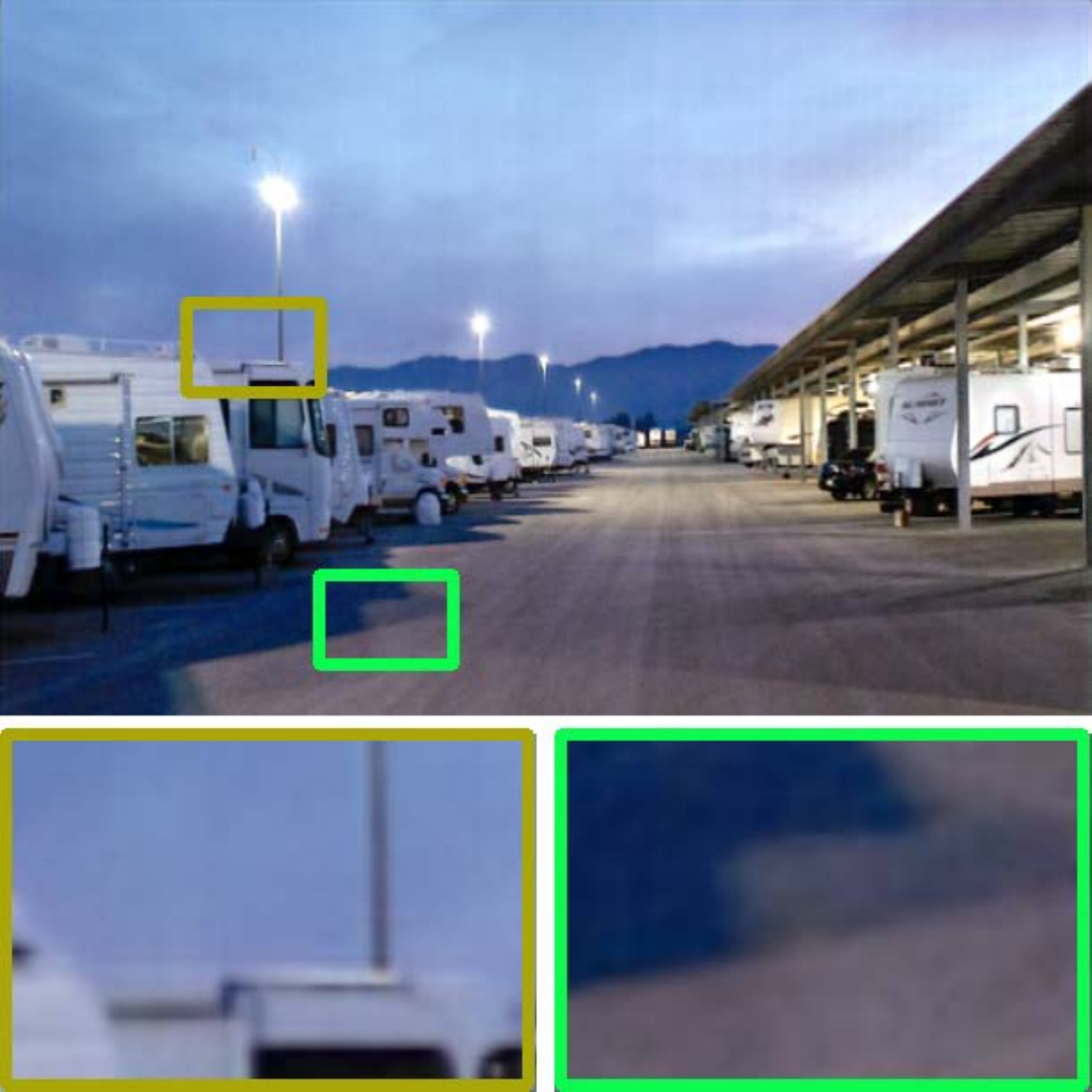}         &
    \includegraphics[width=\swten]{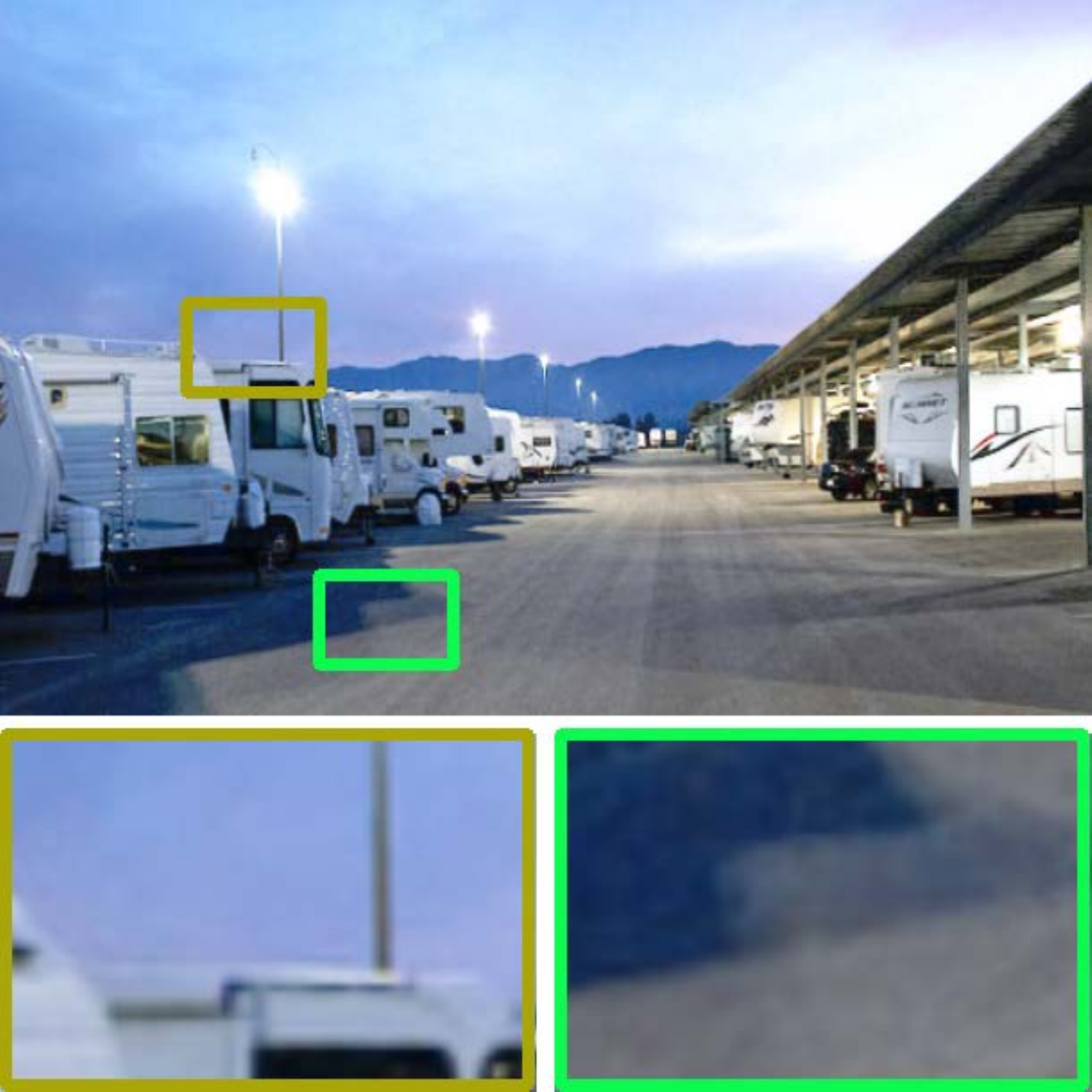}        \\
    \includegraphics[width=\swten]{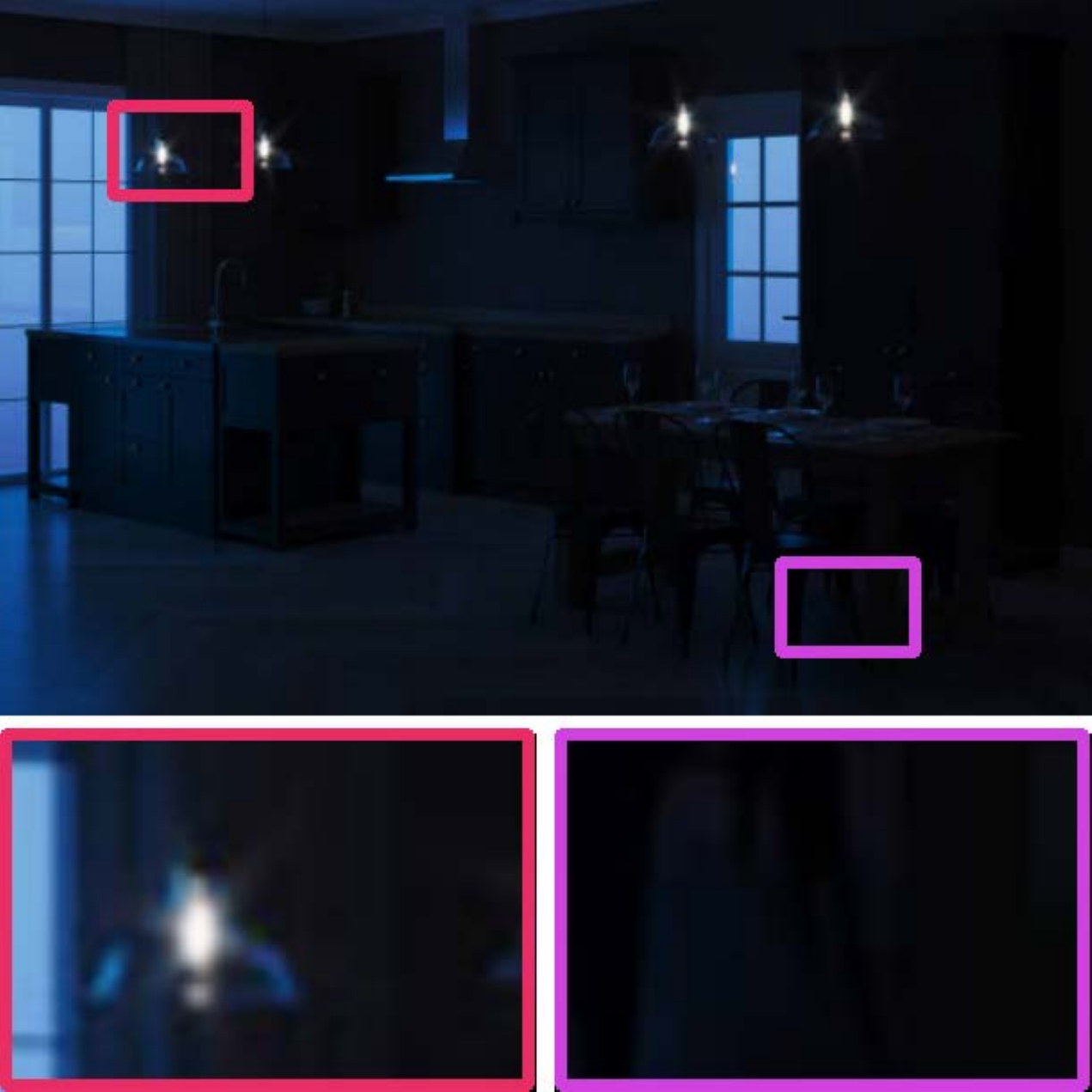}      &
    \includegraphics[width=\swten]{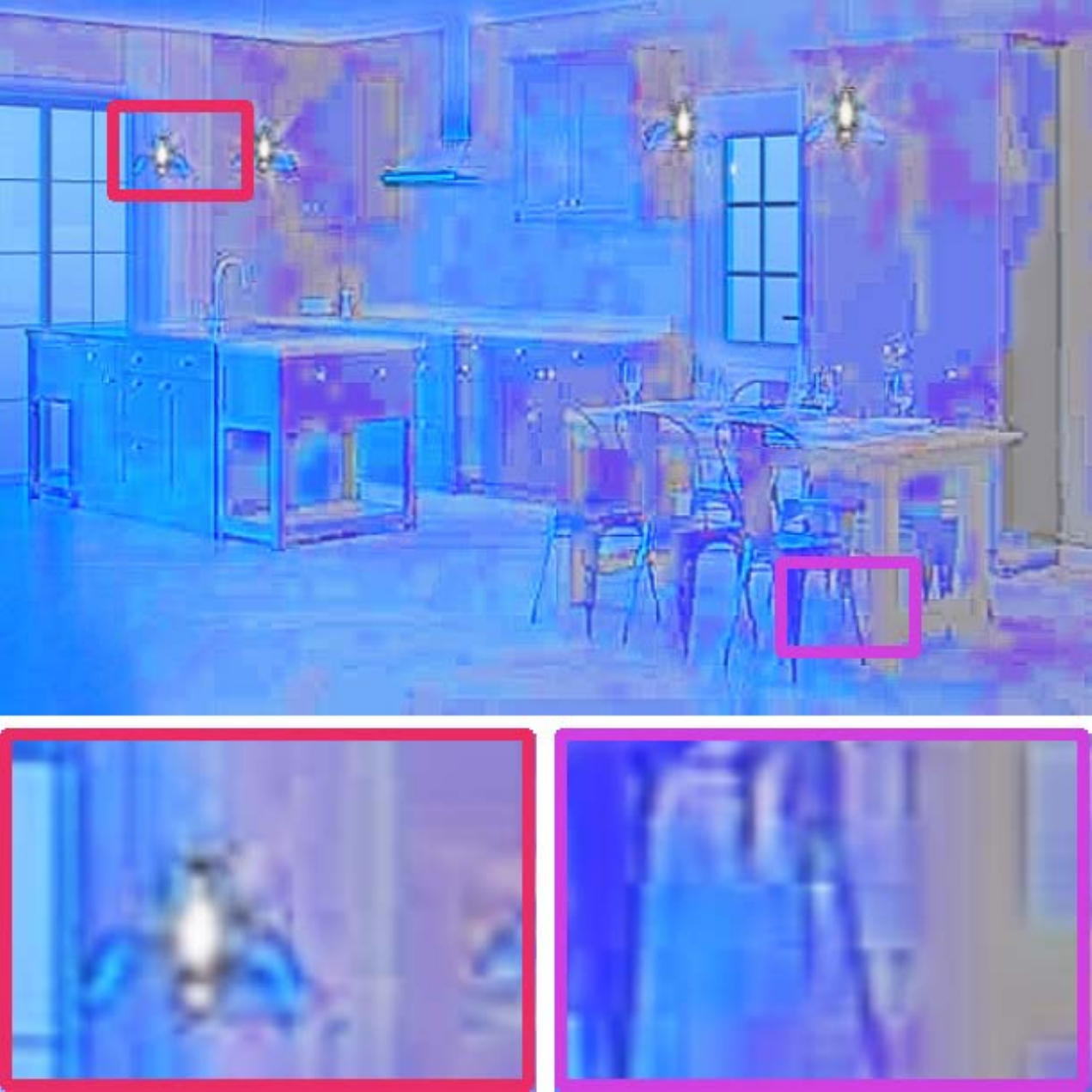} &
    \includegraphics[width=\swten]{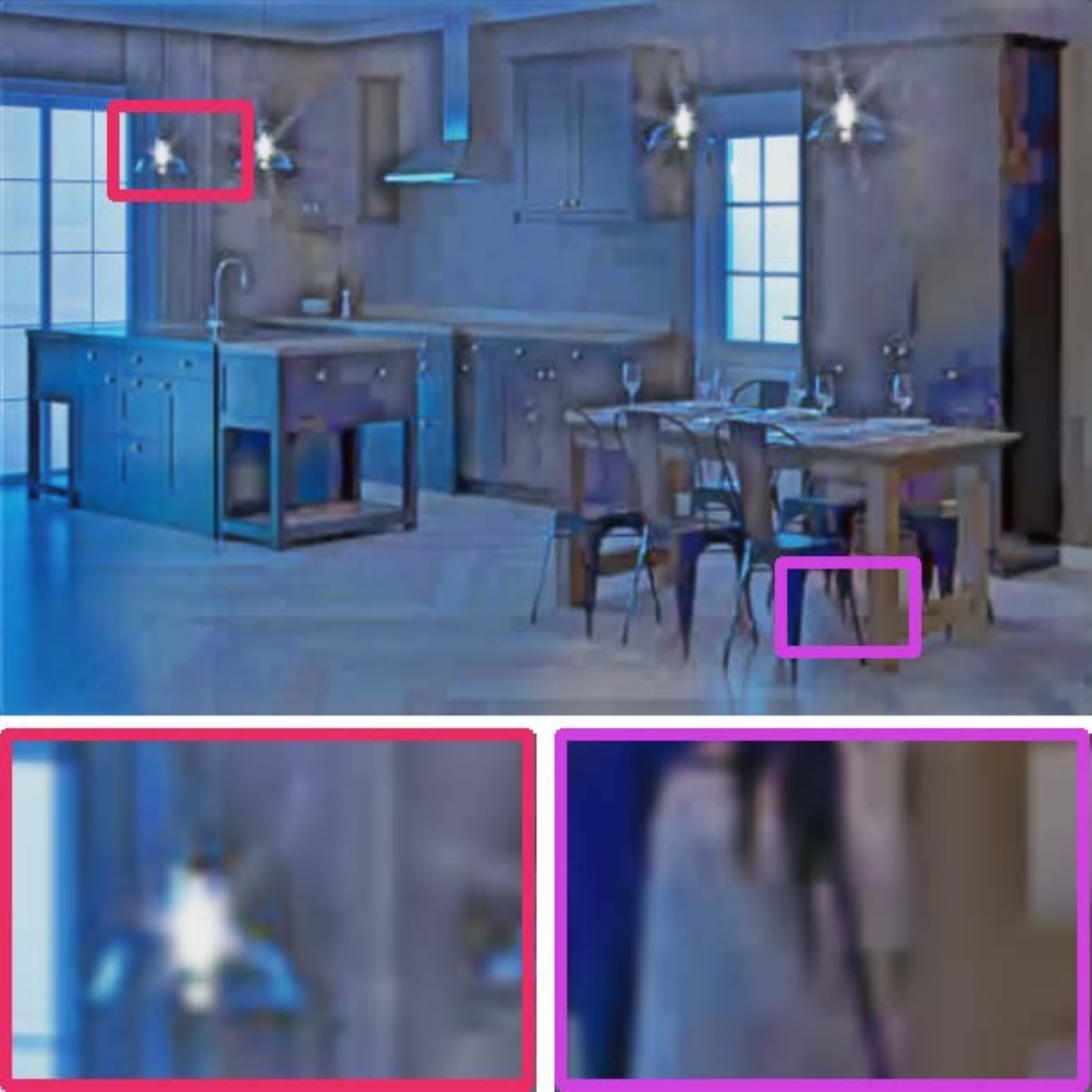}       &
    \includegraphics[width=\swten]{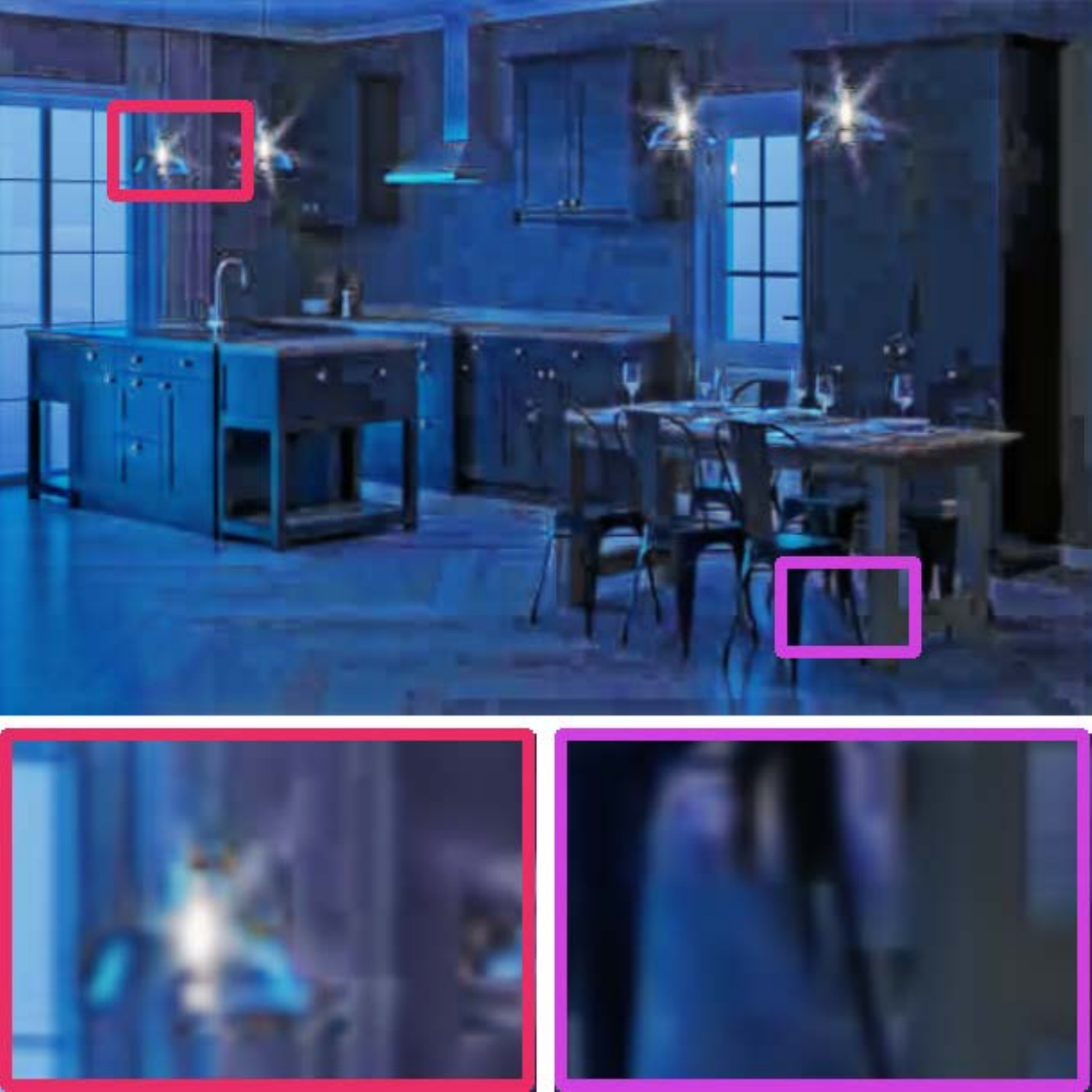}     &
    \includegraphics[width=\swten]{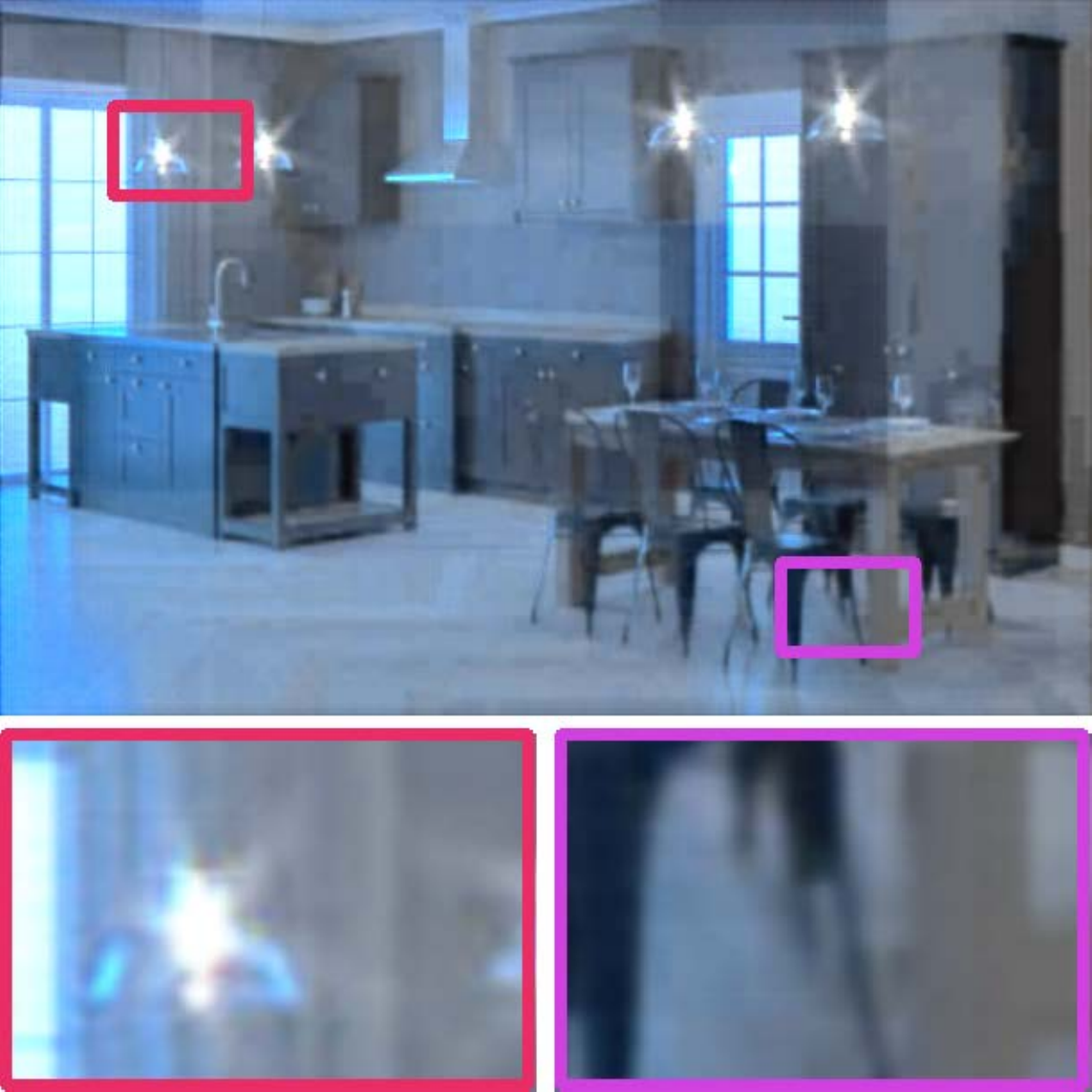}    &
    \includegraphics[width=\swten]{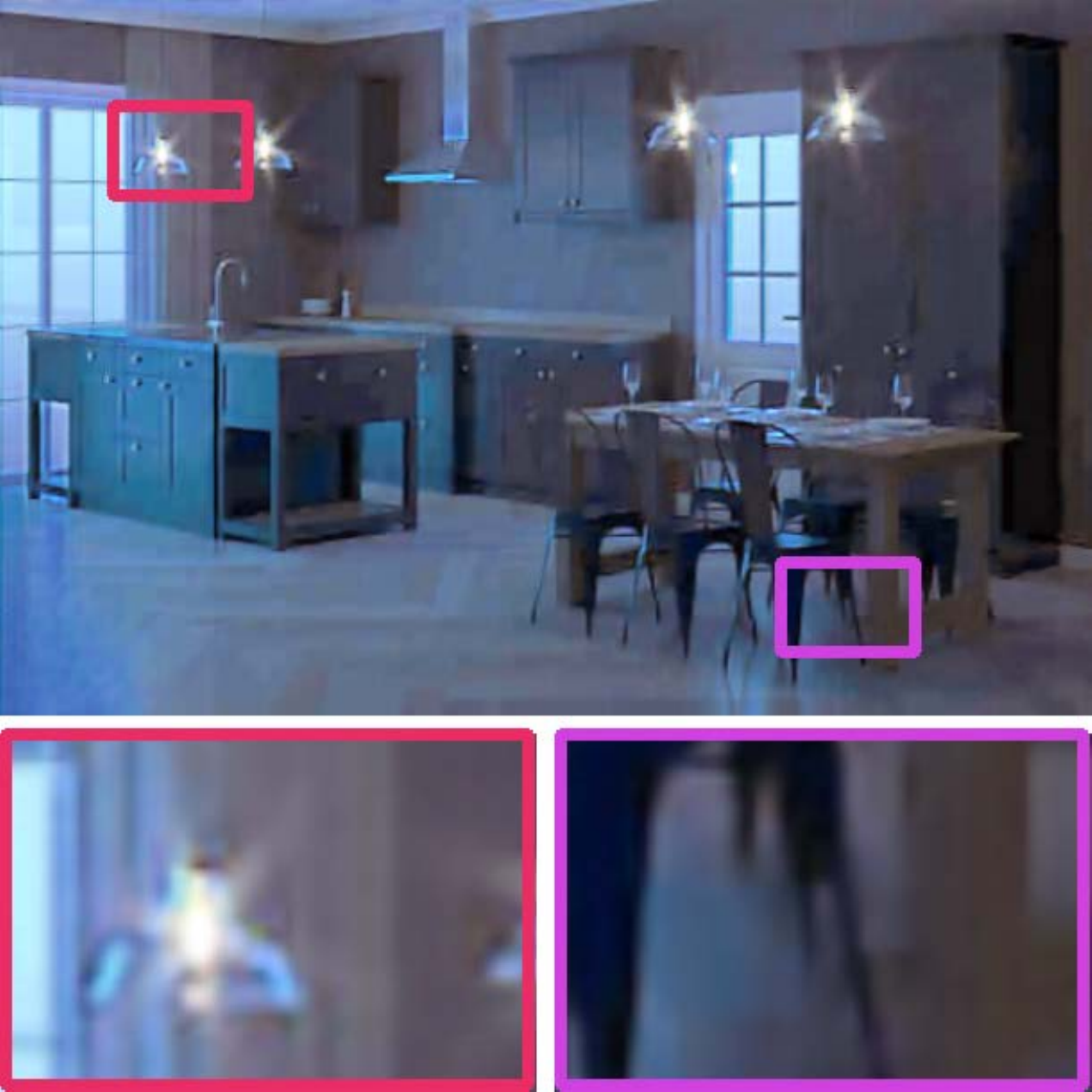}  &
    \includegraphics[width=\swten]{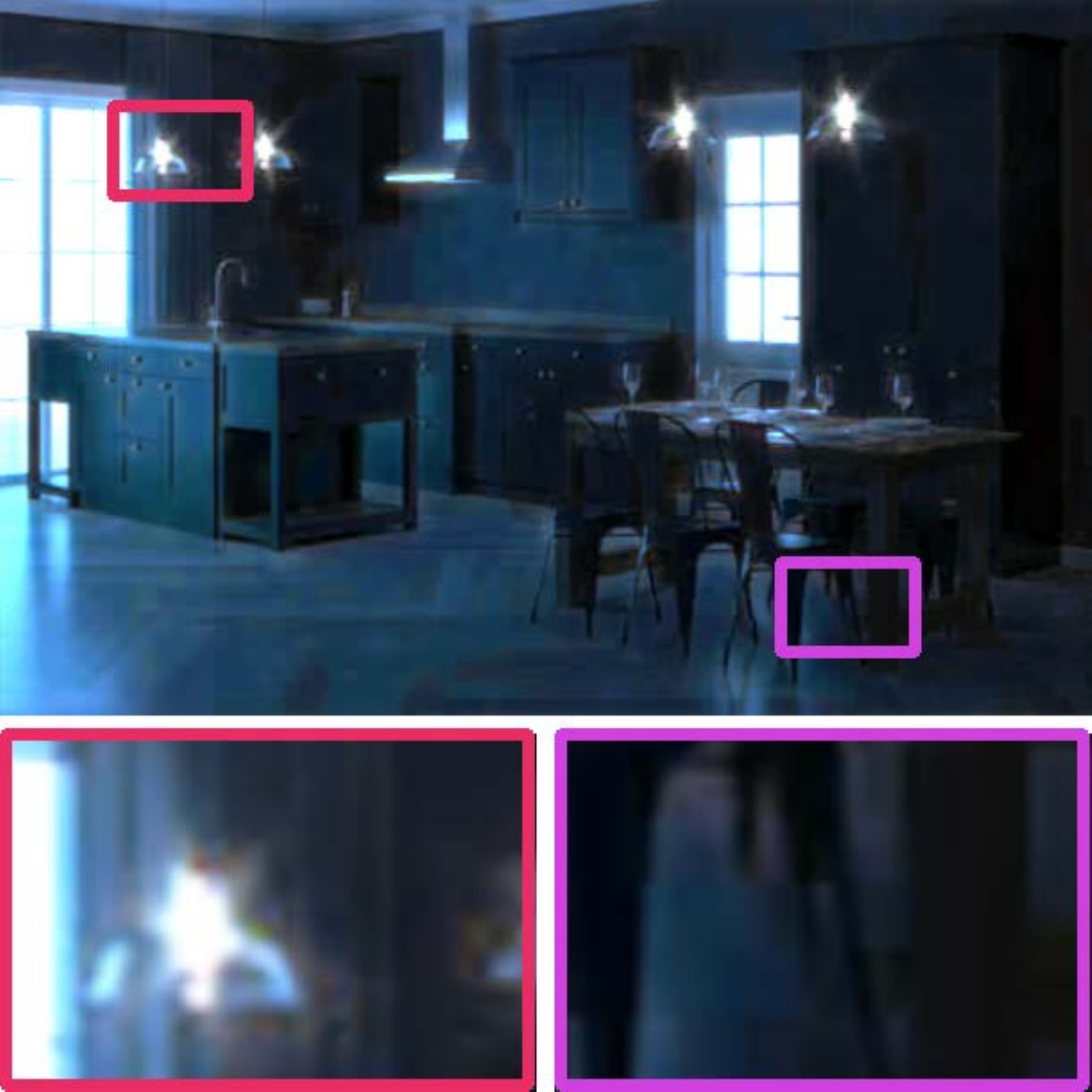}     &
    \includegraphics[width=\swten]{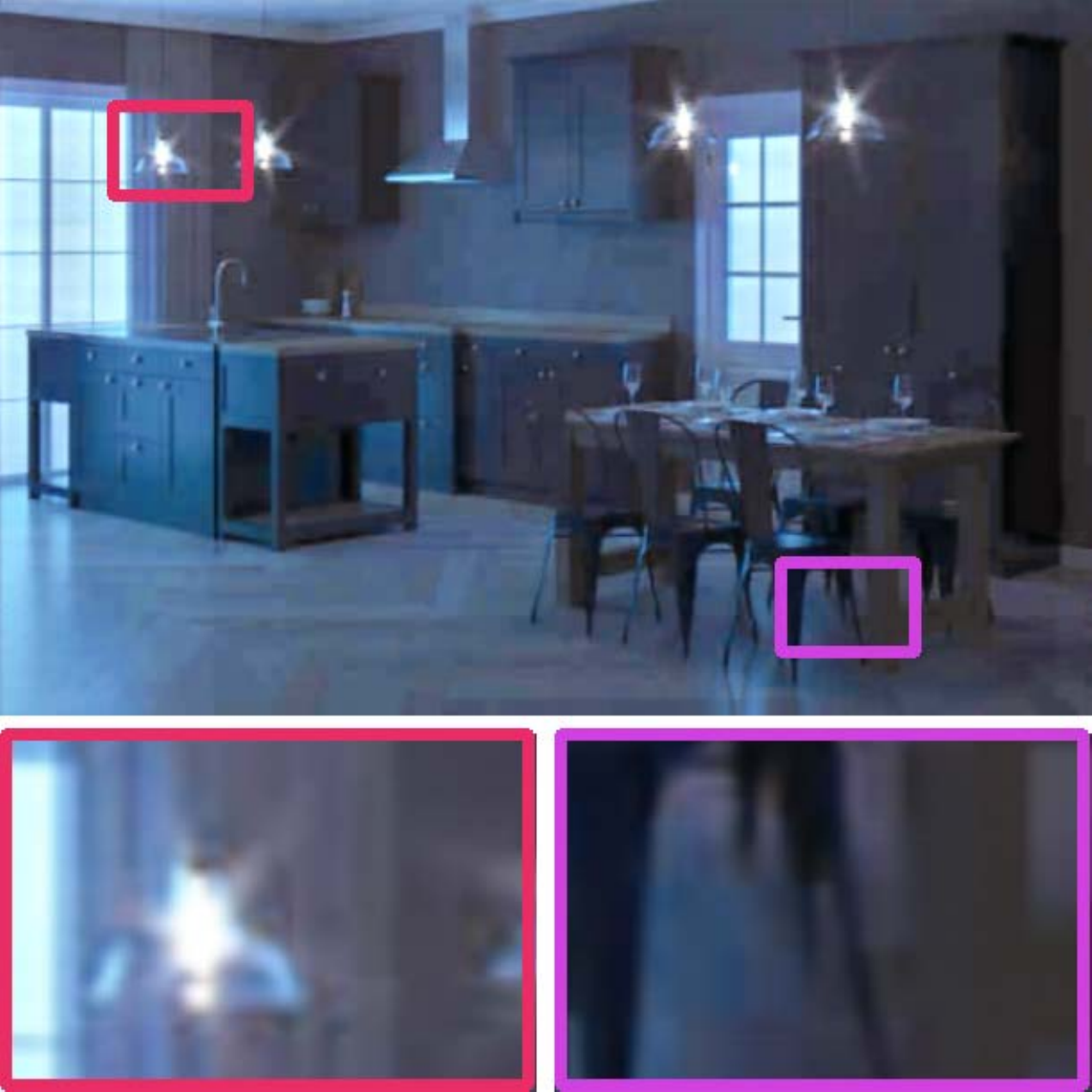}   &
    \includegraphics[width=\swten]{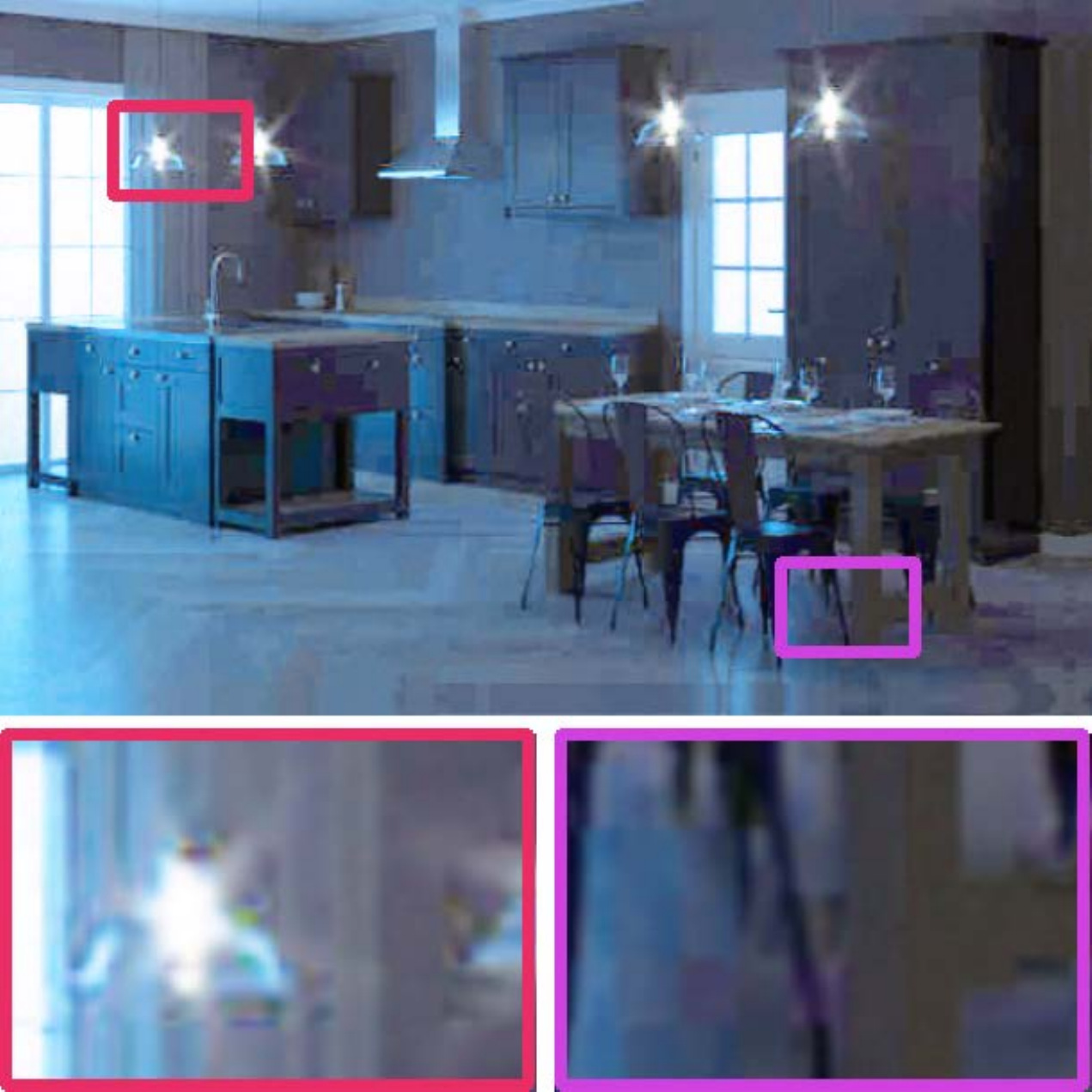} \\
    Input  & RetinexNet  & KinD & KinD++ & Uformer & Restormer & MIRNet & LLFormer  & LLDiffusion
\end{tabular}
\end{center}
    \caption{Visual comparison on DICM~\cite{liu2021benchmarking} (top), MEF~\cite{ma2015perceptual} (second row), NPE~\cite{wang2013naturalness} (third row) and our collected RWT (bottom). Other comparison methods cause color distortion or blurry/noisy textures. LLDiffusion can better enhance visibility and restore the image details. 
    }
    \label{fig:real-result}
\end{figure*}%
\textbf{Performance in real-world scenarios}. We further provide quantitative and qualitative comparisons of LLDiffusion with state-of-the-art LLIE methods on four real-world datasets. Table~\ref{tab:real-word} reports quantitative comparison results regarding PI and ILNIQE non-reference metrics. 
LLDiffusion outperforms all other methods across all evaluation metrics on the DICM, NEF, and RWT datasets. On NPE, LLDiffusion ranks first and second places regarding PI and ILNIQE respectively. In addition, Fig.~\ref{fig:real-result} shows qualitative results of different methods. The low-light images in the first column contain complex degradation in real-world scenarios. Other methods exhibit color distortion 
and cannot enhance the images well. For example, RetinexNet~\cite{wei2018deep}, KinD~\cite{zhang2019kindling}, and Uformer~\cite{wang2022uformer} produce the artifact of color distortion problem and fail to adjust the image contrast well. 
Recent methods Restormer~\cite{zamir2022restormer}, MIRNet~\cite{wang2022low}, and LLFormer~\cite{wang2022ultra} cannot remove the low light in the enhanced images (see the patches in the 5th-7th columns). However, LLDiffusion still works well in adjusting contrast and enhances more realistic image details in dark areas. Therefore, these comparison results show that integrating degradation representations in the diffusion module can significantly improve the robustness and generalization of the model in real-world scenarios.

\begin{figure*}
	\centering
 	 \includegraphics[width=0.70\textwidth]{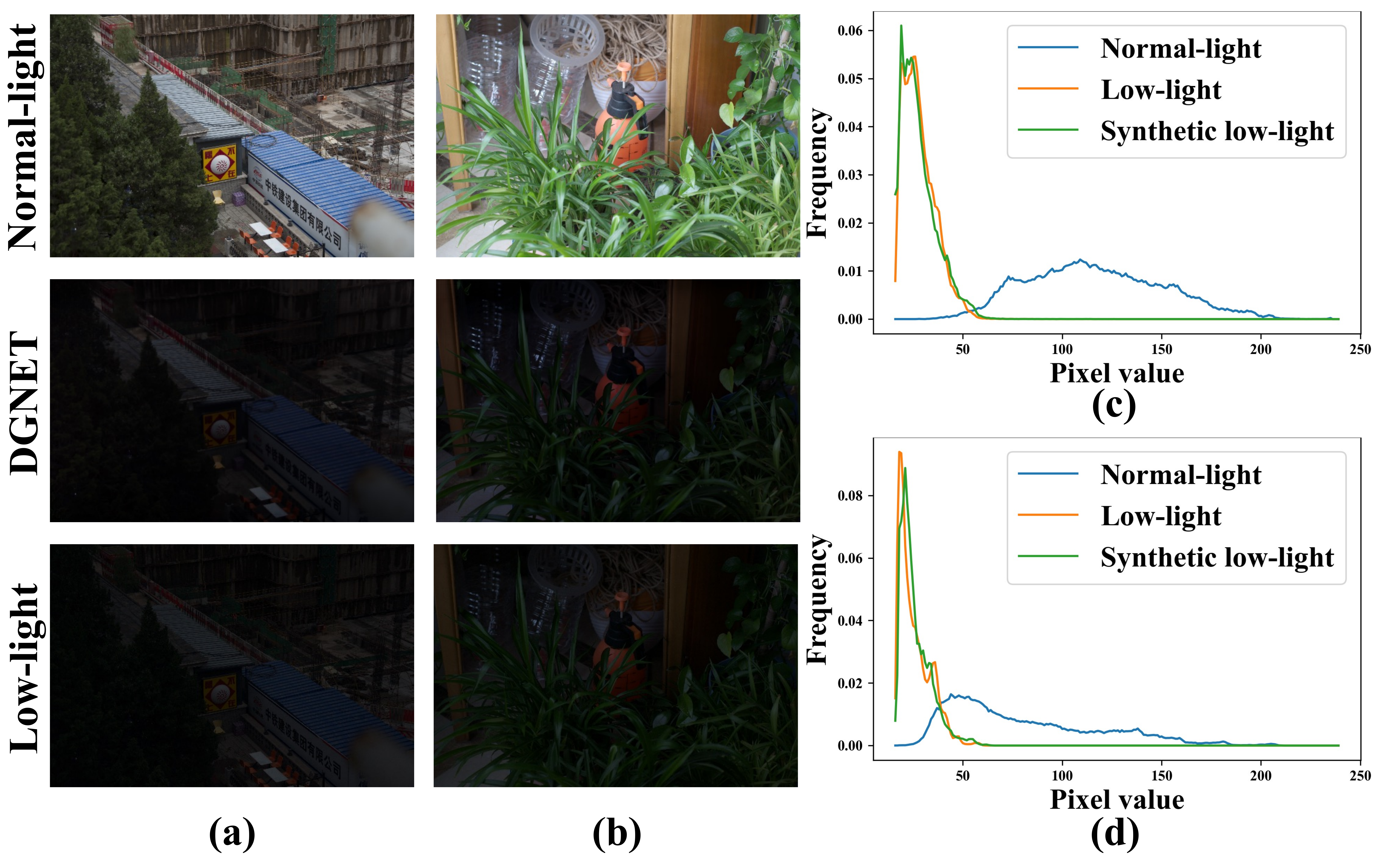}
  \caption{(a) and (b) are examples of synthetic images generated by DGNET. (c) and (d) are the corresponding Y-channel histograms in YCbCr color space~\cite{wei2018deep}.} 
	\label{fig:degradation}
\end{figure*}

\subsection{Ablation Studies}

We conduct ablation studies with regard to three aspects: the core components of the LLDiffusion, the generation of the pre-trained encoder $\mathbf{E}$, and the effect of different diffusion models.

\textbf{A. The core components of the LLDiffusion}. We conduct an ablation study by progressively reducing (1) the encoder module, (2) the DGNET module, and (3) the color map. The model without the encoder module does not use degradation representations, the model without DGNET does not benefit from joint learning, and the model without using the color map as a condition cannot benefit in enhancing color and brightness consistency from the image prior. Table~\ref{tab:network_structure} shows the results. Without integrating degradation representations (\ie without the encoder module) and the degradation network (\ie without the DGNET module) in the model, the PSNR values are reduced by $1.31$ dB and $1.03$ dB, respectively. Our LLDiffusion achieves the best performance in terms of LPIPS, which shows that our LLDiffusion produces the best perceived visual effect. In addition, we conduct an ablation experiment on the color map, and the results show that our LLDiffusion embedded with the color map as a condition achieves superior results compared to the variant without the color map as the condition. It shows that using the color map as an image prior to the diffusion can enrich the saturation and reduce color distortion, leading to improved image quality. To verify whether the DGNET truly learns the degradation process, we show some synthetic images produced by the learned degradation generation network in Fig.~\ref{fig:degradation}. The output images of DGNET are close to the low-light images, indicating that our DGNET can accurately learn the low-light degradation process and thus benefits the model.

%%%%%%% Three parts for ablation study 
\begin{table}[t]
\centering
\caption{{\bf Ablation study} of the core components used in LLDiffusion.}
\begin{tabular}{l|c|c|c|c|c}
    \toprule
    \multirow{2}{*}{Model} & \multicolumn{2}{c|}{Components} & \multirow{2}{*}{PSNR$\uparrow$} &\multirow{2}{*}{SSIM$\uparrow$} & \multirow{2}{*}{LPIPS$\downarrow$}\\ \cline{2-3}
     &encoder&DGNET& & & \\
    \midrule
    baseline & \xmark & \xmark &23.34 &0.803&0.191 \\ 
    w/o DGNET &\cmark &\xmark &23.62 &0.818&0.175 \\ 
    w/o color map & \cmark& \cmark & 23.98 &0.821&0.162\\ 
  LLDiffusion & \cmark& \cmark &\textbf{24.65} & \textbf{0.843} &\textbf{0.075}\\
    \bottomrule
\end{tabular}
\label{tab:network_structure}
\end{table}

% %%%%%%% Three parts for ablation study 
% \begin{table}[t]
% \centering
% \caption{{\bf Ablation study} of the LLDiffusion network structure.}
%  % \vspace{-2mm}
% % \vspace{-0.2cm}
% % \scalebox{0.82}{
% \begin{tabular}{l|c|c|c}
%     \toprule
%     \multirow{2}{*}{Model} & \multicolumn{2}{c|}{Components} & \multirow{2}{*}{PSNR} \\ \cline{2-3}
%      &encoder&DGNET& \\
%     \midrule
%     w/o degradation representations& \xmark & \xmark &23.34 \\ 
%     w/o degradation network &\cmark &\xmark &23.62  \\ 
%     LLDiffusion w/o color map & \cmark& \cmark&\textbf{23.98} \\ 
%     \rowcolor{Gray}  LLDiffusion & \cmark& \cmark&\textbf{24.65} \\
%     \bottomrule
% \end{tabular}
% % }
% % \vspace{-1mm}
% \label{tab:network_structure}
% \end{table}

\textbf{B. Generalization of the pre-trained encoder $\mathbf{E}$}. We further investigate the generalization ability of the proposed encoder for degradation representation learning. We first jointly train the proposed LLDiffusion on the LOL dataset to train the encoder $\mathbf{E}$, then freeze the encoder and train only the DDDM for low-light enhancement on the LOL dataset and evaluate the DDDM on the VE-LOL test set. Compared with the recent best method MIRNet~\cite{zamir2020learning} ($38.10$/$0.889$/$0.106$ in PSNR/SSIM/LPIPS), the proposed LLDiffusion achieves better performance ($30.04$/$0.901$/$0.055$ in PSNR/SSIM/LPIPS). The results show that our pre-trained encoder achieves strong generalization ability by encoding the degradation representations from low-light images.

\begin{table}[t]
\centering
\caption{
Comparison results with different diffusion-based models SR3~\cite{saharia2022image} and WeatherDiffusion~\cite{ozdenizci2022restoring} on the LOL dataset~\cite{wei2018deep}.}
 % \vspace{-2mm}
\begin{tabular}{l|c|c|c}
    \toprule
    Model & PSNR & SSIM &LPIPS\\
    \midrule
    SR3~\cite{saharia2022image}                    &  20.78 &0.612&0.243 \\
    WeatherDiffusion~\cite{ozdenizci2022restoring} & 22.41 &0.781& 0.145\\ % 17.48
    LLDiffusion & \textbf{24.65} & \textbf{0.843}& \textbf{0.075} \\
    \bottomrule
\end{tabular}
\label{tab:diffusion_models}
   % \vspace{-3mm}
\end{table} 
\textbf{C. Different diffusion models}. To the best of our knowledge, no diffusion model has been specifically designed for LLIE. Therefore, we have selected two existing diffusion models from other relevant tasks, retrained them for LLIE, and compared their performance. Two recent diffusion models (SR3~\cite{saharia2022image} and WeatherDiffusion~\cite{ozdenizci2022restoring}) are chosen to verify the effectiveness of the LLDiffusion embedding with the degradation representations. SR3~\cite{saharia2022image} is a diffusion model for image super-resolution, and WeatherDiffusion~\cite{ozdenizci2022restoring} is designed to deal with image restoration under different weather conditions. For a fair comparison, these two models are re-trained on the LOL dataset. Table~\ref{tab:diffusion_models} presents the quantitative results, which clearly demonstrate the superior performance of our LLDiffusion method compared to existing diffusion models. The results reveal significant improvements achieved by LLDiffusion in various evaluation metrics. For instance, when compared to SR3, LLDiffusion showcases remarkable enhancements with approximately 3.87 dB and 0.231 improvements in terms of PSNR and SSIM, respectively. Notably, LLDiffusion outperforms all other methods in terms of LPIPS, indicating its ability to generate more realistic and visually pleasing results.

%------------------------------------------------------------------------
\section{Limitations and Future Work}\label{sec:limitations}
%%%%%%%%%% use a short paragraph to introduce the limitations and Future work
Our proposed LLIE approach, LLDiffusion, has demonstrated superior performance compared to previous LLIE methods on popular LLIE datasets. However, there is still room for improvement in the performance of LLDiffusion. For example, incorporating new training techniques such as pre-trained strategies or integrating Retinex theory into LLDiffusion can further enhance its potential for low-light image enhancement. Additionally, we can improve the performance of LLDiffusion by increasing the width and depth of the latent map encoder, which currently has a simple structure. Finally, while LLDiffusion is currently designed for processing images, we are exploring whether it can also be used for low-light video enhancement in future work.

\section{Conclusion} \label{sec:conclusion}
%%%%%%%% write the conclusion part of this work 
In this paper, we propose a novel degradation-aware learning scheme that integrates degradation into the image enhancement process. The motivation behind this scheme is to leverage the knowledge of degradation patterns to improve the quality and visual appeal of low-light images. To achieve this, we propose a joint model that effectively learns degradation representations from low-light images. To utilize the learned degradation representations, we propose LLDiffusion that introduces a dynamic degradation-aware diffusion module in the diffusion model. This module plays a crucial role in the image enhancement process by conditioning both the color map and the degradation representations. By incorporating these conditioning factors, the diffusion module can guide the enhancement process in a more informed and precise manner, leading to improved image quality. We conduct quantitative evaluations and comparisons with existing LLIE methods on widely-used benchmark datasets. Experimental results demonstrate that our LLDiffusion outperforms existing LLIE methods both quantitatively and qualitatively.

\section*{Acknowledgement}
This work was supported in part by the National Natural Science Foundation of China (Grant No. 61672273, 61832008), in part by Shenzhen Science and Technology Program (No. JSGG20220831093004008, JCYJ20220818102012025).

\section*{Data Availability Statement}
The datasets generated during and/or analyzed during the current study are available in the 
awesome-low-light-image-enhancement repository, with the link as https://github.com/dawnlh/awesome-low-light-image-enhancement.
{\small
\bibliographystyle{spmpsci}
% \bibliography{egbib}
\bibliography{paper}
}

\end{document}